\documentclass[letterpaper, 10 pt, conference]{ieeeconf}  %

\IEEEoverridecommandlockouts                              %

\usepackage{amsmath} %

\usepackage{amssymb}  %
\usepackage{color}
\usepackage[table,xcdraw]{xcolor}
\usepackage{siunitx}
\usepackage{hhline}
\usepackage{tikz}
\usetikzlibrary{calc}
\usepackage{graphicx} %

\let\labelindent\relax
\usepackage{enumitem}

\usepackage{booktabs}
\usepackage{multirow}
\usepackage{xifthen}
\usepackage{nicefrac}

\usepackage[noadjust]{cite}

\usepackage{lipsum}
\usepackage{todonotes}

\makeatletter
\let\NAT@parse\undefined
\makeatother
\usepackage[bookmarks=false]{hyperref}

\usepackage{dblfloatfix}

\usepackage{fancyhdr}
\usetikzlibrary{positioning}

\title{\LARGE \bf
PanopticNDT: Efficient and Robust Panoptic Mapping%
}

\def\theauthors{Daniel Seichter, Benedict Stephan, Söhnke Benedikt Fischedick, Steffen Müller, Leonard Rabes, and \\Horst-Michael Gross}
\author{\theauthors%
\thanks{Authors are with Neuroinformatics and Cognitive Robotics Lab, Technische Universit\"at Ilmenau, 98693 Ilmenau, Germany. Contact: \hfill\newline
{\scriptsize{\tt daniel.seichter@tu-ilmenau.de}, ORCID: {\tt
\href{https://orcid.org/0000-0002-3828-2926}{0000-0002-3828-2926}}}}%
\thanks{This work has received funding from the German Federal Ministry of Education and Research (BMBF) to the project MORPHIA (16SV8426).}%
}

\makeatletter
\newcommand\thefontsize[1]{Fontsize for \string#1 : {#1 \f@size pt}}
\IEEEtriggercmd{\reset@font\normalfont\fontsize{7.9pt}{8.9pt}\selectfont}
\makeatother
\IEEEtriggeratref{1}

\makeatletter
\def\appendixtitle{
    \twocolumn[%
        \begin{center}%
            \vskip0.25in{\LARGE [\thesection~Appendix]\\[1mm]\@title\par}\vskip1.0em\par%
            {\lineskip.5em\sublargesize\theauthors\par}%
            \vspace{10mm}
        \end{center}%
    ]%
}
\makeatother

\begin{document}

\maketitle

\newboolean{isarxiv}
\setboolean{isarxiv}{true}
\ifthenelse{\boolean{isarxiv}}{%
    \renewcommand{\headrulewidth}{0pt}
    \fancypagestyle{fancyfirstpage}{%
        \fancyhf{}%
        \fancyhead[C]{%
            \footnotesize%
            \raisebox{0.3cm}[0cm][0cm]{%
                \parbox{\textwidth}{
                    \centering
                    \textcolor{gray}{%
                        © 2023 IEEE.  Personal use of this material is permitted. 
                        Permission from IEEE must be obtained for all other uses, in any current or future media, including reprinting/republishing this material for advertising or promotional purposes, creating new collective works, for resale or redistribution to servers or lists, or reuse of any copyrighted component of this work in other works.
                    }%
                }%
            }%
        }
        \fancyfoot[C]{%
            \footnotesize%
            \textcolor{gray}{\thepage}%
        }
    }
    \fancypagestyle{fancypage}{%
        \fancyhf{}%
        \fancyfoot[C]{%
            \footnotesize%
            \textcolor{gray}{\thepage}%
        }        
    }    
    \thispagestyle{fancyfirstpage}
    \pagestyle{fancypage}
}{%
    \thispagestyle{empty}%
    \pagestyle{empty}%
}%

\begin{abstract}
As the application scenarios of mobile robots are getting more complex and challenging, scene understanding becomes increasingly crucial.
A mobile robot that is supposed to operate autonomously in indoor environments must have precise knowledge about what objects are present, where they are, what their spatial extent is, and how they can be reached; i.e., information about free space is also crucial.
Panoptic mapping is a powerful instrument providing such information.
However, building 3D panoptic maps with high spatial resolution is challenging on mobile robots, given their limited computing capabilities.
In this paper, we propose PanopticNDT~--~an efficient and robust panoptic mapping approach based on occupancy normal distribution transform (NDT) mapping. 
We evaluate our approach on the publicly available datasets Hypersim and ScanNetV2. 
The results reveal that our approach can represent panoptic information at a higher level of detail than other state-of-the-art approaches while enabling real-time panoptic mapping on mobile robots.
Finally, we prove the real-world applicability of PanopticNDT with qualitative results in a domestic application.
\end{abstract}

\section{Introduction}
\label{sec:introduction}
Mobile robots that are supposed to operate autonomously in complex indoor environments must have a detailed scene understanding.
The robot needs to have precise semantic knowledge about the scene as well as each object within.
Panoptic mapping combines and integrates this information over time into a 3D representation, enabling the robot to have a broad understanding of its environment.

\begin{figure}[!t]
    \vspace{1mm}%
	\centering%
	\hspace*{-0.7mm}%
	\begin{tikzpicture}[scale=1.0]%
	    \node[anchor=north west] at (0, 0){%
	        \includegraphics[width=0.995\columnwidth, trim=2cm 0cm 0.6cm 3cm, clip]{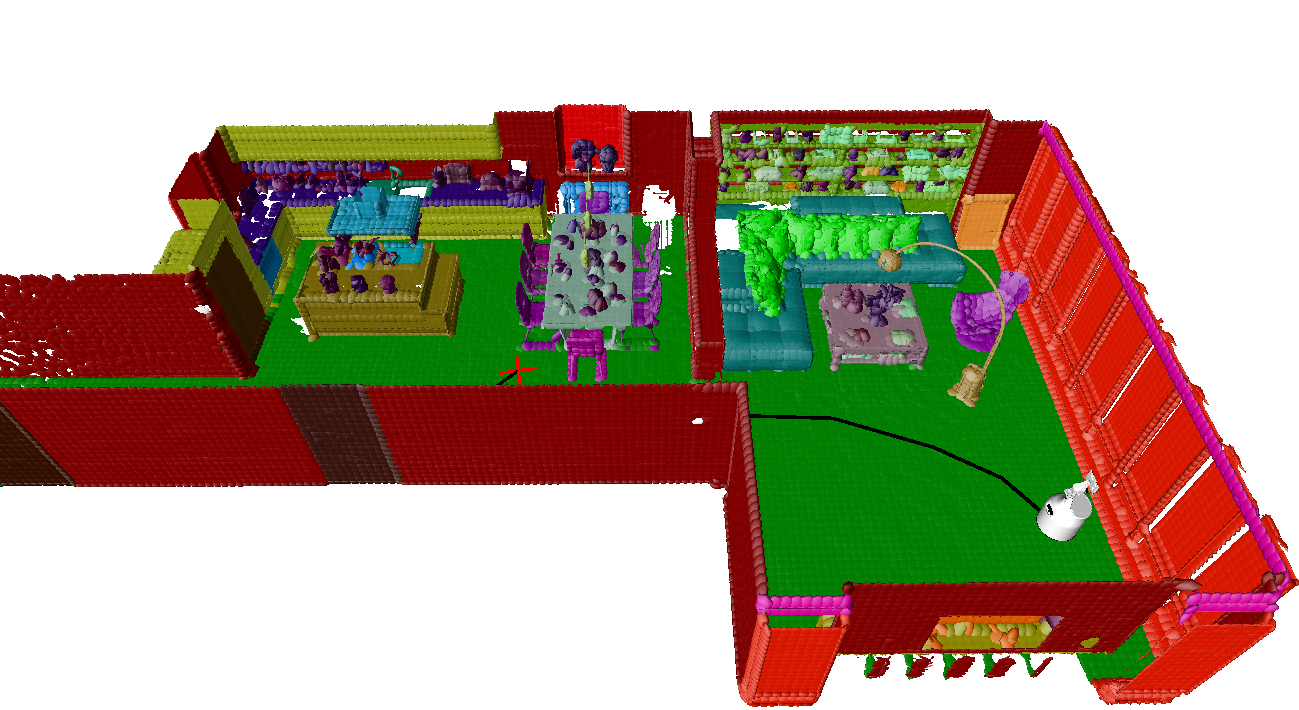}%
	    };
	    \node[anchor=north west] at (0.6, -2.35){%
	        \fcolorbox{black!40}{white}{%
	            \includegraphics[width=0.38\columnwidth, trim=2cm 0cm 0.6cm 3cm, clip]{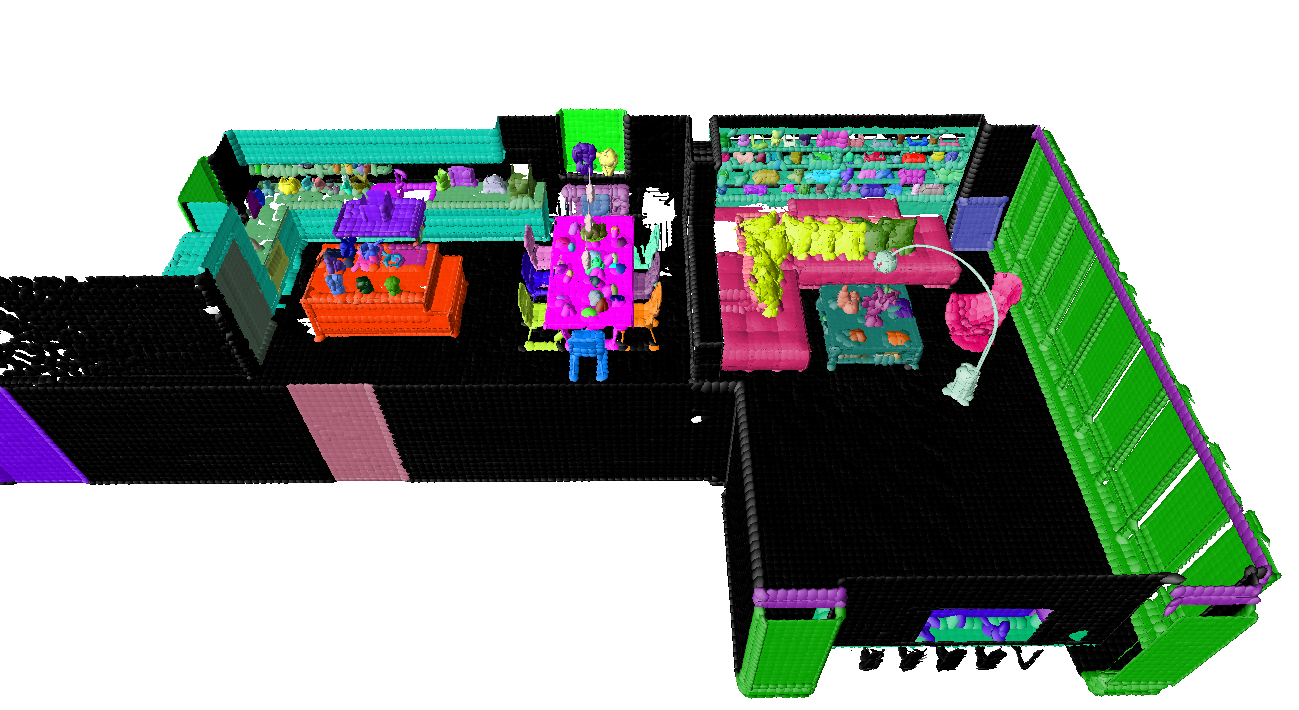}%
	        }
	    };	    
	\end{tikzpicture}%
	\vspace{-5mm}
    \caption{%
        Panoptic occupancy NDT (P-NDT) map built with predicted panoptic segmentation of EMSANet~\cite{emsanet2022ijcnn} and voxel size of 10\si{\centi\meter} for scene \emph{ai\_051\_001} of the Hypersim~\cite{hypersim-iccv2021} test split. %
        Bottom: corresponding instance map. %
        Best viewed in color at 300\%. %
        Black indicates \emph{no\_instance}, see Fig.~\ref{fig:experiments:hypersim_radar_chart} for semantic colors. %
        Panoptic is visualized by small color differences.%
    }%
    \label{fig:introduction:eyecatcher}
    \vspace{-5mm}
\end{figure}

In our research projects CO-HUMANICS~\cite{Cohumanics-Fischedick-ISR-2023} and MORPHIA~\cite{morphia-isr2022}, our mobile robots act autonomously in domestic environments and allow relatives as well as caregivers to stay in contact with care-dependent people to let them participate in social life.
We aim to enable inexperienced operators to remotely control our robots in such an environment.
As shown in Fig~\ref{fig:introduction:eyecatcher}, the operator should be able to send the robot to a specific target, e.g., the table with eight chairs.
Moreover, it should find a suitable waiting position, which, for example, involves not blocking other chairs or taking a suitable position for further inspection via the camera.
As domestic environments are typically dynamic, we rely on a robust long-term localization via RTABMap~\cite{rtabmap-JFR-2019} and build short-term panoptic 3D representations of the current environment periodically.
This enables gaining and representing panoptic knowledge while still being able to react to small but important changes, such as chair arrangements.
In mobile robotics, voxel-based 3D representations are common due to their efficient processing~\cite{octomap-AR-2013, gpoctomap-ICRA-2016, bkgi-octomap-ICRA-2017}. 
While there are already approaches for semantic mapping~\cite{semantic-octree-ICRA-2015, semantic-3d-occ-IROS-2017, semantic-gp-map-ArXiV-2017, semantic-bsk-map-RAL-2020} extending these classical approaches, research focusing on panoptic mapping is rare.
Existing approaches typically do not focus on mobile applications and, thus, do not meet our requirements for efficient and robust panoptic mapping.

Therefore, in this paper, we propose panoptic normal distribution transform~(PanopticNDT) mapping~--~an approach that addresses the requirements of such a scenario. 
It builds upon the fast and robust occupancy NDT mapping~\cite{Einhorn-ECMR-2013-NDT, Einhorn-PHD-2019-NDT} and its semantic extension~\cite{semanticmapping2022icra}. 
We examine how to further incorporate and track instance information in order to realize panoptic NDT maps. 
We show that our approach enables precise and efficient panoptic mapping at sub-voxel level at a framerate of \textasciitilde2.75\si{\hertz} on our mobile robots.

For evaluating, we conduct experiments on two indoor benchmark datasets: the large-scale synthetic Hypersim~\cite{hypersim-iccv2021} dataset and the real-world ScanNetV2 dataset~\cite{scannet-cvpr2017}.
Unlike other approaches, we first evaluate the capability of our approach to represent panoptic information independently of a preceding segmentation step, i.e., we use the ground-truth annotations for mapping.
Subsequently, we switch to panoptic predictions of our EMSANet~\cite{emsanet2022ijcnn}~--~an efficient RGB-D panoptic segmentation approach~--~to evaluate mapping in a more realistic setting.
For reporting results, we rely on the ScanNetV2 benchmark pipeline.
We extend the existing pipeline with a complementary panoptic evaluation task in 2D and 3D and prepare Hypersim to be used in that pipeline.
The quantitative results of our experiments demonstrate the performance and robustness of our approach.
Finally, we present qualitative results in a real domestic environment, showing the real-world performance of our approach.

In order to enable other researchers to have a similar setup and to compare to our panoptic mapping approach, we share the pipeline for data preparation and evaluation as well as the training details and weights of all neural networks involved
at GitHub: \href{https://github.com/TUI-NICR/panoptic-mapping}{\small \url{https://github.com/TUI-NICR/panoptic-mapping}}.

\section{Related Work}
\label{sec:related_work}
Although many representations have been proposed~\cite{mesh-grasping-MMM-2017, octomap-AR-2013, shape-repr-CVPR-2018}, in mobile robotics, voxel-based representations are preferred due to their efficient processing and memory requirements.
In the following, we summarize voxel-based representations and, subsequently, focus on integrating additional semantic and panoptic information.

\subsection{Voxel-based Data Representations}
\label{sec:related_work:voxel}
In voxel-based representations, 3D space is divided into a regular grid of voxels with a predefined size.
These voxels are typically managed using optimized data structures, such as octrees~\cite{octomap-AR-2013} or voxel hashing~\cite{voxelhasing-acm-2013}.
To enable downstream applications, each voxel usually stores an occupancy value, indicating its state either as occupied, free, or unknown.

However, the precision of such maps still highly depends on the used voxel size. 
Decreasing the voxel size heavily increases memory and computational requirements and, thus, is not feasible for mobile applications.
Therefore, other approaches attempt to additionally model the surface of each voxel in order to enable higher precision at the same grid resolution.
For example, truncated signed distance fields~(TSDF)~\cite{tsdf1996pamcgit, voxblox-iros-2017} store the distance to the nearest surface, allowing a more accurate representation of its shape and position.
However, for efficiency reasons, occupancy is only modeled and stored in a truncated area around the surface of objects. 
This can lead to missing information about unseen areas required for downstream applications in mobile robotics, such as obstacle avoidance or path planning.
In contrast to that, surfel maps~\cite{surfel-GCR-2012} use so-called surface elements characterized by the eigenvectors of the data distribution inside a voxel. 
However, this representation makes use of the flat surface assumption that becomes insufficient when modeling objects smaller than the grid resolution. 
A similar representation is the normal distribution transform (NDT)~\cite{ndt-IROS-2003-Biber, ndt-JFR-2007-Magnusson, Einhorn-ECMR-2013-NDT}. 
In contrast to surfel maps, NDT maintains the whole covariance matrix of the data distribution inside a voxel to model its spatial extents in terms of a normal distribution.

As the NDT representation is able to capture a lot of details~\cite{ndt-ICRA-2011-Stoyanov} even with larger voxel sizes and can be efficiently updated~\cite{Einhorn-ECMR-2013-NDT}, we follow this approach.

\subsection{Semantic Mapping}
\label{sec:related_work:semantic_mapping}
In contrast to the aforementioned approaches, semantic mapping approaches do not only store an occupancy value in each voxel but also semantic information.
The semantic information is retrieved by a preliminary semantic segmentation step using conditional random fields~(CRF)~\cite{semantic-octree-ICRA-2015} or approaches based on convolutional neural networks~(CNN)~\cite{semantic-3d-occ-IROS-2017, semanticmapping2022icra}.
Most approaches have in common that a histogram is used to model the information.
Approaches, such as~\cite{semantic-gp-map-ArXiV-2017, bkgi-octomap-ICRA-2017}, create a multi-class problem~(free$\,$+$\,$semantic classes) to track occupancy and semantic information in a single histogram.
However, as this requires sampling data points for free space from range measurements, these approaches are susceptible to over-representing the free space class. 
As shown in~\cite{semanticmapping2022icra}, this effect gets further intensified when incorporating noisy segmentations.
Therefore, \cite{semantic-octree-ICRA-2015, semantic-3d-occ-IROS-2017, semanticmapping2022icra} decouple both information and maintain occupancy and semantic independently.

In~\cite{semanticmapping2022icra}, we have shown that semantic information can be integrated in the occupancy normal distribution transform mapping of~\cite{Einhorn-ECMR-2013-NDT, Einhorn-PHD-2019-NDT}, enabling efficient and precise semantic mapping at sub-voxel level.
Therefore, our panoptic mapping approach builds on top of this work.

\subsection{Panoptic Mapping}
\label{sec:related_work:panoptic_mapping}
Panoptic mapping is challenging as it aims to create a representation of the environment including both, semantic and instance information, where each instance ID must be globally unique.
Thus, panoptic mapping faces the additional challenge that observations are not independent anymore:~instances already present in the map must be matched with the instances of the current observation before integrating.
Similar to semantic mapping, panoptic mapping is usually done in multiple steps, featuring one or multiple approaches for panoptic segmentation, matching, and the mapping step.
These steps are often further followed by a map refinement. 

Most approaches in this context build upon Voxblox~\cite{voxblox-iros-2017}~--~a method that creates voxel-based TSDF maps.
Panoptic Fusion~\cite{panoptic-fusion-2019-iros}, for example, additionally stores a single panoptic ID and a corresponding weight in each voxel.
For panoptic segmentation, the output of PSPNet~\cite{PSPNet-cvpr2017} and Mask R-CNN~\cite{MaskRCNN-iccv2017} is combined.
Instance inconsistency is resolved by projecting the current map back to the camera plane and an intersection over union (IoU)-based matching.
The weight in a voxel is increased with matching observations and decreased when observations do not match.
The panoptic label gets updated or replaced depending on that weight. 
Voxblox++~\cite{voxblox++-ral-2019} also extends~\cite{voxblox-iros-2017} but only focuses on instance mapping.
An additional geometric depth segmentation divides each instance into segments.
The map represents these segments while additionally keeping track of corresponding instance labels.
The integration of new observations is similar to~\cite{panoptic-fusion-2019-iros} but done in 3D.
In~\cite{cad-panoptic-mapping-2021-icra}, a further extension to panoptic mapping is proposed. 
However, it mainly focuses on fitting CAD models into the map.
The recent Panoptic Multi-TSDFs~\cite{panoptic-multi-tsdf-2022-icra} also builds on top of Voxblox but takes a different approach.
It uses a collection of submaps to represent the geometry of individual entities such as object instances, the background class, or free space.
By combining these submaps and explicitly modeling free space, a full volumetric map can be derived.
Each submap has a different voxel resolution depending on the semantic class it is supposed to represent.
To efficiently integrate panoptic information, a label tracking approach based on IoU, similar to~\cite{panoptic-fusion-2019-iros}, is used.

All approaches have in common that they do not model the underlying distribution of panoptic observations integrated into the map, making them less robust to misclassifications.
Moreover, none of the proposed pipelines for panoptic mapping is real-time capable on a mobile robot due to computationally expensive segmentation approaches, such as Mask R-CNN or PSPNet, or additional map refinement.

\section{Efficient and Robust Panoptic NDT-Mapping}
\label{sec:main}
\begin{figure}[!b]
    \vspace{-5mm}
	\centering
	\begin{tikzpicture}[scale=1.0]
	    \node at (0, 0){%
	        \includegraphics[width=0.95\columnwidth]{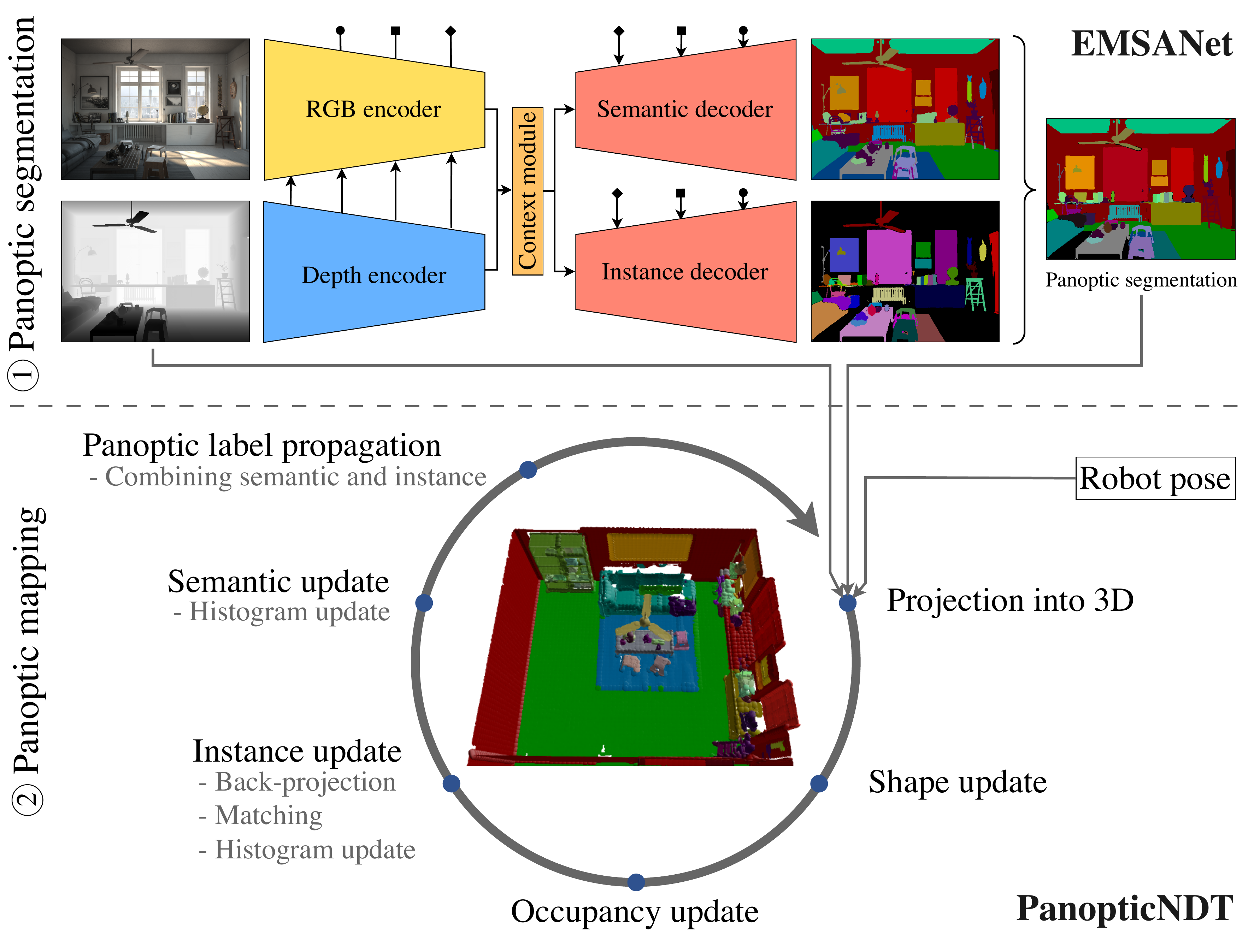}
	    };
	\end{tikzpicture}%
	\vspace{-6mm}
    \caption{%
        Overview of our two-step approach for panoptic mapping.
    }%
    \label{fig:main:system}
\end{figure}

Our panoptic mapping pipeline is depicted in~Fig.~\ref{fig:main:system} and is implemented using the middleware for robotic applications~(MIRA)~\cite{MIRA-Einhorn-IROS-2012}.
Given a precise localization in the environment, our approach comprises two steps. 
We first apply EMSANet~\cite{emsanet2022ijcnn}~--~an efficient RGB-D panoptic segmentation approach that extends ESANet~\cite{esanet2021icra}~--~to the current set of input images (color and depth).
Unlike other approaches~\cite{semantic-octree-ICRA-2015, semantic-3d-occ-IROS-2017, semantic-gp-map-ArXiV-2017, semantic-bsk-map-RAL-2020, 3d-semantic-segementation-2019-wacv, panoptic-fusion-2019-iros, voxblox++-ral-2019, cad-panoptic-mapping-2021-icra, panoptic-multi-tsdf-2022-icra}, we decided in favor of an RGB-D approach, as depth provides complementary information that helps to segment cluttered indoor scenes~\cite{esanet2021icra, emsanet2022ijcnn}.
Afterward, the obtained panoptic segmentation, the depth image, and the current pose are passed to the mapping stage.
In the following, we describe both parts of the pipeline in detail.

\subsection{Panoptic Segmentation}
\label{sec:main:segmentation}
Panoptic segmentation aims to assign a panoptic label~$P(\boldsymbol{u}) \in \mathcal{P}$ to each input pixel, with $\boldsymbol{u} \in \mathbb{R}^2$ denoting the corresponding normalized image coordinates.
To derive a pixel's panoptic label, both a semantic prediction~$L(\boldsymbol{u}) \in \mathcal{L}$ and an instance prediction~$Z(\boldsymbol{u}) \in \mathcal{Z}$ must be merged.
Panoptic segmentation distinguishes two categories of semantic classes:~\emph{thing} classes~$\mathcal{L}^{\text{Th}}$~--~such as chair, table, or cabinet~--~for countable objects; and \emph{stuff} classes~$\mathcal{L}^{\text{St}}$~--~such as wall, floor, or ceiling~--~for non-countable objects, such that $\mathcal{L}^{\text{Th}} \cup \mathcal{L}^{\text{St}} = \mathcal{L}$ and $\mathcal{L}^{\text{Th}} \cap \mathcal{L}^{\text{St}} = \emptyset$.
An instance ID~$z \in \mathbb{N}_{>0}$ is only assigned if a pixel belongs to one of the thing class, and $z=0$ denotes \emph{no\_instance}.

For efficiency reasons, we use the lightweight EMSANet-R34-NBt1D~(enhanced dual ResNet34-based encoder utilizing the Non-Bottleneck-1D block (NBt1D)~\cite{ERFNet-its2018}) for panoptic segmentation.
EMSANet~\cite{emsanet2022ijcnn} predicts dense semantic and instance information in a bottom-up fashion~(see Fig.~\ref{fig:main:system}).
The first task decoder directly assigns a semantic class to each pixel.
The second task decoder predicts instance centers and instance center assignments.
Both instance predictions are fused for pixels belonging to thing classes to obtain a class-agnostic instance segmentation. 
Note, unlike other top-down approaches~\cite{panoptic-fusion-2019-iros, voxblox++-ral-2019, cad-panoptic-mapping-2021-icra} that utilize instance segmentation approaches, such as Mask-RCNN~\cite{MaskRCNN-iccv2017}, this bottom-up approach leads to consistent instance IDs, i.e., there are no overlaps that must be resolved to determine a single instance ID for each pixel.
The panoptic label $P(\boldsymbol{u})$ finally combines semantic and instance information in a tuple and assigns the most frequent semantic class for each instance:
\begin{equation}
\label{eq:panoptic_label_2d}
P(\boldsymbol{u}) =%
\left\{%
    \begin{aligned}
        &\langle \underset{\boldsymbol{u}^\prime \in \{\boldsymbol{i} | Z(\boldsymbol{i}) = Z(\boldsymbol{u})\}}{\mathrm{mode}(L(\boldsymbol{u}^\prime))}, Z(\boldsymbol{u}) \rangle &\,& L(\boldsymbol{u}) \in \mathcal{L}^{\text{Th}} \\
        &\langle L(\boldsymbol{u}),  0 \rangle &\,&\mathrm{otherwise.}
    \end{aligned}
\right.%
\end{equation}
Note, Eq.~\ref{eq:panoptic_label_2d} emphasizes the instance prediction, i.e., the semantic prediction might be overruled by the semantic mode for a predicted instance. 
In the following, we denote the elements in the tuple $P(\boldsymbol{u})$ as $\langle L^P(\boldsymbol{u}), Z^P(\boldsymbol{u}) \rangle$.
Furthermore, we extend EMSANet to also provide dense confidence scores for the panoptic label and both elements to account for weak predictions in the subsequent mapping stage.
We denote these scores as $S^{L^P}(\boldsymbol{u})$, $S^{Z^P}(\boldsymbol{u})$, and $S^P(\boldsymbol{u}) = S^{L^P}(\boldsymbol{u}) \cdot S^{Z^P}(\boldsymbol{u})$ for semantic, instance, and panoptic, respectively.

\subsection{Panoptic Mapping: Overview}
\label{sec:main:mapping_overview}
The panoptic mapping stage aims to integrate the predicted panoptic labels $P(\boldsymbol{u})$ over time into a consistent 3D map.
Unfortunately, direct integration, as for semantic mapping, is not possible.
The predicted instance IDs~$Z^P(\boldsymbol{u})$ are inconsistent for the same object over multiple images processed in the first stage. 
Therefore, an instance matching and tracking step is required.
This step can be done in 2D or 3D.
We perform this step in 2D and project the 3D map back to the camera plane to match instances, similar to~\cite{panoptic-fusion-2019-iros, voxblox++-ral-2019, cad-panoptic-mapping-2021-icra, panoptic-multi-tsdf-2022-icra}. 
Afterward, the instance and semantic information in each voxel of the map is updated according to the current observation.
We introduce the underlying data representation in Sec.~\ref{sec:main:mapping_data} and describe the instance and semantic update step in Sec.~\ref{sec:main:mapping_instance} and Sec.~\ref{sec:main:mapping_semantic}, respectively. 
Note, unlike other approaches~\cite{panoptic-fusion-2019-iros, cad-panoptic-mapping-2021-icra, panoptic-multi-tsdf-2022-icra}, we follow the class-agnostic bottom-up idea of EMSANet also for mapping, i.e., semantic and instance information are mapped independently.
The only link between both is different processing depending on the stuff and thing class assignment.
Especially for indoor environments, cluttered scenes may impede panoptic segmentation.
Directly mapping panoptic information is susceptible to misclassifying related semantic classes, e.g., classifying an armchair as chair and sofa or a cabinet as counter and shelf in subsequent images. 
We observe a great performance and robustness boost when decoupling semantic and instance information.
To derive the panoptic label for each voxel, we finally perform a panoptic-label-propagation step similar to Eq.~\ref{eq:panoptic_label_2d}. 
We introduce this step in Sec.~\ref{sec:main:mapping_panoptic_labels}.
Unlike~\cite{panoptic-fusion-2019-iros, panoptic-multi-tsdf-2022-icra}, we do not apply any subsequent map refinement, as this notably increases runtime. 

\subsection{Panoptic Mapping: Map and Data Representation}
\label{sec:main:mapping_data}
We build on top of the occupancy normal distribution transform (NDT) mapping approach presented in~\cite{Einhorn-ECMR-2013-NDT, Einhorn-PHD-2019-NDT} and its semantic extension~\cite{semanticmapping2022icra}.
Compared to traditional voxel maps~\cite{octomap-AR-2013}, 3D occupancy NDT maps do not only store an occupancy value per voxel but also shape information, describing the underlying surface in terms of a normal distribution.
These normal distributions can be updated efficiently in an incremental fashion~\cite{Einhorn-ECMR-2013-NDT}.
To integrate panoptic information, we extend the tuple of data fields held in each voxel~$v$ of the map to:
$$%
v = \langle v^{\text{Shape}}, v^{\text{Occ}}, v^L, v^Z, v^P \rangle
$$
The shape data field~$v^{\text{Shape}}$ stores a mean vector~$\boldsymbol{\mu}(v) \in \mathbb{R}^3$ and a covariance matrix~$\boldsymbol{\Sigma}(v) \in \mathbb{R}^{3{\times}3}$, describing the normal distribution of the surface. 
The occupancy data field~$v^{\text{Occ}}$ holds the occupancy~$o(v)$ as log odd.
As we focus on panoptic mapping, we refer to~\cite{Einhorn-ECMR-2013-NDT, Einhorn-PHD-2019-NDT} for the exact update process of these data fields.
The additional data fields~$v^L$, $v^Z$, store information for semantic and instances as histogram~$\boldsymbol{h}^L(v) \in \mathbb{R}^{|\mathcal{L}|}$ and $\boldsymbol{h}^Z(v) \in \mathbb{R}^{|\mathcal{Z}^{\text{3D}}|}$, respectively. 
Moreover, we track the total number of increments for both histograms, denoted as ${n}^L(v) \in \mathbb{N}$ and ${n}^Z(v) \in \mathbb{N}$.
The panoptic information~$P(v) \in \mathcal{P}^{\text{3D}}$ is stored in~$v^P$.

Note that we simplified the data representation for explanation purposes. 
In fact, for efficiency reasons, we also store the sum of both histograms and the proportion of stuff classes in the semantic histogram.
Moreover, the instance histogram is sparse, with a maximum of 16 entries holding only entries for globally unique instance IDs observed for that voxel.
We further keep this smaller histogram ordered and replace the smallest entry if the limit of 16 is reached.

The voxels are managed in an octree structure.
In the following, we denote the mapping from $\boldsymbol{u}$ to a voxel~$v$ as:
\begin{equation}
\label{eq:utov}
v = \mathrm{utov}(\boldsymbol{u}) := \mathrm{findnode}(\boldsymbol{T}\boldsymbol{K}^{-1}\left[\boldsymbol{u},1\right]^TD(\boldsymbol{u}))    
\end{equation}
with $\boldsymbol{T}$, $\boldsymbol{K}$ being the current extrinsic and intrinsic camera parameters. $D(\boldsymbol{u})$ is the current depth value along the ray, and $\mathrm{findnode}(\boldsymbol{\cdot})$ queries a voxel based on a point in 3D.
Given this forward mapping, we can derive a corresponding mapping $\mathrm{vtou}(v)$ that maps a voxel's information back to a set of coordinates $\mathcal{U}$ on the camera plane according to the NDT information of $v$. 
Note that the mappings are done once per frame.

\subsection{Panoptic Mapping: Instance Update}
\label{sec:main:mapping_instance}
We first perform the instance matching step to resolve the inconsistency in the instance IDs. 
Let $\mathcal{V}$ denote the set of voxels in the camera frustum affected by the forward mapping of the current observation (see Eq.\ref{eq:utov}); then, we first project each relevant instance $\tilde{z} \in \mathcal{Z}^{\text{3D}}$ back to the camera plane and create a mask $\boldsymbol{M}^{\tilde{z}}$ for each instance. 
The elements $M^{\tilde{z}}(\boldsymbol{u})$ are updated as follows:
\begin{equation}
    \begin{gathered}
        M^{\tilde{z}}(\boldsymbol{u}^\prime) = %
        \left\{%
        \begin{aligned}
            & 1 &\,& \mathrm{isthing}(v) \land \mathrm{isintopz}(v, \tilde{z}) \\
            & 0 &\,& \mathrm{otherwise}
        \end{aligned}
    \right.%
    \\%
    \boldsymbol{u^\prime} \in \mathrm{vtou}(v), \quad v \in \mathcal{V}, \quad \tilde{z} \in \mathcal{Z}^{\text{3D}}.
    \end{gathered}
\end{equation}
Here, $\mathrm{isthing}(v)$ ensures that voxels currently considered as stuff are ignored even if they contain any instance information. 
The function applies a threshold $\theta^{\text{St}}$ to the stuff proportion in the semantic histogram and is defined as:
\begin{equation}
    \label{eq:main:isthing}
    \begin{gathered}
        \mathrm{isthing}(v) := \frac{%
            \sum_{l \in \mathcal{L}^{\text{St}}} h^L_l(v)
        }{%
            \sum_{l \in \mathcal{L}} h^L_l(v)
        } < \theta^{\text{St}}.
    \end{gathered}
\end{equation}
Note that the instance information of previous observations is preserved and might be reactivated if the stuff/thing assignment of $v$ changes. 
Tuning this threshold further helps to ignore false positives due to an imprecise segmentation at object borders that blended into the background~(often stuff classes). 
The function $\mathrm{isintopz}(v,z)$ ensures that only the most relevant instances~(with the highest histogram scores) are back-projected. 
It is defined as:
\begin{equation}
    \begin{gathered}
        \mathrm{isintopz}(v, z) := \frac{%
            \sum_{i = 0}^{o_i = z} h^Z_{o_i}(v)
        }{%
            \sum_{i \in \mathcal{Z}^{\text{3D}}} h^Z_i(v)
        } \ge 1 - \theta^{\text{B}}
    \\[1mm]
    \text{with}\quad\boldsymbol{o} = \mathrm{argsort}^{\uparrow}(\boldsymbol{h}^Z(v)).
    \end{gathered}
\end{equation}

We choose $\theta^{\text{B}} = 0.8$ to project only \textasciitilde80\% of the instance knowledge back for matching.
This excludes only sporadically observed instances and, thus, speeds up matching. 

Given the back-projected instance masks, we then compute the intersection over union~(IoU)~--~also known as  Jaccard index~--~between all back-projected instances and the currently observed instances:
\begin{equation}
    J(z_a, z_b) = \mathrm{IoU}(\{\boldsymbol{u}^\prime | M^{z_a}(\boldsymbol{u}^\prime){=}1\}, \{\boldsymbol{u}^\prime | Z^P(\boldsymbol{u}^\prime){=}z_b\}).
\end{equation}
Finally, we derive the matched instance ID~$\hat{Z}(\boldsymbol{u})$ based on:
\begin{equation}
    \label{eq:main:matching}
    \hat{z} = \left\{%
        \begin{aligned}
            &\operatornamewithlimits{\mathrm{argmax}}_{\tilde{z} \in \mathcal{Z}^{\text{3D}}}(J(\tilde{z}, z)) &\,&\operatornamewithlimits{\mathrm{max}}_{\tilde{z} \in \mathcal{Z}^{\text{3D}}}(J(\tilde{z}, z)) > \theta^{\text{M}} \\
            &\mathrm{newz}(z) &\,& \operatornamewithlimits{\mathrm{max}}_{\tilde{z} \in \mathcal{Z}^{\text{3D}}}(J(\tilde{z}, z)) \le \theta^{\text{N}} \\
            &0 &\,& \hspace{1mm}\mathrm{otherwise}\\
        \end{aligned}
    \right.
\end{equation}
with $\hat{z} = \hat{Z}(\boldsymbol{u})$ and $z = Z^P(\boldsymbol{u})$.
The parameters~$\theta^{\text{M}}$ and $\theta^{\text{N}}$, with $\theta^{\text{M}} \ge \theta^{\text{N}}$, define a minimum threshold for accepting matches and an upper limit for creating new instances, respectively, and $\mathrm{newz}(z)$ defines a function that creates a new globally unique instance ID that is consistent for a given~$z$ within an update step.
Note that choosing $\theta^{\text{N}} < \theta^{\text{M}}$ restricts instance creation and further allows increasing $\theta^{\text{M}}$ in order to prevent under-segmentation.
Moreover, note that the instance matching is not exclusive, i.e., multiple predicted instances can match the same map instance. 
This compensates for temporary over-segmentation in the prediction.

Given the matched instances, the final instance update is straightforward.
If the panoptic score of a matched instance exceeds a minimum threshold of~$\theta^Z$, we update the map and increment the corresponding histogram bin in each voxel by the panoptic score as well as update the observation counter:
\begin{equation} 
    \label{eq:main:instance_update}
    \begin{gathered}
        h^Z_{\hat{Z}(\boldsymbol{u}^\prime)}(\mathrm{utov}(\boldsymbol{u}^\prime)) \mathrel{+}= S^P(\boldsymbol{u}^\prime)
        ,\;\;\,
        n^Z(\mathrm{utov}(\boldsymbol{u}^\prime)) \mathrel{+}= 1
        \\
        \boldsymbol{u}^\prime \in \left\{ \boldsymbol{u} | \hat{Z}(\boldsymbol{u}) \neq 0 \land S^P(\boldsymbol{u}) > \theta^Z \right\}.
    \end{gathered}
\end{equation}
We use the panoptic score instead of the instance score to account for a possible inconsistent semantic in the first stage.

\subsection{Panoptic Mapping: Semantic Update}
\label{sec:main:mapping_semantic}
The semantic update is performed subsequently to ensure consistent semantic information during back-projection.
As the semantic predictions are independent, the update step is much simpler.
If the semantic score of a panoptic label exceeds a minimum threshold of~$\theta^L$, we increment the corresponding histogram bin in each voxel by the semantic score and update the observation counter:
\begin{equation} 
    \label{eq:main:semantic_update}
    \begin{gathered}
        h^L_{L^P(\boldsymbol{u}^\prime)}(\mathrm{utov}(\boldsymbol{u}^\prime)) \mathrel{+}= S^{L^P}(\boldsymbol{u}^\prime)
        ,\;
        n^L(\mathrm{utov}(\boldsymbol{u}^\prime)) \mathrel{+}= 1%
        \\%
        \boldsymbol{u}^\prime \in \left\{ \boldsymbol{u} | S^{L^P}(\boldsymbol{u}) > \theta^L \right\}.
    \end{gathered}
\end{equation}

\subsection{Panoptic Mapping: Panoptic Label Propagation}
\label{sec:main:mapping_panoptic_labels}
To derive the panoptic label $P(v) \in \mathcal{P}^{\text{3D}}$ for each voxel, we follow a similar approach as for 2D:
\begin{equation}
\label{eq:panoptic_label_3d}
\hspace*{-2mm}
P(v) =%
\left\{%
    \begin{aligned}
        &\langle \underset{l \in \mathcal{L}^{\text{Th}}}{\mathrm{argmax}}(\underset{v^\prime \in \{i | \hat{Z}(i) = \hat{Z}(v)\}}{\mathrm{sum}(h_l^L(v^\prime)))}, \hat{Z}(v) \rangle &\,& \hspace{-2mm}\mathrm{pthing}(v) \\
        &\langle \hat{L}(v),  0 \rangle &\,&\hspace{-2mm}\mathrm{otherw.}
    \end{aligned}
\right.%
\end{equation}
with $\hat{L}(v) = \mathrm{argmax}_{l \in \mathcal{L}}(h_l^{L}(v))$ and $\hat{Z}(v) = \mathrm{argmax}_{i \in \mathcal{Z^{\text{3D}}}}(h_i^{Z}(v))$.
Here, $\mathrm{pthing}(v)$ determines whether to propagate thing information. 
It extends the previous $\mathrm{isthing}(v)$ and further incorporates the \mbox{observation counters:}
\begin{equation}
    \label{eq:main:propthing}
    \mathrm{pthing}(v) := \mathrm{isthing}(v) \land \frac{n^Z(v)}{n^L(v)} \ge \theta^{O}
\end{equation}
An observation ratio less than the threshold $\theta^O$ marks a voxel to currently contain only weak instance knowledge, i.e., even if it belongs to thing, only a fraction of $\theta^O$ of the observations entered the map as valid instance observations.
Note, due to Eq.~\ref{eq:panoptic_label_3d}, this may introduce a garbage segment for thing classes. 
However, it prevents over-segmentation and is of great importance for real-world application, as it indicates thing areas that could not be separated into instances.
\section{Experiments}
\label{sec:experiments}
We evaluate the performance of our proposed panoptic mapping approach on the publicly available datasets Hypersim~\cite{hypersim-iccv2021} and ScanNetV2~\cite{scannet-cvpr2017} in 3D as well as in 2D.
In the following, we first introduce both datasets and describe the exact evaluation protocol.
Subsequently, we provide details about the training of EMSANet and give further implementation details.
Finally, we present quantitative and qualitative results demonstrating the performance of PanopticNDT.

\begin{table*}[!b]
\vspace{-5mm}
\centering
\caption{%
    Results on Hypersim validation and test split when mapping with ground truth~(top) and predicted segmentation~(bottom) of EMSANet~\cite{emsanet2022ijcnn}~(center). %
    See Sec.~\ref{sec:experiments:eval} for details on the reported metrics. %
    Legend: %
        gray: mapping with semantic only instead of panoptic information (black), %
        *: re-application in our data pipeline, %
        ${}^{\dagger}$: mapping with inputs at 640${\times}$480 instead of 1024${\times}$768 pixels, %
        $x$\si{\centi\meter}: voxel size, %
        w/o NDT: PanopticNDT but evaluated without shape information, %
        and FPS:~average update rate in frames per second on the hardware of our mobile robot, i.e., for single-threaded mapping: Intel NUC11PHKi7C~(Intel i7-1165G7) and for EMSANet: NVIDIA Jetson AGX Orin~(Jetpack 5.0.2, TensorRT 8.4, Float16, 30W$\,$/$\,$50W). %
}
\vspace{-2.5mm}
\resizebox{\textwidth}{!}{%
\begin{tabular}{%
@{\hspace{1mm}}l%
@{\hspace{2mm}}l%
@{\hspace{6mm}}%
c@{\hspace{2mm}}c@{\hspace{1mm}}c%
@{\hspace{6mm}}%
c@{\hspace{2mm}}c@{\hspace{1mm}}c%
@{\hspace{10mm}}%
c@{\hspace{2mm}}c@{\hspace{1mm}}c%
@{\hspace{6mm}}%
c@{\hspace{2mm}}c@{\hspace{1mm}}c%
@{\hspace{10mm}}%
c@{\hspace{1mm}}}
\toprule
&                     & \multicolumn{6}{c@{\hspace{13mm}}}{\bf Valid}                           & \multicolumn{6}{c@{\hspace{13mm}}}{\bf Test}                               &      \\
&                     & \multicolumn{3}{c@{\hspace{4mm}}}{\bf 3D}   & \multicolumn{3}{c@{\hspace{8mm}}}{\bf 2D}   & \multicolumn{3}{c@{\hspace{4mm}}}{\bf 3D}      & \multicolumn{3}{c@{\hspace{8mm}}}{\bf 2D}   &      \\                     
&                     & \bf mIoU$\,{}^\uparrow$     & \bf mIoU${}_\text{P}^{\,\uparrow}$ & \bf PQ$\,{}^\uparrow$    & \bf mIoU$\,{}^\uparrow$     & \bf mIoU${}_\text{P}^{\,\uparrow}$ & \bf PQ$\,{}^\uparrow$    & \bf mIoU$\,{}^\uparrow$        & \bf mIoU${}_\text{P}^{\,\uparrow}$ & \bf PQ$\,{}^\uparrow$    & \bf mIoU$\,{}^\uparrow$    & \bf mIoU${}_\text{P}^{\,\uparrow}$ & \bf PQ$\,{}^\uparrow$     & \bf FPS$\,{}^\uparrow$  \\
\midrule
\multirow{7}{*}{\rotatebox{90}{\bf~Ground Truth}}%
& PanopticNDT (5\si{\centi\meter})%
    & 91.84 & 91.36 & 76.04    & 84.46 & 84.18 & 65.78    & 87.22 & 86.74 & 73.30    & 84.68 & 84.34 & 67.67    & 1.10 \\
& PanopticNDT (10\si{\centi\meter})%
    & 90.31 & 89.04 & 73.24    & 76.84 & 75.87 & 53.40    & 87.01 & 86.24 & 69.39    & 77.54 & 77.02 & 57.03    & 2.73 \\
& PanopticNDT (20\si{\centi\meter})%
    & 83.30 & 81.03 & 61.52    & 64.71 & 63.37 & 37.57    & 82.07 & 80.29 & 59.86    & 64.46 & 63.35 & 40.77    & 3.35 \\
\cmidrule{2-15}
& Panoptic Multi-TSDFs~\cite{panoptic-multi-tsdf-2022-icra}*%
    & ---   & 70.82 & 51.00    & ---   & ---   & ---      & ---   & 69.69 & 54.69    & ---   &  ---  &  ---    & 2.95${}^{\dagger}$ \\
& PanopticNDT (5\si{\centi\meter}, w/o NDT)%
    & 91.33 & 90.94 & 75.55    & 70.37 & 70.60 & 48.47    & 86.84 & 86.40 & 72.80    & 71.85 & 71.87 & 52.17   & --- \\
& SemanticNDT~\cite{semanticmapping2022icra}* (10\si{\centi\meter}) 
    & \color[HTML]{737373}90.31 & --- &  ---      & \color[HTML]{737373}76.84 & --- & ---      & \color[HTML]{737373}87.01 & --- & ---      & \color[HTML]{737373}77.54 & --- & ---      & 4.32 \\
\midrule
\midrule
& EMSANet%
    & --- & --- & ---    & \color[HTML]{737373}49.74 & 49.12 & 34.95    & --- & --- & ---    & \color[HTML]{737373}46.66 & 44.66 & 29.77    & 24.0$\,$/$\,$35.5    \\
\midrule
\multirow{6}{*}{\rotatebox{90}{\bf~Prediction}}%
& PanopticNDT (5\si{\centi\meter})%
    & 45.53 & 45.10 & 30.37    & 52.50 & 53.50 & 34.23    & 44.86 & 45.09 & 26.99    & 48.82 & 50.27 & 32.36    & 1.11 \\
& PanopticNDT (10\si{\centi\meter})%
    & 45.43 & 44.56 & 31.08    & 48.93 & 49.00 & 30.44    & 45.34 & 45.20 & 27.54    & 46.10 & 47.01 & 28.82    & 2.73 \\
& PanopticNDT (20\si{\centi\meter})%
    & 43.60 & 42.14 & 28.84    & 42.94 & 42.53 & 22.70    & 44.29 & 44.08 & 25.98    & 40.24 & 40.47 & 22.90    & 3.41 \\
\cmidrule{2-15}
& Panoptic Multi-TSDFs~\cite{panoptic-multi-tsdf-2022-icra}* %
    & --- & 30.85 & 15.92    & --- & --- & ---    & --- & 30.59 & 15.98    & --- & --- & ---    & 1.25${}^{\dagger}$ \\    
& SemanticNDT~\cite{semanticmapping2022icra}* (10\si{\centi\meter}) 
    & \color[HTML]{737373}44.31 & --- & ---    & \color[HTML]{737373}48.45 & --- &  ---     & \color[HTML]{737373}44.80 & --- & ---    & \color[HTML]{737373}45.56 & --- & ---    & 4.32 \\

\bottomrule
\end{tabular}
\label{tab:results_hypersim}
}
\end{table*}

\subsection{Datasets}
\label{sec:experiments:data}
Both Hypersim and ScanNetV2 feature images for RGB and depth, dense panoptic annotations as well as the corresponding intrinsic and extrinsic camera parameters. 
Both datasets comprise a training, validation, and test split.

\textit{Hypersim~\cite{hypersim-iccv2021}:} \enspace %
Hypersim is a synthetic dataset with 77,400 photorealistic images of 461 indoor scenes rendered in 774 camera trajectories.
Each trajectory is generated by a random walk and consists of 100 camera poses, ensuring reasonable visual coverage.
The rendered images have a spatial resolution of 1024${\times}$768 pixels.
However, some scenes were rendered using tilt-shift photography, which is incompatible with the simpler pinhole camera model used in robotic frameworks such as MIRA and ROS.
This becomes a problem when projecting depth images into 3D space for mapping.
Unfortunately, re-rendering the images is difficult as it requires buying all underlying assets. 
Therefore, we projected all annotations into 3D space using the provided camera parameters and then back to the simpler pinhole camera.
This inevitably leads to an information loss of \textasciitilde5.3\% as it is impossible to back-project all pixels without introducing ambiguities.
However, this approach is still better than excluding scenes.
We follow~\cite{semanticmapping2022icra} to further filter invalid trajectories~(void/single semantic class only, missing textures, invalid depth) from training and validation split.

\textit{ScanNetV2~\cite{scannet-cvpr2017}:} \enspace %
ScanNetV2, on the other hand, is a real-world dataset for indoor scene understanding.
It comprises 2.5M images of 807 indoor scenes captured in 1613 camera trajectories.
Depth and RGB were captured at a spatial resolution of 640${\times}$480 and 1280${\times}$960 pixels, respectively.
As the dataset is created from video sequences, it contains many similar images. 
We use a subsampling of 5 in our pipeline to reduce the number of samples.

Both datasets provide annotations for the 40 semantic classes of NYUv2~\cite{NYUv2-eccv2012}. 
Following~\cite{panoptic-fusion-2019-iros, emsanet2022ijcnn}, we treat wall, floor, and ceiling to be background~(stuff classes).
The ScanNetV2 benchmark further excludes 20 classes due to few annotations, which we do as well but only for ScanNetV2.
We use the training split for network training and the validation split for tuning hyperparameters in our pipeline. 
The test split is only used for reporting results.
For further details on data processing, we refer to our GitHub repository.

\subsection{Evaluation Protocol}
\label{sec:experiments:eval}
We build upon the publicly available ScanNetV2 benchmark pipeline~\cite{scannet-cvpr2017} and evaluate our approach in 3D and 2D.
Unfortunately, this pipeline does not feature a panoptic evaluation.
Therefore, we extend the existing pipeline with a complementary panoptic evaluation task. 
The evaluation code is shared at GitHub.
The mean intersection over union~(mIoU) is reported for evaluating semantic information and the panoptic quality~(PQ)~\cite{Panopticsegmentation-cvpr2019} as well as the mIoU$_P$ for evaluating panoptic information and the semantic of panoptic information, respectively.
As it is part of ScanNetV2 instance evaluation on the hidden test split, we also report the average precision at 50\% overlap~(AP${}_{\text{50}}$) for evaluating instances.
Note, according to~\cite{Panopticsegmentation-cvpr2019}, PQ and AP track closely and an improvement in AP is expected to also improve PQ.

\textit{3D Evaluation:} \enspace %
In the ScanNetV2 benchmark, evaluation in 3D is done  by mapping created representations to annotated ground-truth point clouds.
While ScanNetV2 provides annotated meshes as ground truth, such ground-truth representations are missing for Hypersim.
Therefore, we generated them ourselves by applying a voxel-grid filter with a voxel size of 1\si{\centi\meter} to the point cloud of each camera trajectory.
The most frequent annotation is used to assign a ground-truth annotation to each voxel.
For mapping points to our NDT map, the na\"ive way would be to assign a ground-truth point by simply treating the NDT map as a voxel map, i.e., by querying the voxel into which the point would fall.
However, as we have shape information in terms of an estimated normal distribution~(see Sec.~\ref{sec:main:mapping_data}), we can make a more informed matching.
We compute the Mahalanobis distances between each ground-truth point and the normal distributions of the corresponding NDT voxel and its immediate neighborhood and assign the annotation based on the smallest distance.
Every ground-truth point that cannot be matched is labeled as \emph{unknown}~(void).

\textit{2D Evaluation:} \enspace %
Although it is natural to evaluate in 3D, there is a major drawback. 
As created representations are mapped to ground-truth representations, the actual spatial resolution of a created representation has less influence, i.e., modeling free space is less relevant.
Therefore, we follow~\cite{semantic-3d-occ-IROS-2017, bkgi-octomap-ICRA-2017, semanticmapping2022icra} and further evaluate back-projections of the map to the camera plane~(see Sec.~\ref{sec:main:mapping_instance}).
This evaluation also allows for a comparison to the raw network predictions and, thereby, assessing the temporal integration is possible.

\subsection{Network Training \& Implementation Details}
\label{sec:experiments:implementation}
We use the PyTorch implementation of EMSANet for network training~\cite{emsanet2022ijcnn}. 
The network input resolution is chosen based on the resolution of the RGB images. %
Network training was done for 500 epochs. %
For optimization, we used SGD with momentum and performed a grid search for determining a suitable learning rate.
For each dataset, we derived the best configuration based on PQ on the corresponding validation split.
We refer to the implementation at GitHub for the exact hyperparameters and the network weights.
The results of EMSANet as well as the inference throughput is reported next to the mapping results in Tab.~\ref{tab:results_hypersim} and Tab.~\ref{tab:results_scannet}.

Mapping is done at the resolution of provided depth images, i.e., at 1024${\times}$768 for Hypersim and at 640${\times}$480 for ScanNetV2.
We further follow~\cite{semantic-3d-occ-IROS-2017, semantic-bsk-map-RAL-2020, semanticmapping2022icra} and limit the maximum mapping distance for Hypersim to a reasonable value of 20\si{\meter}.
Independent of the dataset, we choose %
the stuff proportion threshold~$\theta^{\text{St}}$~(see Eq.~\ref{eq:main:isthing}) to 0.9; %
the thresholds for matching instances~$\theta^{\text{M}}$ and creating new instances~$\theta^{\text{N}}$~(see Eq.~\ref{eq:main:matching}) to 0.2 and 0.1;
and the instance to semantic observation ratio threshold~$\theta^{O}$ (see Eq.~\ref{eq:main:propthing}) to 0.25.
The score thresholds~$\theta^{L}$ and~$\theta^{Z}$~(see Eq.~\ref{eq:main:instance_update} and Eq.~\ref{eq:main:semantic_update}) are set to 0.7 and 0.4 for Hypersim; and to 0.7 and 0.1 for ScanNetV2.

\begin{table*}[!b]
\vspace{-5mm}
\centering
\caption{%
    Results on ScanNetV2 validation and hidden test split when mapping with predicted segmentation of EMSANet~\cite{emsanet2022ijcnn}. %
    See Sec.~\ref{sec:experiments:eval} for details on the reported metrics. %
    Legend: %
        gray: mapping with semantic only instead of panoptic information (black), %
        *: re-application in our data pipeline, %
        $x$\si{\centi\meter}: voxel size, %
        ${\dagger}$: expected to be much slower~(see Sec.~\ref{sec:related_work:panoptic_mapping}), %
        and FPS:~average update rate in frames per second on %
        our mobile robot, i.e., for single-threaded mapping: Intel NUC11PHKi7C (Intel i7-1165G7) and for EMSANet: NVIDIA Jetson AGX Orin (Jetpack 5.0.2, TensorRT 8.4, Float16, 30W$\,$/$\,$50W). %
}
\vspace{-2.5mm}
\resizebox{0.98\textwidth}{!}{%
\begin{tabular}{%
@{\hspace{2mm}}l%
@{\hspace{6mm}}%
c@{\hspace{2mm}}c@{\hspace{1mm}}c@{\hspace{2.5mm}}c%
@{\hspace{6mm}}%
c@{\hspace{2mm}}c@{\hspace{1mm}}c@{\hspace{2.5mm}}c%
@{\hspace{10mm}}%
c@{\hspace{1mm}}c%
@{\hspace{6mm}}%
c@{\hspace{1mm}}c%
@{\hspace{10mm}}%
c@{\hspace{1mm}}}
\toprule
                     & \multicolumn{8}{c@{\hspace{13mm}}}{\bf Valid}                           & \multicolumn{4}{c@{\hspace{13mm}}}{\bf Test}                               &      \\
                     & \multicolumn{4}{c@{\hspace{4.5mm}}}{\bf 3D}   & \multicolumn{4}{c@{\hspace{8mm}}}{\bf 2D}   & \multicolumn{2}{c@{\hspace{6mm}}}{\bf 3D}      & \multicolumn{2}{c@{\hspace{10mm}}}{\bf 2D}   &      \\                     
                     & \bf mIoU$\,{}^\uparrow$     & \bf mIoU${}_\text{P}^{\,\uparrow}$ & \bf PQ$\,{}^\uparrow$ & \bf AP${}_{50}^{\,\uparrow}$   & \bf mIoU$\,{}^\uparrow$     & \bf mIoU${}_\text{P}^{\,\uparrow}$ & \bf PQ$\,{}^\uparrow$ & \bf AP${}_{50}^{\,\uparrow}$    & \bf mIoU${}_\text{P}^{\,\uparrow}$ & \bf AP${}_{50}^{\,\uparrow}$      & \bf mIoU${}_\text{P}^{\,\uparrow}$ & \bf AP${}_{50}^{\,\uparrow}$     & \bf FPS$\,{}^\uparrow$  \\
\midrule
EMSANet%
    & --- & --- & --- & ---    & \color[HTML]{737373}70.99 & 66.19 & 58.22 & 41.94    & --- & ---    & 60.0 & 38.0    & 15.4$\,$/$\,$23.1     \\
\midrule
PanopticNDT (5\si{\centi\meter})   & 70.07 & 69.37 & 57.38 & 52.79    & 68.31 & 67.71 & 52.48 & 39.18    & ---   & ---     & ---  & ---     & 1.25  \\
PanopticNDT (10\si{\centi\meter})  & 69.47 & 68.39 & 59.19 & 52.65    & 67.70 & 66.17 & 50.48 & 37.76    & 68.1  & 50.9    & 64.8 & 39.8    & 7.77  \\
\midrule
Panoptic Multi-TSDFs~\cite{panoptic-multi-tsdf-2022-icra}* %
    & --- & 47.29 & 34.75 & ---    & --- & --- & --- & ---    & --- & ---    & --- & ---    & 5.85     \\
Panoptic Fusion~\cite{panoptic-fusion-2019-iros} %
    & --- & --- & 33.5 & ---    & --- & --- & --- & ---    & 52.9 & 47.8     & --- & ---    & ${\dagger}$   \\
SemanticNDT~\cite{semanticmapping2022icra}* (10\si{\centi\meter})%
 & \color[HTML]{737373}69.59 & --- & --- & ---     & \color[HTML]{737373}67.71 & --- & --- & ---    & --- & ---    & --- & ---    & 8.73  \\
\bottomrule
\end{tabular}
\label{tab:results_scannet}
}
\end{table*}

\subsection{Results on Hypersim}
\label{sec:experiments:results_hypersim}
First, we evaluate our panoptic mapping approach on Hypersim in a setting assuming perfect data, i.e., using the provided ground-truth panoptic segmentation as well as perfect depth data for mapping.
The goal is to determine how well panoptic information can be represented in the resulting maps.
Tab.~\ref{tab:results_hypersim}~(upper part) lists the results and further compares them to Panoptic Multi-TSDFs~\cite{panoptic-multi-tsdf-2022-icra}, a voxel-based version of our approach (ignoring NDT shape information), and the semantic NDT approach of~\cite{semanticmapping2022icra} we build upon.

For both splits similar trends emerge.
First, in 3D, PanopticNDT performs better with decreasing voxel size, while the step from 10\si{\centi\meter} to 5\si{\centi\meter} is already much smaller than the step from 20\si{\centi\meter} to 10\si{\centi\meter}.
A voxel size of 5\si{\centi\meter} may already be close to the upper limit of our approach for representing panoptic information in 3D.
All settings perform significantly better than Panoptic Multi-TSDFs.
In 2D, i.e., when accounting for modeling free space accurately, the steps between the voxel sizes are larger for all metrics, however, again, the steps get smaller.
Nevertheless, the results indicate that Hypersim contains a lot of small objects that require accurate modeling.
The comparison to the simple voxel-based version of our approach (denoted as w/o NDT) shows that the NDT representation enables choosing the next coarser voxel size while still achieving at least the same precision. 

\begin{figure}[b!]
    \centering
    \vspace{-4mm}
    \includegraphics[width=\linewidth]{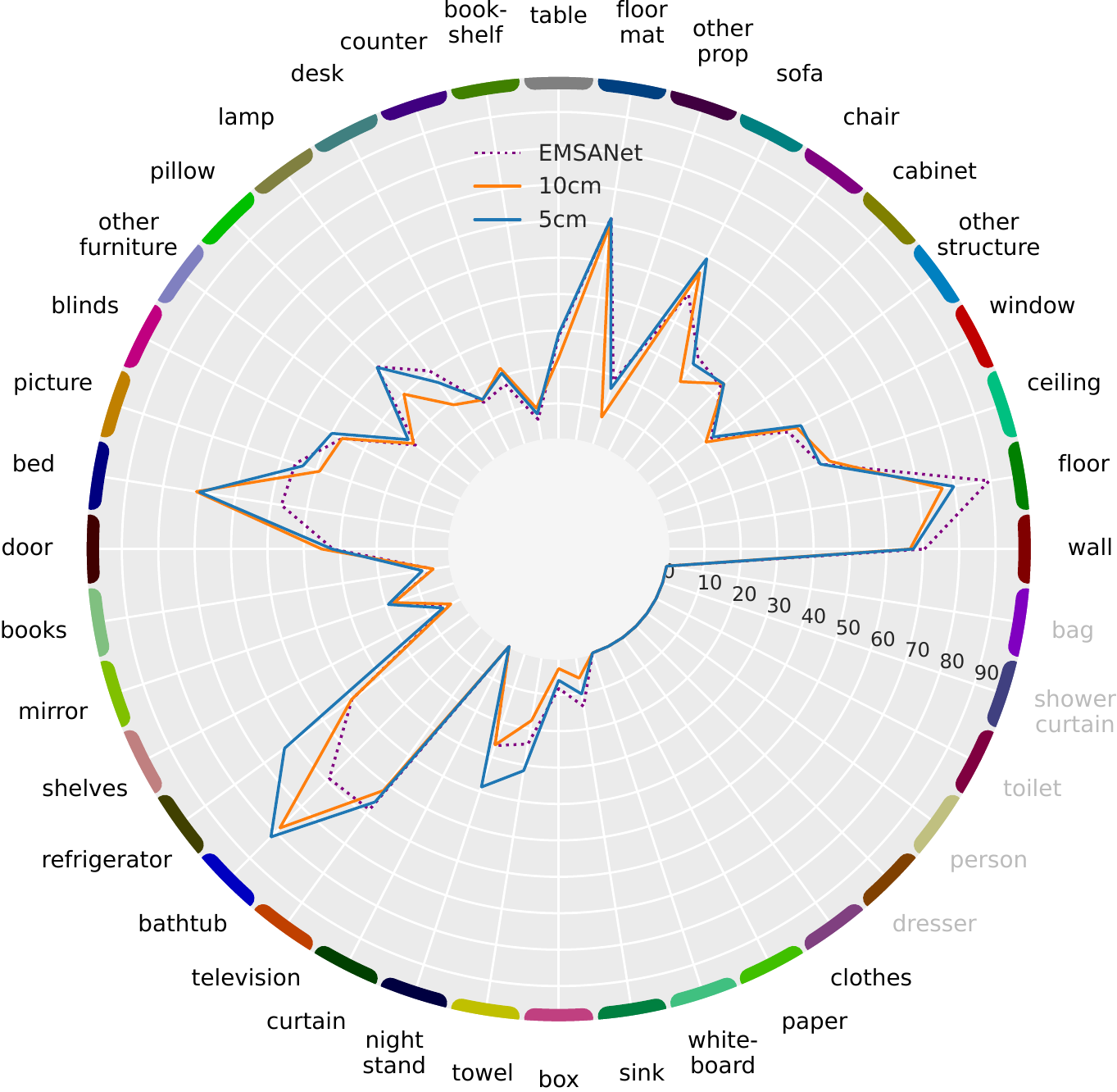}
    \vspace{-7mm}
    \caption{
        Per-class 2D panoptic quality on the Hypersim test split for EMSANet and when mapping with its predictions and voxel sizes of 10\si{\centi\meter} and 5\si{\centi\meter}. %
        Classes printed in gray do not appear in the test split.
    }
    \label{fig:experiments:hypersim_radar_chart}
    \vspace{-0.5mm}
\end{figure}

Next, we use the outputs of EMSANet (see Tab.~\ref{tab:results_hypersim}~(center) for results) as input for panoptic mapping.
The results of the 3D evaluation in Tab.~\ref{tab:results_hypersim} (lower part) show that PanopticNDT performs similarly regardless of the used voxel size. 
This indicates that the performance of our approach is mainly limited by the quality of the preceding panoptic segmentation.
However, the results of the 2D evaluation suggest that our mapping approach is able to improve segmentation by integrating observations over time.
For the test split, merging semantic and instance predictions of EMSANet to derive panoptic segmentations (input for panoptic mappings) leads to a drop in mIoU from 46.66\% to 44.66\%. 
Integrating these merged panoptic observations over time in a panoptic NDT map with a voxel size of 10\si{\centi\meter} already almost closes this gap~(mIoU: 46.10\%) even if there is some unavoidable loss due to the much coarser map representation~(see results with ground truth).
Further combining the aggregated semantic and instance knowledge in the map by propagating panoptic labels additionally improves results~(mIoU: 47.01\%). 
For the smaller voxel size of 5\si{\centi\meter}, this effect is even stronger and also affects PQ.
Fig.~\ref{fig:experiments:hypersim_radar_chart} depicts the per-class PQs and compares to the EMSANet result. 
It becomes obvious that the increase in PQ comes from improvements across most thing classes.
Only for the stuff class floor is some larger decrease.
Fig.~\ref{fig:experiments:qualitative_results}~(top) further visualizes qualitative results.

The performance of Panoptic Multi-TSDFs~\cite{panoptic-multi-tsdf-2022-icra} notably decreases when mapping with EMSANet outputs.
We observed that this drop is mainly caused by mapping issues for smaller objects in thing classes.
We assume that the submap approach is not able to account for such noisy observations.

\subsection{Results on ScanNetV2}
\label{sec:experiments:results_scannet}
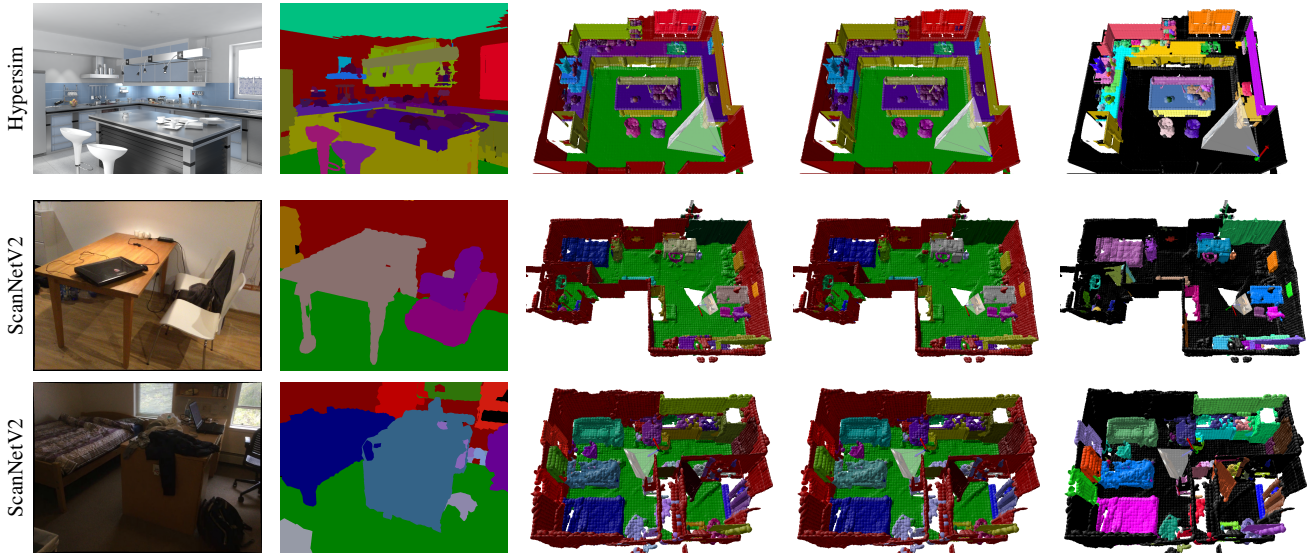
\begin{figure*}[!t]
   \centering 
   \resizebox{0.986\textwidth}{!}{%
   \begin{tikzpicture}
       \node (hypersim_color) {\includegraphics[width=3cm, height=2.25cm]{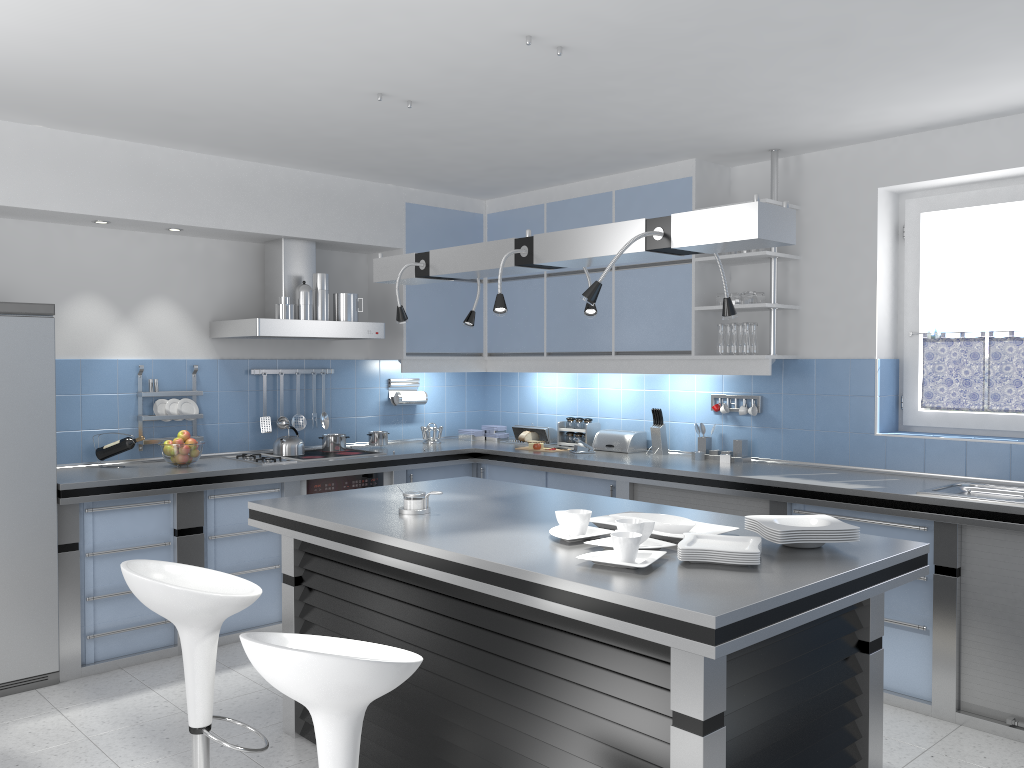}};%
       \node[right=0 of hypersim_color] (hypersim_backproj) {\includegraphics[width=3cm, height=2.25cm]{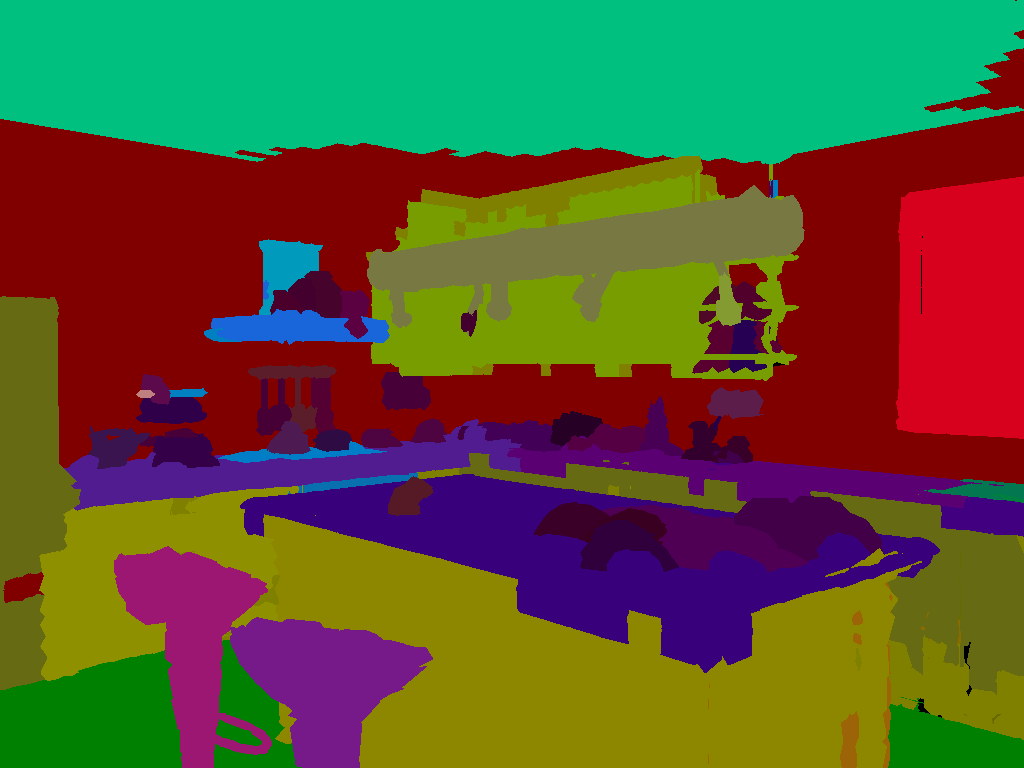}};%
       \node[right=0 of hypersim_backproj] (hypersim_panoptic){\includegraphics[height=2.25cm, trim={0 2cm 0 1.7cm}, clip]{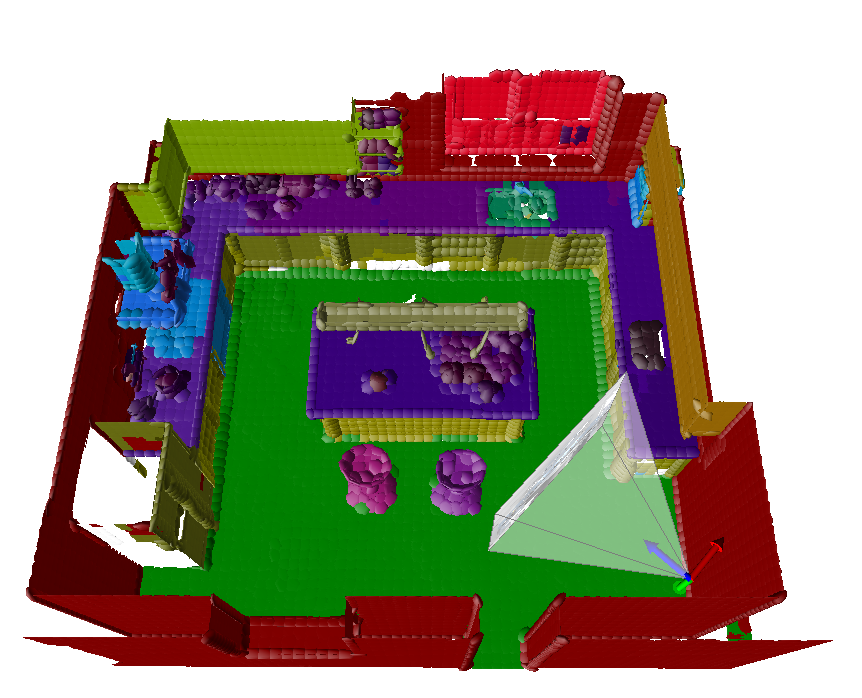}};%
       \node[right=0 of hypersim_panoptic] (hypersim_semantic){\includegraphics[height=2.25cm, trim={0 2cm 0 1.7cm}, clip]{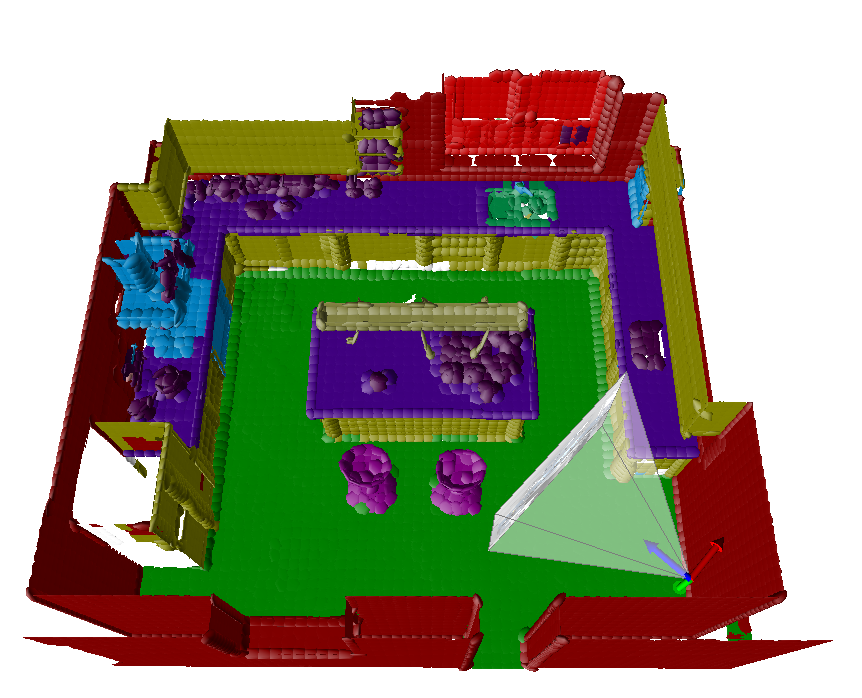}};%
       \node[right=0 of hypersim_semantic] (hypersim_instance) {\includegraphics[height=2.25cm, trim={0 2cm 0 1.7cm}, clip]{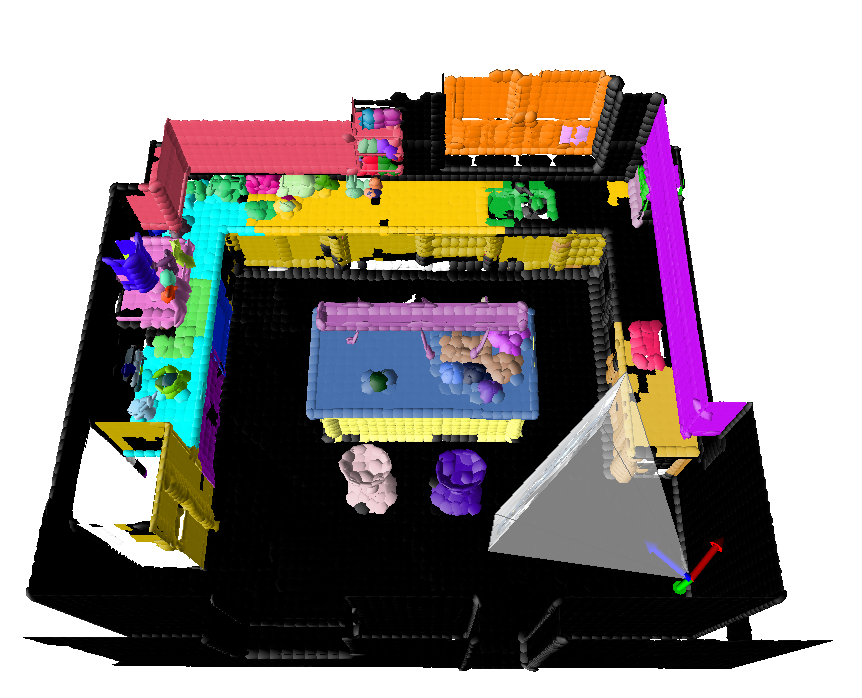}};%
       
       \node[below=0.1 of hypersim_color](scannet_color_1) {\includegraphics[width=3cm, height=2.25cm]{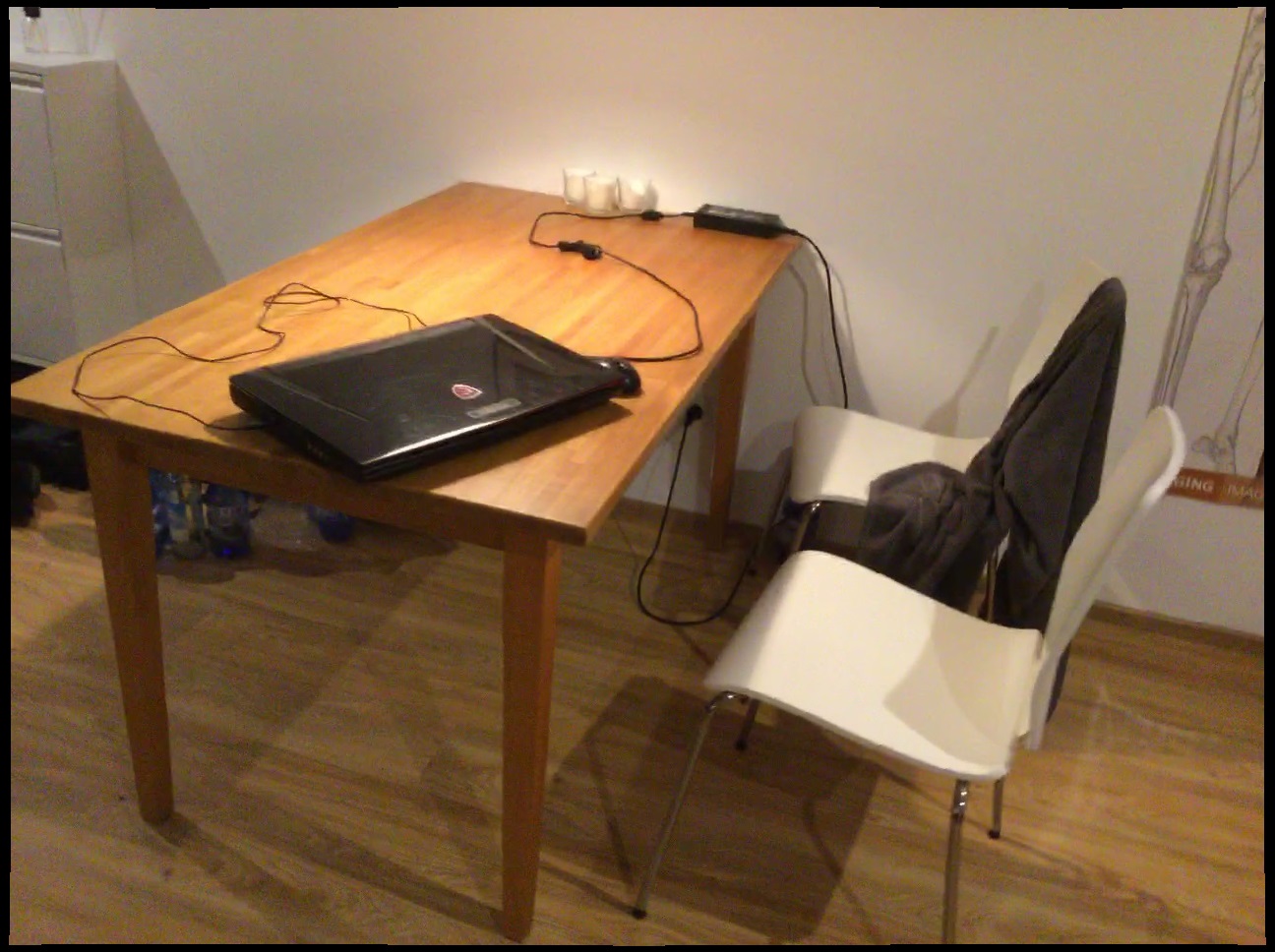}};%
       \node[right=0 of scannet_color_1] (scannet_backproj_1) {\includegraphics[width=3cm, height=2.25cm]{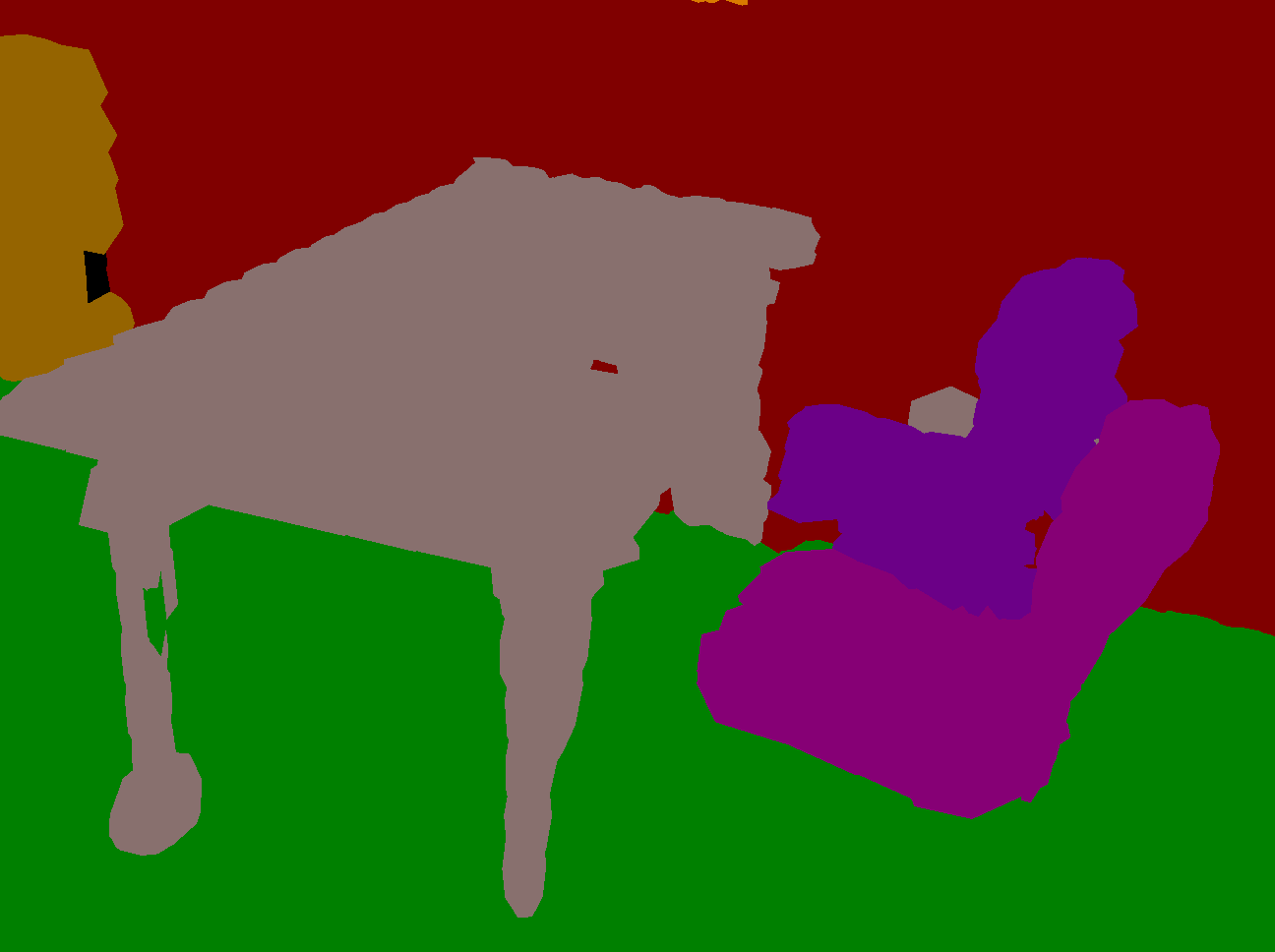}};%
       \node[right=0 of scannet_backproj_1] (scannet_panoptic){\includegraphics[height=2.25cm, trim={0 0.5cm 0 3.2cm}, clip]{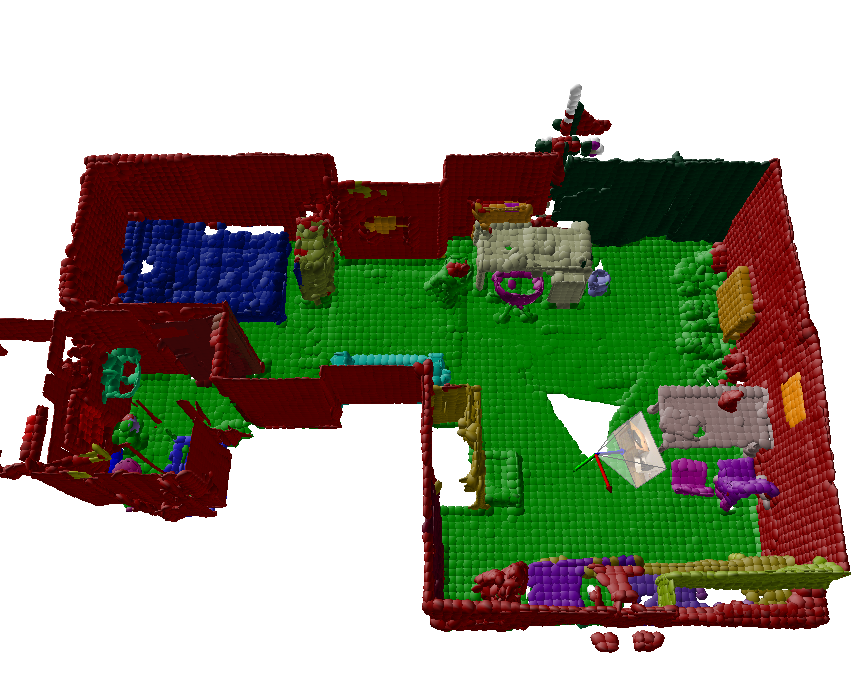}};%
       \node[right=0 of scannet_panoptic] (scannet_semantic){\includegraphics[height=2.25cm, trim={0 0.5cm 0 3.2cm}, clip]{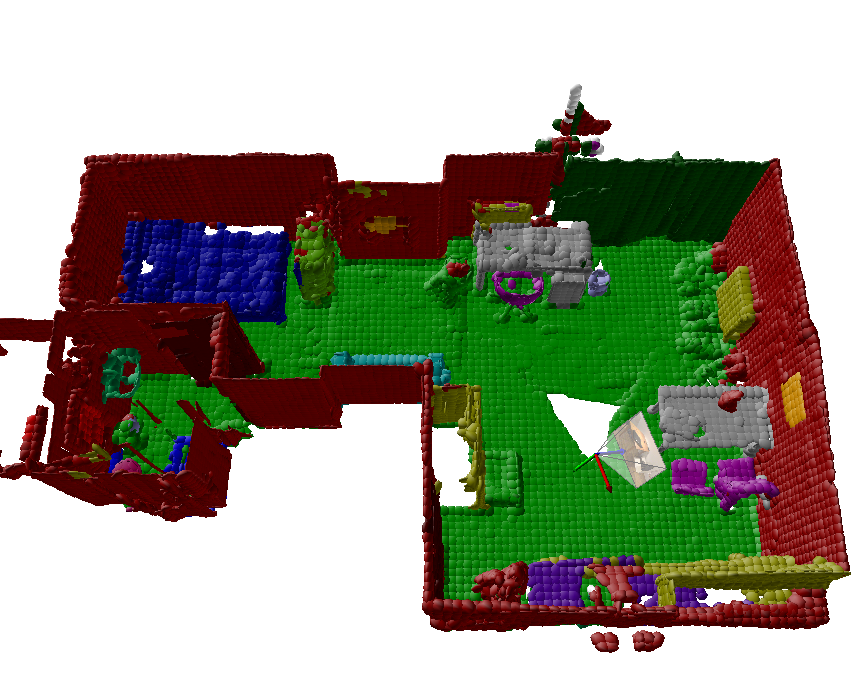}};%
       \node[right=0 of scannet_semantic] (scannet_instance) {\includegraphics[height=2.25cm, trim={0 0.5cm 0 3.2cm}, clip]{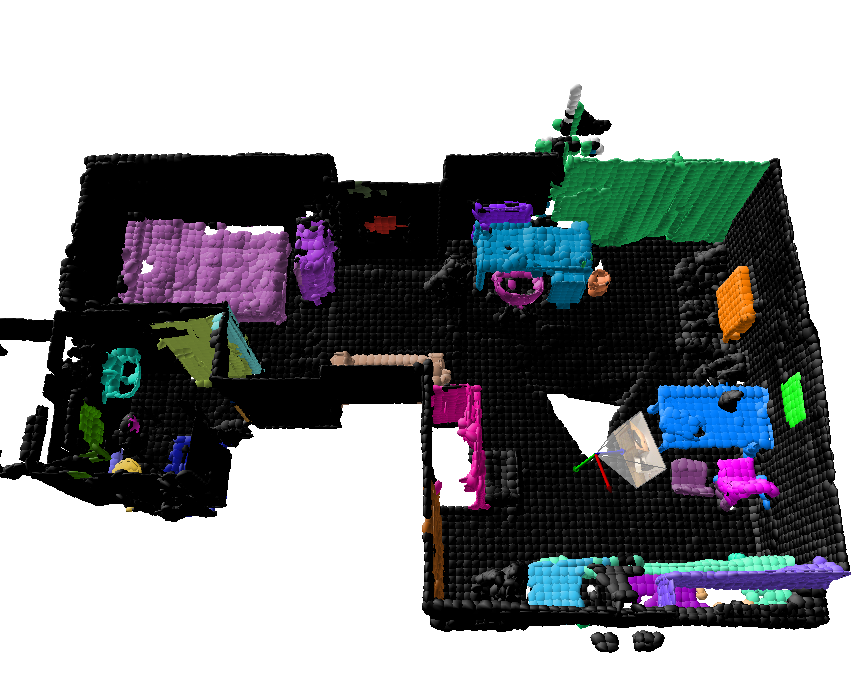}};%

       \node[below=-0.1 of scannet_color_1] (scannet_color_2) {\includegraphics[width=3cm, height=2.25cm]{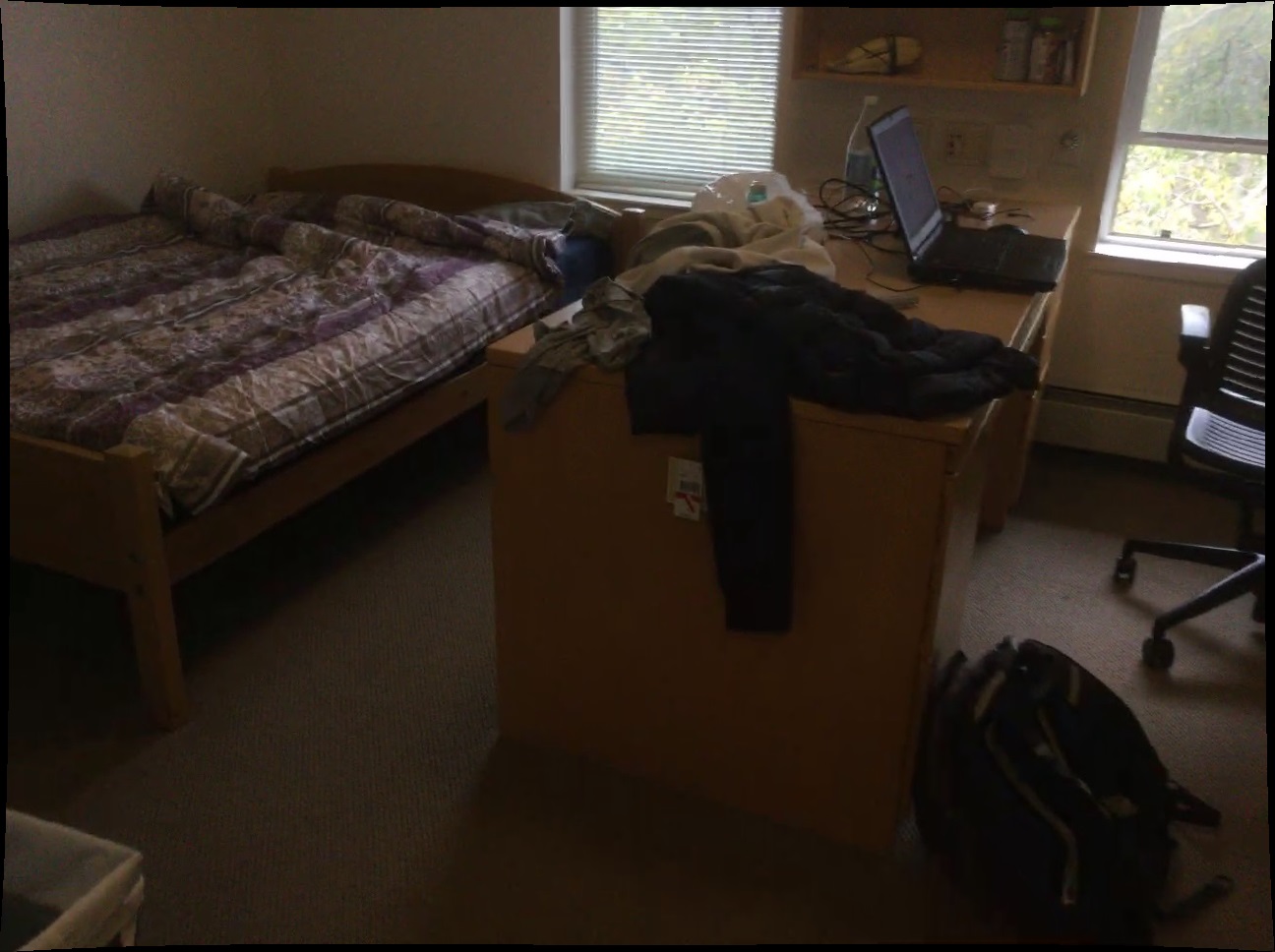}};%
       \node[right=0 of scannet_color_2] (scannet_backproj_2) {\includegraphics[width=3cm, height=2.25cm]{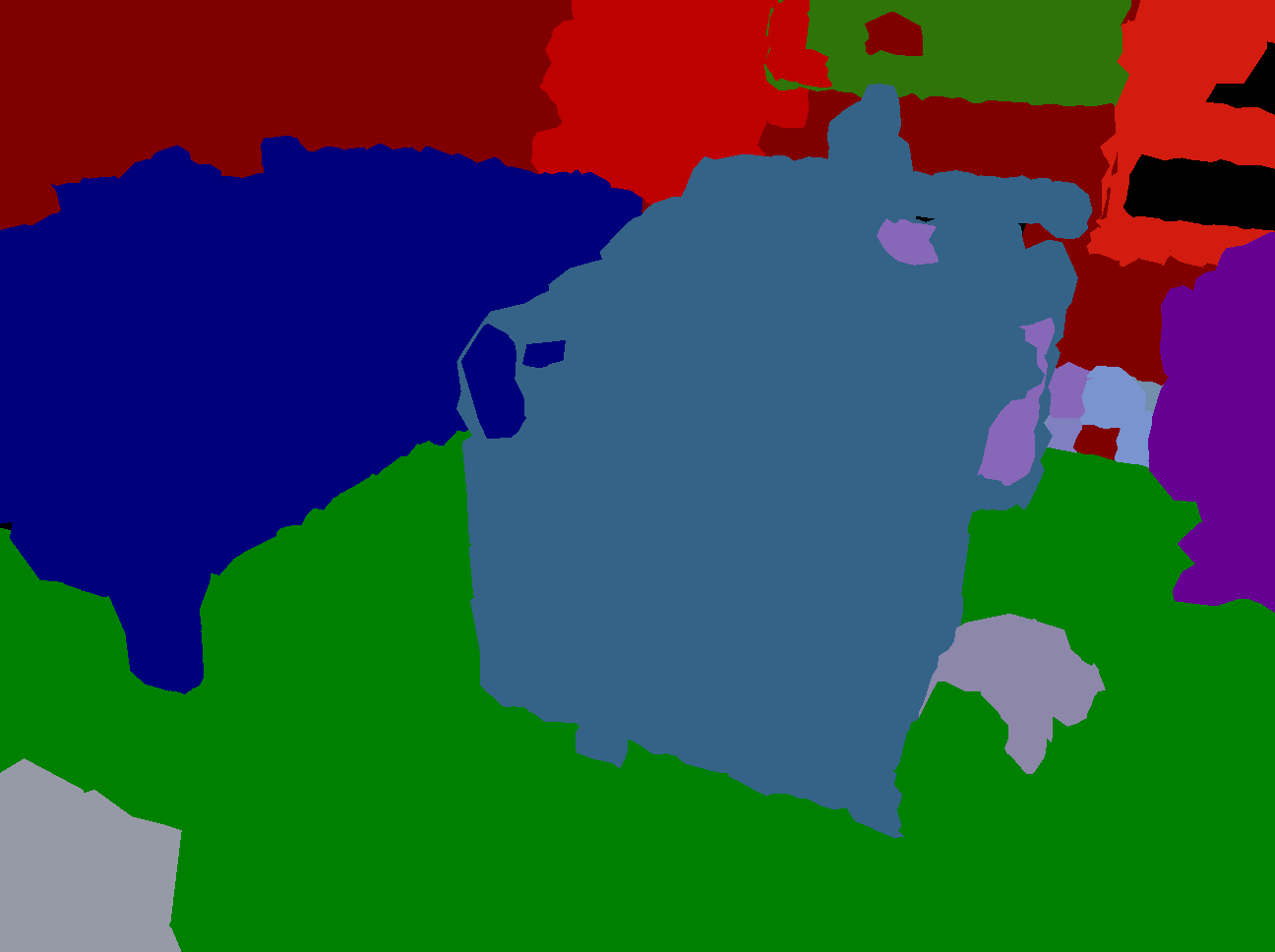}};%
       \node[right=0 of scannet_backproj_2] (scannet_panoptic_test){\includegraphics[height=2.25cm, trim={0 0.5cm 0 3.2cm}, clip]{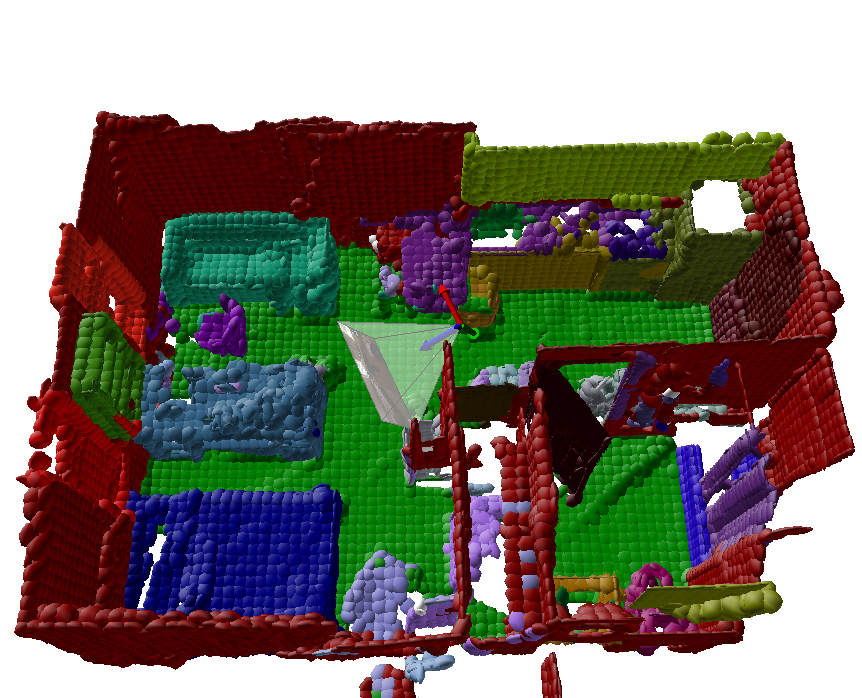}};%
       \node[right=0 of scannet_panoptic_test] (scannet_semantic_test){\includegraphics[height=2.25cm, trim={0 0.5cm 0 3.2cm}, clip]{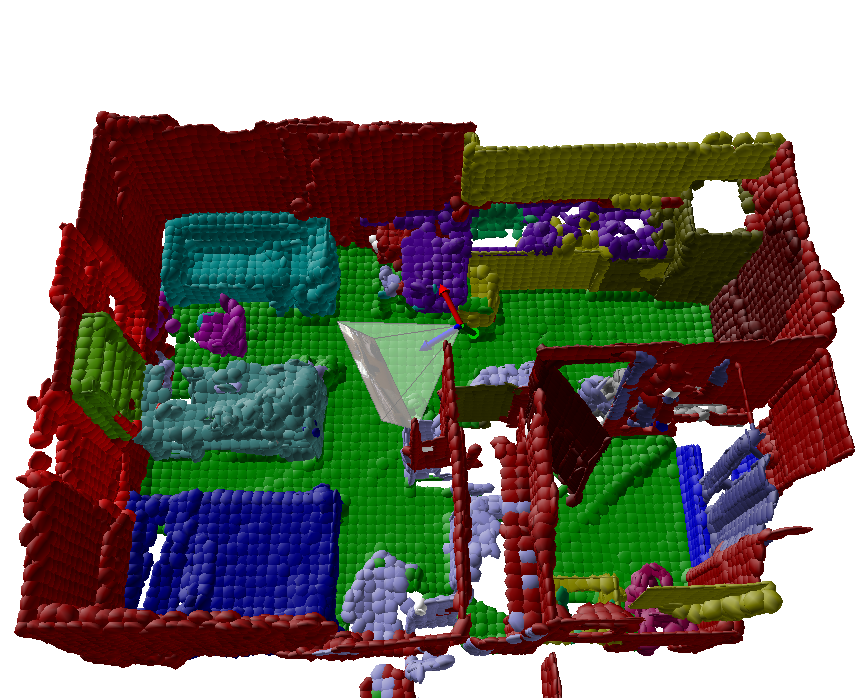}};%
       \node[right=0 of scannet_semantic_test] (scannet_instance_test) {\includegraphics[height=2.25cm, trim={0 0.5cm 0 3.2cm}, clip]{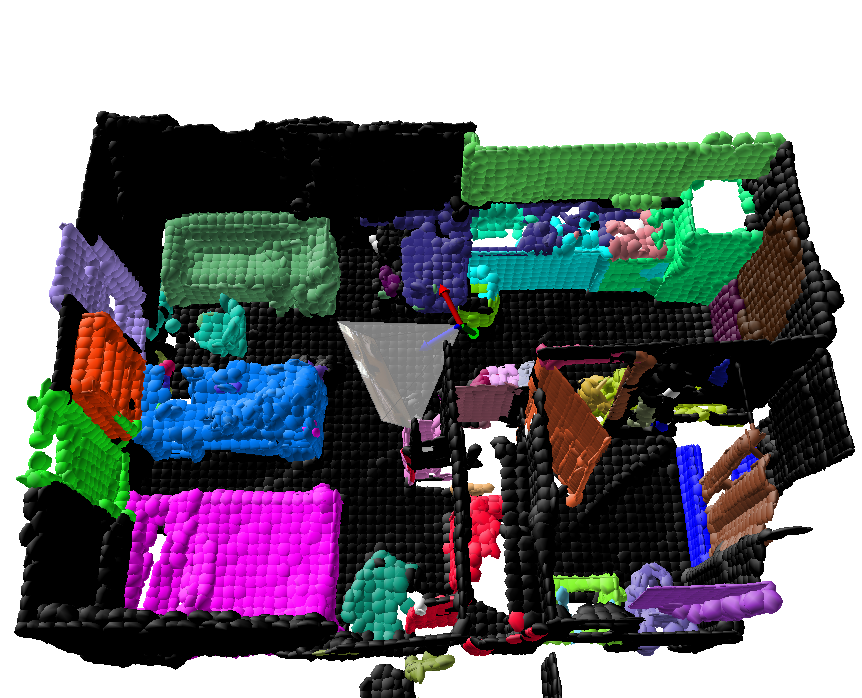}};%

       \node[left= 0.1 of hypersim_color, rotate=90, anchor=mid] (hypersim_label) {\footnotesize Hypersim};%
       \node[left= 0.1 of scannet_color_1, rotate=90, anchor=mid] {\footnotesize ScanNetV2};%
       \node[left= 0.1 of scannet_color_2, rotate=90, anchor=mid] {\footnotesize ScanNetV2};%
    \end{tikzpicture}%
    }
    \vspace{-4mm}%
    \caption{%
        Qualitative results for scene \emph{ai\_001\_10} of the Hypersim test split (top) and for \emph{scene\_0757\_00} and \emph{scene\_0761\_00} of the hidden ScanNetV2 test split~(bottom) when mapping with EMSANet predictions and voxel size 10\si{\centi\meter}. %
        Left to right: color image and panoptic back-projection for the given camera pose (see 3D scenes), panoptic, panoptic semantic, and panoptic instance NDT map. 
        Best viewed in color at 200\%. %
        Black indicates \emph{void/no\_instance}, for the semantic colors, we refer to Fig.~\ref{fig:experiments:hypersim_radar_chart}. %
        Panoptic labels are visualized by small color differences based on the semantic color.%
     }        
    \label{fig:experiments:qualitative_results}
    \vspace{-6mm}
\end{figure*}

With the results of our experiments on Hypersim at hand, we evaluate our mapping pipeline on the real-world dataset ScanNetV2. 
The results are listed in Tab.~\ref{tab:results_scannet}.
Although depth data is not perfect anymore, the results are higher than for Hypersim.
This is due to limiting the semantic classes to frequent classes and less small objects.
However, PQ results also indicate that imperfect depth data affect back-projection and matching. 
Nevertheless, compared to the state-of-the-art approaches Panoptic Multi-TSDFs and Panoptic Fusion, our PanopticNDT approach is still much more stable.
We assume that the bottom-up design of our approach as well as modeling the distributions for semantic and instance observations instead of storing a single label are the main reasons for better stability.
This also holds true when evaluating on the hidden test split. 
Fig.~\ref{fig:experiments:qualitative_results} shows qualitative results proving the real-world applicability of our approach.

\section{Application in Real-World Scenario}
\label{sec:application}
In addition to just evaluating on benchmark datasets, we further present qualitative results in a real-world application in the context of our MORPHIA project~\cite{morphia-isr2022}.
After an initial mapping phase using RTAB-Map, it is switched to localization mode to provide a reliable long-term localization.
For efficient short-term panoptic mapping, we rely on our proposed PanopticNDT with a voxel size of 10\si{\centi\meter}.
Panoptic segmentation is done using EMSANet-R34-NBt1D~\cite{emsanet2022ijcnn} trained on NYUv2~\cite{NYUv2-eccv2012}, Hypersim~\cite{hypersim-iccv2021}, SUNRGB-D~\cite{SUNRGBD-cvpr2015}, and ScanNetV2~\cite{scannet-cvpr2017}.
Even though training data do not contain any images of our Kinect Azure sensor setup, the results prove the real-world applicability of our approach. 
\ifthenelse{\boolean{isarxiv}}{%
    For impressions, we refer to \href{https://youtu.be/xS9jCEKO-Uw}{\small \url{https://youtu.be/xS9jCEKO-Uw}}.
}{%
    We refer to the attached video for impressions in this scenario.
}
\section{Conclusion}
\label{sec:conclusion}
In this paper, we have presented a novel approach for incorporating panoptic information into 3D representations. 
The approach has been integrated into efficient occupancy NDT maps, realizing panoptic occupancy NDT (PanopticNDT) maps. 
Our experiments on Hypersim and ScanNetV2 reveal that our approach represents panoptic information at a higher level of detail than other state-of-the-art approaches and enables real-time panoptic mapping on mobile robots.
We have further demonstrated that the approach can be deployed to real-world applications, enabling mobile robots to gain a strong scene understanding of their environment.

\ifthenelse{\boolean{isarxiv}}{%
}{%
    \addtolength{\textheight}{0cm}   %
}%

\bibliographystyle{bib/IEEEtran}
\bibliography{bib/literature}

\ifthenelse{\boolean{isarxiv}}{%
    \clearpage%
    \addtocounter{section}{1}
\appendixtitle

\label{sec:appendix}
Due to the space restrictions, some details had to be omitted from the main part of the paper. 
We present them here. 
In Sec.~\ref{sec:appendix:details_networks} and~\ref{sec:appendix:details_tsdf}, we give further details on the training of EMSANet~\cite{emsanet2022ijcnn} and on our adapted implementation of Panoptic Multi-TSDFs~\cite{panoptic-multi-tsdf-2022-icra} that serves for comparing PanopticNDT to Panoptic Multi-TSDFs.
Afterward, in Sec.~\ref{sec:appendix:quantitative_results} and~\ref{sec:appendix:qualitative_results}, we present omitted quantitative and qualitative results.
Finally, in Sec.~\ref{sec:appendix:application}, we give more details on the network and the pipeline used for real-world application.

\subsection{Further Network Details}
\label{sec:appendix:details_networks}
Network training was done with a batch size of 8 and without any further pre-training. 
Only the encoder was pre-trained on the ImageNet dataset~\cite{emsanet2022ijcnn}.
As the number of samples in both Hypersim and ScanNetV2 is large, we used a random subsampling in each epoch of 30\% and 25\% for Hypersim and ScanNetV2, respectively.
For ScanNetV2, we further applied a fixed subsampling of 50 to both the training and the validation split.
For both networks, the full training command including all hyperparameters is shared in our GitHub repository next to the network weights.
As we focus on fast inference for real-world applications, we did not apply any test-time tricks, such as horizontal flipping or multiscale inputs, when extracting panoptic segmentations for mapping. 

\subsection{Further Details on Panoptic Multi-TSDFs}
\label{sec:appendix:details_tsdf}
To apply Panoptic Multi-TSDFs~\cite{panoptic-multi-tsdf-2022-icra} in our data pipeline, we adapted the implementation available at GitHub\footnote{%
    GitHub Panoptic Multi-TSDFs: \url{https://github.com/ethz-asl/panoptic_mapping/} (online: 28.07.2023)
}.
We created additional data loaders for both Hypersim and ScanNetV2.
Independently of the dataset, mapping was done with inputs at the default resolution of 640${\times}$480 pixels.
Moreover, as both datasets follow the same semantic classes, we set the voxel sizes only once and used them for both datasets likewise.
We followed the already predefined sizes for assigning the voxel size for each class.
For \emph{void} and the stuff classes wall, floor, and ceiling, the preset large~(5\si{\centi\meter}) was used. 
For thing classes, the preset small~(2\si{\centi\meter}) or medium~(3\si{\centi\meter}) was used depending the object appearance.
Note that this means that Panoptic Multi-TSDFs features much finer voxel sizes than our proposed PanopticNDT.
For the remaining hyperparameters, we performed a grid search.
While most hyperparameters remained at their default value, we identified three hyperparameters with larger impact.
Tab.~\ref{tab:tsdf_parameters} lists these hyperparameters and assigns the values used for reporting results in our experiments.
It is important to note that we achieved better results when disabling the submap deactivation~(last parameter in Tab.~\ref{tab:tsdf_parameters}) completely. 
However, this would also mean that the change detection~--~one of the core features of Panoptic Multi-TSDFs~--~gets deactivated. 
Therefore, we decided for a compromise and set the deactivation parameter to 20~frames.

As Panoptic Multi-TSDFs itself does not provide a dedicated pipeline for panoptic evaluation, we implemented such a pipeline ourselves in order to be able to compare our PanopticNDT to Panoptic Multi-TSDFs.
For evaluation in 3D, i.e., mapping the created representations to ground-truth point clouds, each point of the ground-truth point cloud is queried inside the map according to the defined order described in \cite{panoptic-multi-tsdf-2022-icra}.
Conflicts due to overlapping submaps are resolved using the signed distance and the belonging probability. 
The latter describes the likelihood of a voxel being part of a submap.
This procedure was also suggested by the authors of \cite{panoptic-multi-tsdf-2022-icra} in a related GitHub issue\footnote{%
    Related GitHub issue: \url{https://github.com/ethz-asl/panoptic_mapping/issues/67} (online: 28.07.2023)
}.
After deriving the 3D representations, we follow the evaluation protocol described in Sec.~\ref{sec:experiments:eval}.

\begin{table}[t!]
    \centering
    \caption{
        Most relevant hyperparameters for Panoptic Multi-TSDFs~\cite{panoptic-multi-tsdf-2022-icra}. %
        * indicates that even higher values may improve panoptic results. 
        However, this would also mean that the change detection~--~one of the core features of Panoptic Multi-TSDFs~--~gets deactivated. 
        Therefore, we decided for a compromise and set this parameter to 20 frames. %
    }
    \vspace{-2.5mm}
    \scriptsize
    \begin{tabular}{@{}l@{\hspace{4mm}}c@{\hspace{4mm}}c@{\hspace{4mm}}c@{}}
        \toprule%
        \textbf{Parameter} & \textbf{Default} & \textbf{Hypersim} & \textbf{ScanNetV2} \\
        \midrule%
        \begin{tabular}[c]{@{}l@{}}\color[HTML]{737373}id\_tracker/\\ match\_acceptance\_threshold\end{tabular}                                & 0.1     & 0.1      & 0.15      \\[2mm]
        \begin{tabular}[c]{@{}l@{}}\color[HTML]{737373}tsdf\_integrator/\\ use\_instance\_classification\end{tabular}                          & false   & true     & true      \\[2mm]
        \begin{tabular}[c]{@{}l@{}}\color[HTML]{737373}map\_management/activity\_manager/\\ deactivate\_after\_missed\_detections\end{tabular} & 5       & 20*      & 20*       \\
        \bottomrule%
    \end{tabular}
    \label{tab:tsdf_parameters}
    \vspace{-4mm}
\end{table}

\subsection{Further Quantitative Results}
\label{sec:appendix:quantitative_results}
Due to the restricted space in the main part, we decided to omit the results for mapping with ground-truth annotations on ScanNetV2.
Tab.~\ref{tab:results_scannet_full} complements Tab.~\ref{tab:results_scannet} and shows the full results for ScanNetV2.
\begin{table*}[t!]
\centering
\caption{%
    Results on ScanNetV2 validation and hidden test split when mapping with ground truth~(top) and predicted segmentation~(bottom) of EMSANet~\cite{emsanet2022ijcnn}~(center). %
    The results listed in this table complement the results shown in Tab.~\ref{tab:results_scannet}.
    See Sec.~\ref{sec:experiments:eval} for details on the reported metrics. %
    Legend:~%
        gray: mapping with semantic only instead of panoptic information (black), %
        *: re-application in our data pipeline, %
        $x$\si{\centi\meter}: voxel size, %
        ${\dagger}$: expected to be much slower~(see Sec.~\ref{sec:related_work:panoptic_mapping}), %
        and FPS:~average update rate in frames per second on %
        our mobile robot, i.e., for single-threaded mapping: Intel NUC11PHKi7C~(Intel i7-1165G7) and for EMSANet:~NVIDIA Jetson AGX Orin (Jetpack 5.0.2, TensorRT 8.4, Float16, 30W$\,$/$\,$50W). %
}
\vspace{-2.5mm}
\resizebox{0.98\textwidth}{!}{%
\begin{tabular}{%
@{\hspace{1mm}}l%
@{\hspace{2mm}}l%
@{\hspace{6mm}}%
c@{\hspace{2mm}}c@{\hspace{1mm}}c@{\hspace{2.5mm}}c%
@{\hspace{6mm}}%
c@{\hspace{2mm}}c@{\hspace{1mm}}c@{\hspace{2.5mm}}c%
@{\hspace{10mm}}%
c@{\hspace{1mm}}c%
@{\hspace{6mm}}%
c@{\hspace{1mm}}c%
@{\hspace{10mm}}%
c@{\hspace{1mm}}}
\toprule
&                     & \multicolumn{8}{c@{\hspace{13mm}}}{\bf Valid}                           & \multicolumn{4}{c@{\hspace{13mm}}}{\bf Test}                               &      \\
&                     & \multicolumn{4}{c@{\hspace{4.5mm}}}{\bf 3D}   & \multicolumn{4}{c@{\hspace{8mm}}}{\bf 2D}   & \multicolumn{2}{c@{\hspace{6mm}}}{\bf 3D}      & \multicolumn{2}{c@{\hspace{10mm}}}{\bf 2D}   &      \\                     
&                     & \bf mIoU$\,{}^\uparrow$     & \bf mIoU${}_\text{P}^{\,\uparrow}$ & \bf PQ$\,{}^\uparrow$ & \bf AP${}_{50}^{\,\uparrow}$   & \bf mIoU$\,{}^\uparrow$     & \bf mIoU${}_\text{P}^{\,\uparrow}$ & \bf PQ$\,{}^\uparrow$ & \bf AP${}_{50}^{\,\uparrow}$    & \bf mIoU${}_\text{P}^{\,\uparrow}$ & \bf AP${}_{50}^{\,\uparrow}$      & \bf mIoU${}_\text{P}^{\,\uparrow}$ & \bf AP${}_{50}^{\,\uparrow}$     & \bf FPS$\,{}^\uparrow$  \\
\midrule
\multirow{6}{*}{\rotatebox{90}{\bf~GT}} & PanopticNDT (5\si{\centi\meter})
    & 89.91 & 90.22 & 80.30 & 94.14    & 85.91 & 85.73 & 68.20 & 77.15    & ---   & ---     & ---  & ---     & 1.23  \\
& PanopticNDT (10\si{\centi\meter})   
    & 87.23 & 87.77 & 78.44 & 91.13    & 82.06 & 81.30 & 61.22 & 62.82    & ---   & ---     & ---  & ---     & 7.51  \\
& PanopticNDT (20\si{\centi\meter})   
    & 81.36 & 82.21 & 72.92 & 82.27    & 72.09 & 71.24 & 48.02 & 44.03    & ---   & ---     & ---  & ---     & 11.47 \\
\cmidrule{2-15}
& Panoptic Multi-TSDFs~\cite{panoptic-multi-tsdf-2022-icra}* %
    & --- & 61.46 & 47.11 & 54.30    & --- & --- & --- & ---    & ---   & ---     & ---  & ---     & 6.45     \\
& SemanticNDT~\cite{semanticmapping2022icra}* (10\si{\centi\meter})%
 & \color[HTML]{737373}87.05 & --- & --- & ---     & \color[HTML]{737373}82.11 & --- & --- & ---    & --- & ---    & --- & ---    & 8.73  \\    
\midrule
\midrule
& EMSANet%
    & --- & --- & --- & ---    & \color[HTML]{737373}70.99 & 66.19 & 58.22 & 41.94    & --- & ---    & 60.0 & 38.0    & 15.4$\,$/$\,$23.1     \\
\midrule
\multirow{7}{*}{\rotatebox{90}{\bf~Prediction}} & PanopticNDT (5\si{\centi\meter})   & 70.07 & 69.37 & 57.38 & 52.79    & 68.31 & 67.71 & 52.48 & 39.18    & ---   & ---     & ---  & ---     & 1.25  \\
& PanopticNDT (10\si{\centi\meter})  & 69.47 & 68.39 & 59.19 & 52.65    & 67.70 & 66.17 & 50.48 & 37.76    & 68.1  & 50.9    & 64.8 & 39.8    & 7.77  \\
& PanopticNDT (20\si{\centi\meter}) & 66.44 & 64.74 & 56.49 & 48.21    & 60.85 & 58.85 & 41.19 & 28.66    & ---   & ---     & ---  & ---     & 10.96 \\
\cmidrule{2-15}
& Panoptic Multi-TSDFs~\cite{panoptic-multi-tsdf-2022-icra}* %
    & --- & 47.29 & 34.75 & 27.24    & --- & --- & --- & ---    & --- & ---    & --- & ---    & 5.85     \\    
& Panoptic Fusion~\cite{panoptic-fusion-2019-iros} %
    & --- & --- & 33.5 & ---    & --- & --- & --- & ---    & 52.9 & 47.8     & --- & ---    & ${\dagger}$   \\
& SemanticNDT~\cite{semanticmapping2022icra}* (10\si{\centi\meter})%
 & \color[HTML]{737373}69.59 & --- & --- & ---     & \color[HTML]{737373}67.71 & --- & --- & ---    & --- & ---    & --- & ---    & 8.73  \\
\bottomrule
\end{tabular}%
\label{tab:results_scannet_full}%
}
\vspace{-4mm}%
\end{table*}

When mapping with PanopticNDT on ScanNetV2 in the ground-truth setting, similar trends as for Hypersim~(see results in Tab.~\ref{tab:results_hypersim}) emerge.
PanopticNDT performs better with decreasing voxel size, while the step from 10\si{\centi\meter} to 5\si{\centi\meter} is already much smaller than the step from 20\si{\centi\meter} to 10\si{\centi\meter}.
A voxel size of 5\si{\centi\meter} may already be close to the upper limit of our approach for representing panoptic information, i.e., even smaller voxel sizes might not necessarily improve results.
However, in contrast to Hypersim, the results for PQ are consistently higher. 
Moreover, especially in 2D, the results for 10\si{\centi\meter} and 5\si{\centi\meter} are much closer. 
Both observations can be attributed to fewer small instances in ScanNetV2.

\subsection{Further Qualitative Results}
\label{sec:appendix:qualitative_results}
Fig.~\ref{fig:appendix:qualitative_results_hypersim} and Fig.~\ref{fig:appendix:qualitative_results_scannet} complement the qualitative results shown in Fig.~\ref{fig:experiments:qualitative_results} in the main part.
For creating 3D visualizations for Panoptic Multi-TSDFs, we take the surface points\footnote{
    During mapping, Panoptic Multi-TSDFs incrementally computes an iso-surface for each submap, describing the submap's surface by a set of points.
} of all submaps and concatenate them in a single point cloud.
For creating 2D visualizations, we follow the approximate back-projection that is already part of the system's internal instance tracker.
To create the back-projection for a given camera pose, the surface points of all visible submaps~--~that are either active or persistent~--~are projected onto the 2D camera plane.
To fill this sparse representation, each point receives an individually-sized rectangle at its center according to its distance to the camera and the voxel size of the related submap.
For back-projecting panoptic information, we use the panoptic label of the submap as fill value for all related rectangles.

The results in Fig.~\ref{fig:appendix:qualitative_results_hypersim} for mapping with ground-truth annotations show that both PanopticNDT and Panoptic Multi-TSDFs are capable of creating panoptic maps.
However, PanopticNDT can handle instances much better than Panoptic Multi-TSDFs. 
Even with the larger voxel size of 10\si{\centi\meter}, a lot of small instances are still present in the resulting map.
When switching to mapping with panoptic predictions of EMSANet, a lot of noise is introduced in the mapping process as the predictions of EMSANet are not perfect, especially for small instances far away from the camera~(as shown in this example).
However, PanopticNDT is still able to integrate such noisy predictions over time in a robust way, creating maps well suited for application.
By contrast, Panoptic Multi-TSDFs almost fails completely in this setting.
The bad performance most likely stems from the random camera trajectories together with the noisy panoptic predictions of EMSANet.
Panoptic Multi-TSDFs expects temporal consistency between incoming frames to properly allocate, track, and deactivate submaps.
Mapping with perfect ground-truth annotations can still be handled well since map regulation can fuse submaps and eliminates erroneous data by running comparisons between old and new information.
This ability is likely impaired by additional errors and inaccuracies introduced by mapping with predictions of EMSANet.

The results in Fig.~\ref{fig:appendix:qualitative_results_scannet} for ScanNetV2 show a similar picture.
In general, ScanNetV2 is less challenging as it has less semantic classes and less small objects. 
However, compared to Hypersim, there is additional noise in the depth measurements, making this dataset a good benchmark for real-world applications.
PanopticNDT performs well in this setting.
As also indicated by the results in Tab.~\ref{tab:results_scannet} and Tab.~\ref{tab:results_scannet_full}, there is no reason to switch to the smallest voxel size in such a setting.
By contrast, Panoptic Multi-TSDFs is struggling again. 
There are a lot of submaps overlapping each other.
Moreover, the upper example shows that resolving overlaps using the change detection may also lead to missing geometry in the resulting map~(the table is not present).

\subsection{Further Details on Real-World Application}
\label{sec:appendix:application}

\setcounter{table}{5}
\begin{table*}[b!]
\vspace{-5mm}
\center
\caption{
    Upper part: Results obtained for EMSANet-R34-NBt1D in the full multi-task settings of~\cite{emsanet2022ijcnn} for real-world applications~(our application network). %
    While the network was trained on NYUv2~(40~semantic classes), Hypersim~(40~semantic classes, with camera model correction presented here), SUNRGB-D~(37(+3)~semantic classes, instance annotations from here~--~PanopticNDT version), and ScanNetV2~(40~semantic classes), the training process was solely monitored based on the performance on the SUNRGB-D test split (37~semantic classes). %
    Thus, the results additionally reported for the NYUv2 test split, the Hypersim validation/test split and the ScanNet validation split~(subsample 100, 40 classes and 20 classes~(=benchmark mode)) are all within the epoch the best performance on SUNRGB-D was reached. %
    The best individual result may be higher.
    Lower part: We further present results when fine-tuning this application network on the individual datasets. 
    We refer to Sec.~V-B in \cite{emsanet2022ijcnn} for details on the reported metrics. %
    Legend: \emph{italic}:~metric used for determining the best checkpoint; {\color[HTML]{9B9B9B} gray}:~best result within the same run; Sem:~semantic segmentation; Sce:~scene classification~(unified classes of~\cite{emsanet2022ijcnn}); Ins:~instance segmentation; Or:~instance orientation estimation; and LR:~learning rate. %
}
\vspace{-2.5mm}
\scriptsize
\resizebox{0.99\textwidth}{!}{%
\begin{tabular}{
    @{\hspace{1mm}}l%
    @{\hspace{2mm}}l%
    c@{\hspace{5mm}}%
    l@{\hspace{5mm}}%
    l@{\hspace{5mm}}%
    c@{\hspace{7mm}}%
    c@{\hspace{2mm}}%
    c@{\hspace{7mm}}%
    c@{\hspace{7mm}}%
    c@{\hspace{2mm}}%
    c@{\hspace{2mm}}%
    c@{\hspace{2mm}}%
    c@{\hspace{1mm}}
    c@{}
}
\toprule
& & & & & \textbf{\begin{tabular}[c]{@{}c@{}}\color[HTML]{737373}Semantic\\ \color[HTML]{737373}decoder\end{tabular}} & \multicolumn{2}{c@{\hspace{9mm}}}{\textbf{\begin{tabular}[c]{@{}c@{}}\color[HTML]{737373}Instance\\ \color[HTML]{737373}decoder\end{tabular}}} & \textbf{\begin{tabular}[c]{@{}c@{}}\color[HTML]{5c5c5c}Scene\\ \color[HTML]{5c5c5c}head\end{tabular}} & \multicolumn{5}{c@{\hspace{9mm}}}{\textbf{\begin{tabular}[c]{@{}c@{}}\hspace{6.5mm}\color[HTML]{737373}Panoptic results\\ \hspace{6.5mm}\color[HTML]{737373}(after merging)\end{tabular}}}  \\
& \textbf{Tasks} & \textbf{Task weights} & \textbf{~~LR} & \textbf{Dataset} & \textbf{mIoU}$\,{}^\uparrow$ & \textbf{PQ}$\,{}^\uparrow$ & \textbf{MAAE}$\,{}^\downarrow$ & \textbf{bAcc}$\,{}^\uparrow$ & \textbf{mIoU}$\,{}^\uparrow$ & \textbf{PQ}$\,{}^\uparrow$ & \textbf{RQ}$\,{}^\uparrow$ & \textbf{SQ}$\,{}^\uparrow$ & \textbf{MAAE}$\,{}^\downarrow$ \\ \midrule
\multirow{10}{*}{\rotatebox{90}{\bf Application network}} & \textbf{Sem$\,$+$\,$Sce$\,$+$\,$Ins$\,$+$\,$Or} & $1{\,:\,}0.25{\,:\,}3{\,:\,}0.5$ & 0.0005 & SUNRGB-D & %
    49.30 & 61.33 & 18.31 & 55.45 & 47.32 & \emph{50.91} & 58.40 & 86.06 & 14.28 \\
& & {\color[HTML]{9B9B9B} } & \multicolumn{1}{l}{{\color[HTML]{9B9B9B} }} &  & %
    {\color[HTML]{9B9B9B} 49.31} & {\color[HTML]{9B9B9B} 62.12} & {\color[HTML]{9B9B9B} 18.24} & {\color[HTML]{9B9B9B} 58.56} & {\color[HTML]{9B9B9B} 47.32} & {\color[HTML]{9B9B9B} \emph{50.91}} & {\color[HTML]{9B9B9B} 58.40} & {\color[HTML]{9B9B9B} 86.22} & {\color[HTML]{9B9B9B} 14.19} \\
\cmidrule{5-14}
& & & & NYUv2 & %
    56.55 & 63.54 & 16.70 & 70.29 & 55.42 & 48.47 & 56.77 & 84.27 & 14.58 \\
\cmidrule{5-14}    
& & & & Hypersim (valid) & %
    28.60 & 47.25 & --- & 30.02 & 27.06 & 17.23 & 21.30 & 74.09 & --- \\
& & & & Hypersim (test) & %
    29.03 & 42.40 & --- & 46.17 & 26.75 & 17.95 & 22.05 & 69.74 & --- \\
\cmidrule{5-14}
& & & & ScanNetV2 (40 classes) & %
    50.84 & 58.68 & --- & 49.01 & 49.32 & 40.49 & 49.13 & 81.20 & --- \\
\cmidrule{5-14}    
& & & & ScanNetV2 (20 classes) & %
    66.96 & 58.60 & --- & 49.01 & 65.27 & 53.92 & 63.71 & 84.15 & --- \\
\midrule
\midrule
\multirow{15}{*}{\rotatebox{90}{\bf ~~{\color[HTML]{9B9B9B}Application network} + Fine-tuning}} & \textbf{Sem$\,$+$\,$Sce$\,$+$\,$Ins$\,$+$\,$Or} & $1{\,:\,}0.25{\,:\,}3{\,:\,}0.5$ & 0.002 & SUNRGB-D & %
    50.86 & 63.40 & 16.95 & 58.55 & 47.88 & \emph{52.48} & 60.03 & 86.30 & 13.02 \\
& & {\color[HTML]{9B9B9B} } & \multicolumn{1}{l}{{\color[HTML]{9B9B9B} }} &  & %
    {\color[HTML]{9B9B9B} 50.91} & {\color[HTML]{9B9B9B} 63.61} & {\color[HTML]{9B9B9B} 16.87} & {\color[HTML]{9B9B9B} 60.17} & {\color[HTML]{9B9B9B} 48.30} & {\color[HTML]{9B9B9B} \emph{52.48}} & {\color[HTML]{9B9B9B} 60.03} & {\color[HTML]{9B9B9B} 86.39} & {\color[HTML]{9B9B9B} 13.00} \\
\cmidrule{4-14}
&  &  & 0.004 & NYUv2 & %
    59.02 & 64.79 & 16.12 & 73.33 & 57.83 & \emph{51.15} & 59.59 & 84.80 & 14.31 \\
& & {\color[HTML]{9B9B9B} } & \multicolumn{1}{l}{{\color[HTML]{9B9B9B} }} &  & %
    {\color[HTML]{9B9B9B} 59.13} & {\color[HTML]{9B9B9B} 65.17} & {\color[HTML]{9B9B9B} 15.76} & {\color[HTML]{9B9B9B} 76.82} & {\color[HTML]{9B9B9B} 57.90} & {\color[HTML]{9B9B9B} \emph{51.15}} & {\color[HTML]{9B9B9B} 60.04} & {\color[HTML]{9B9B9B} 84.99} & {\color[HTML]{9B9B9B} 13.86} \\
\cmidrule{4-14}
&  &  & 0.001 & Hypersim (valid) & %
    49.06 & 53.90 & --- & 32.89 & 47.80 & \emph{34.22} & 39.98 & 82.88 & --- \\
& & {\color[HTML]{9B9B9B} } & \multicolumn{1}{l}{{\color[HTML]{9B9B9B} }} &  & %
    {\color[HTML]{9B9B9B} 49.06} & {\color[HTML]{9B9B9B} 54.22} & {\color[HTML]{9B9B9B} ---} & {\color[HTML]{9B9B9B} 35.48} & {\color[HTML]{9B9B9B} 47.80} & {\color[HTML]{9B9B9B} \emph{34.22}} & {\color[HTML]{9B9B9B} 82.98} & {\color[HTML]{9B9B9B} 39.98} & {\color[HTML]{9B9B9B} ---} \\
\cmidrule{5-14}
&  &  &  & Hypersim (test) & %
    50.23 & 49.99 & --- & 54.62 & 47.58 & 31.62 & 37.17 & 73.06 & --- \\
\cmidrule{4-14}
&  &  & 0.002 & ScanNetV2 (40 classes) & %
    52.08 & 59.67 & --- & 49.65 & 50.19 & \emph{41.72} & 50.13 & 79.42 & --- \\
& & {\color[HTML]{9B9B9B} } & \multicolumn{1}{l}{{\color[HTML]{9B9B9B} }} &  & %
    {\color[HTML]{9B9B9B} 52.26} & {\color[HTML]{9B9B9B} 60.28} & {\color[HTML]{9B9B9B} ---} & {\color[HTML]{9B9B9B} 50.32} & {\color[HTML]{9B9B9B} 50.25} & {\color[HTML]{9B9B9B} \emph{41.72}} & {\color[HTML]{9B9B9B} 50.53} & {\color[HTML]{9B9B9B} 80.51} & {\color[HTML]{9B9B9B} ---} \\
\cmidrule{4-14}
&  &  & 0.0005 & ScanNetV2 (20 classes) & %
    70.52 & 70.60 & --- & 50.46 & 67.65 & \emph{58.47} & 67.65 & 85.94 & --- \\
& & {\color[HTML]{9B9B9B} } & \multicolumn{1}{l}{{\color[HTML]{9B9B9B} }} &  & %
    {\color[HTML]{9B9B9B} 70.81} & {\color[HTML]{9B9B9B} 71.13} & {\color[HTML]{9B9B9B} ---} & {\color[HTML]{9B9B9B} 51.33} & {\color[HTML]{9B9B9B} 67.69} & {\color[HTML]{9B9B9B} \emph{58.47}} & {\color[HTML]{9B9B9B} 67.65} & {\color[HTML]{9B9B9B} 86.07} & {\color[HTML]{9B9B9B} ---} \\
\bottomrule
\end{tabular}%
}
\label{tab:emsanet_application}
\end{table*}
\setcounter{table}{4}

As already described in the main part and shown in the video, we also use EMSANet-R34-NBt1D~\cite{emsanet2022ijcnn} for predicting panoptic segmentations in our real-world applications. 
The application network processes inputs at the Hypersim input resolution~(1024${\times}$768 pixels) but was trained with samples from the four datasets NYUv2~\cite{NYUv2-eccv2012}, Hypersim, SUNRGB-D~\cite{SUNRGBD-cvpr2015}, and ScanNetV2.
We decided in favor of combining these datasets, as each dataset contributes additional information to the training process.
Hypersim contains a lot of (small) instances that greatly boost instance segmentation. 
We observed that pre-training on Hypersim already helps a lot, but adding Hypersim examples to the target training has an even larger impact on the generalization capability of the final network for real-world application.

However, as Hypersim itself is large and provides only perfect depth data, we limited the number of samples and used a random subsampling of 10\% in each epoch.
The remaining three datasets are captured in real environments and, thus, provide valuable information for application.
We used the raw depth in all datasets for training to force the network to cope with incomplete depth data.
To prevent ScanNetV2 from prevailing the real-world part of the training data, we used a random subsampling of 25\% in each epoch.
NYUv2 and SUNRGB-D are used along with the additional annotations provided in~\cite{emsanet2022ijcnn}.
Note that we used both datasets even though SUNRGB-D already contains all samples of NYUv2.
However, in the SUNRGB-D version, the last three semantic classes are omitted and mapped to \emph{void}. 
Using both datasets at the same time increases the number of real-world examples for these three classes.
For SUNRGB-D, we further refined the pipeline of~\cite{emsanet2022ijcnn} for extracting instance annotations from 3D bounding boxes.
We refer to the version of SUNRGB-D with refined instance annotations presented in this paper as PanopticNDT version. 
Tab.~\ref{tab:sunrgbd} shows that the total number of available instance annotations could be greatly increased.
The refined annotations are shared in our GitHub repository.
The best checkpoint was chosen based on the performance on the SUNRGB-D test split, as it has the biggest impact on our application due its variability in scenes and cameras.
\begin{table}[!t]
    \caption{
        Comparison between annotations provided for the SUNRGB-D dataset~\cite{SUNRGBD-cvpr2015} in the EMSANet publication~\cite{emsanet2022ijcnn}~(EMSANet version) and the refined annotations proposed in this publication~(PanopticNDT version).
    }
    \scriptsize
    
    \centering%
    \vspace{-2.5mm}
    \begin{tabular}{@{}l@{\hspace{8mm}}lrrr@{}}
        \toprule
        & \textbf{Split} & \textbf{\#$\,$Images} & \textbf{\#$\,$Instances}    & \textbf{\#$\,$Orientations} \\ \midrule
        \textbf{SUNRGB-D}                         & train & 5,285    & 18,171    & 13,076  \\
        (EMSANet version~\cite{emsanet2022ijcnn}) & test  & 5,050    & 16,961    & 12,440  \\ 
        \midrule
        \textbf{SUNRGB-D}                         & train & 5,285    & 24,184    & 19,292  \\
        (PanopticNDT version)                     & test  & 5,050    & 23,769    & 19,409  \\ \bottomrule
    \end{tabular}
    \label{tab:sunrgbd}%
    \vspace{-6mm}
\end{table}
\setcounter{table}{6}

Tab.~\ref{tab:emsanet_application} shows the results obtained for our final application network.
Note that EMSANet-R34-NBt1D~\cite{emsanet2022ijcnn} is a multi-task approach that also predicts instance orientations as well as an overall scene label for a given input.
We also integrate both information over time in our NDT representation, enabling our mobile robots to gain an even stronger scene understanding. 
However, this is not in the scope of this publication.
The results in the upper part of Tab.~\ref{tab:emsanet_application} indicate that the application network benefits from being trained on multiple datasets. 
For the real-world datasets NYUv2 and SUNRGB-D, the results already exceed those from~\cite{emsanet2022ijcnn}.
The results for Hypersim and ScanNetV2 reflect that combined training puts less focus on these datasets. 
However, when comparing the results for ScanNetV2 to those in Tab~\ref{tab:results_scannet}, the results are still good, considering that the model was trained on all 40 classes instead of only 20\footnote{
    For the ScanNetV2 benchmark, networks are typically trained directly on 20 classes, as distinguishing 20 classes is easier than distinguishing 40 classes.
    Our application network instead was trained on 40 classes to enable combining all four datasets. 
    For evaluating the 20-classes setting, we simply mapped predictions for ignored classes to void.
}.
\begin{figure*}[!b]
    \vspace{-3mm}%
    \centering%
    \begin{tikzpicture}[inner sep=0pt]     
        \newcommand{\includeimageddd}[1]{\includegraphics[height=1.92cm, trim={0 2cm 1cm 1.7cm}, clip]{#1}}%
        \newcommand{\includeimagedd}[1]{\includegraphics[height=1.92cm]{#1}}%
        \node (gt_ndt_05_panoptic) {\includeimageddd{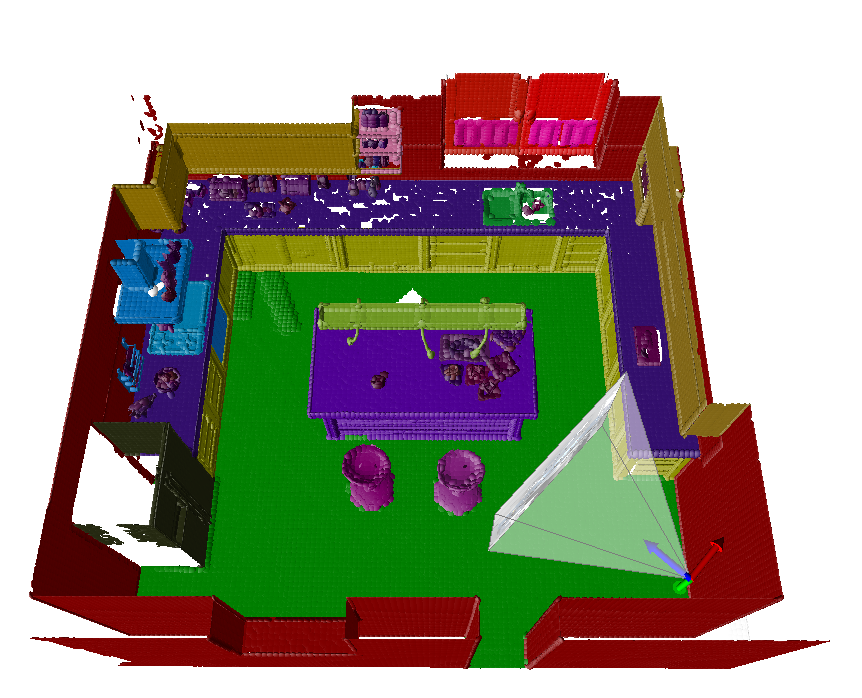}};%
        \node[right=-0.5mm of gt_ndt_05_panoptic] (gt_ndt_05_semantic) {\includeimageddd{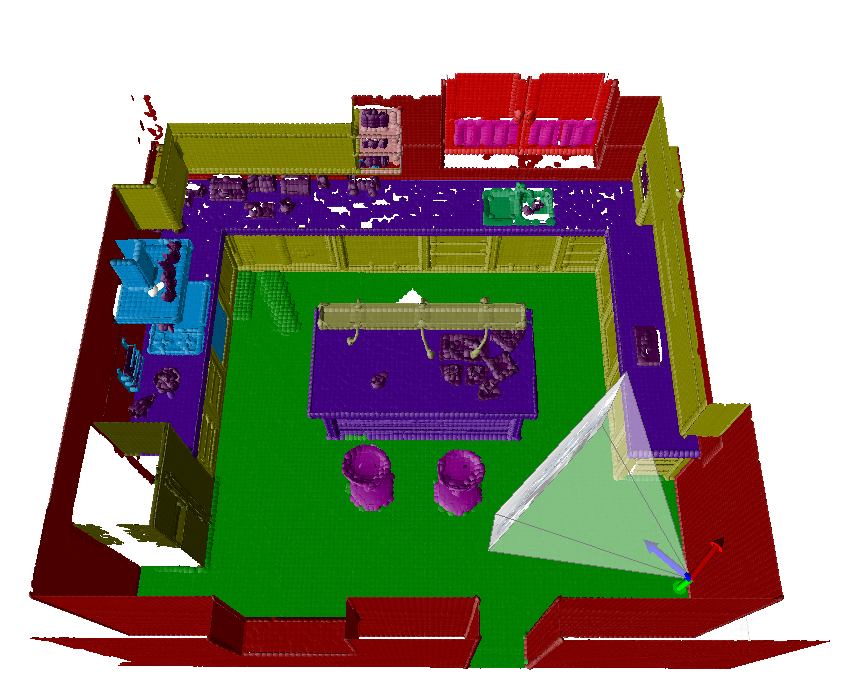}};%
        \node[right=-0.5mm of gt_ndt_05_semantic] (gt_ndt_05_instance) {\includeimageddd{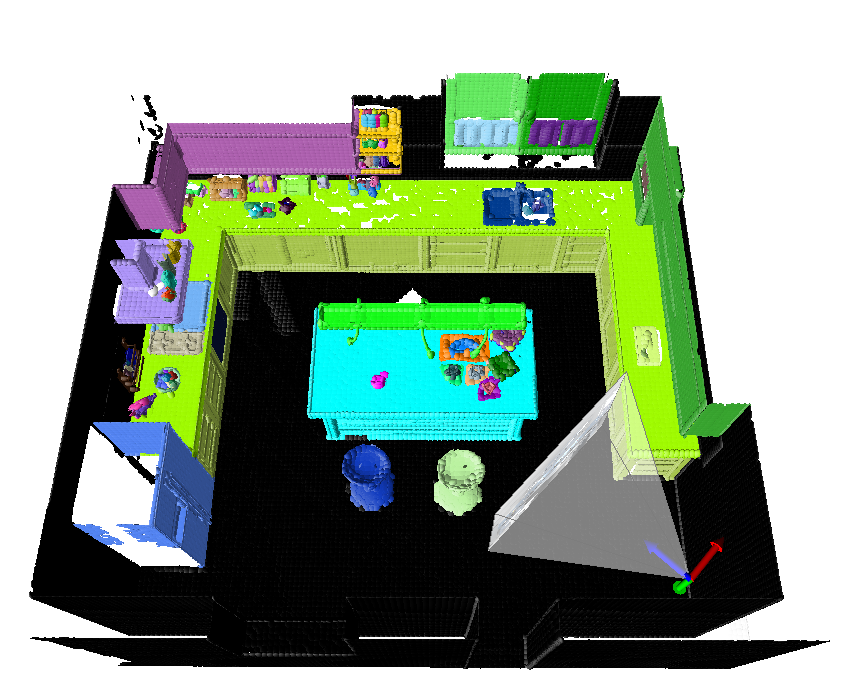}};%
        \node[right=4mm of gt_ndt_05_instance] (gt_ndt_05_panoptic_2d) {\includeimagedd{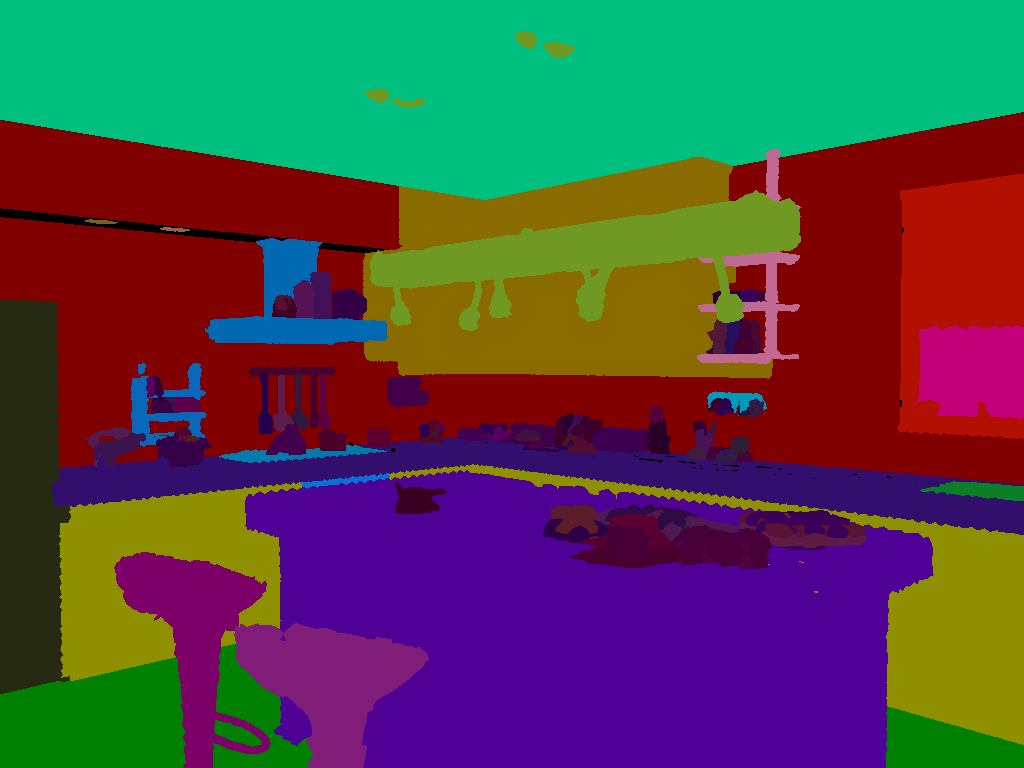}};%
        \node[right=2mm of gt_ndt_05_panoptic_2d] (gt_ndt_05_semantic_2d) {\includeimagedd{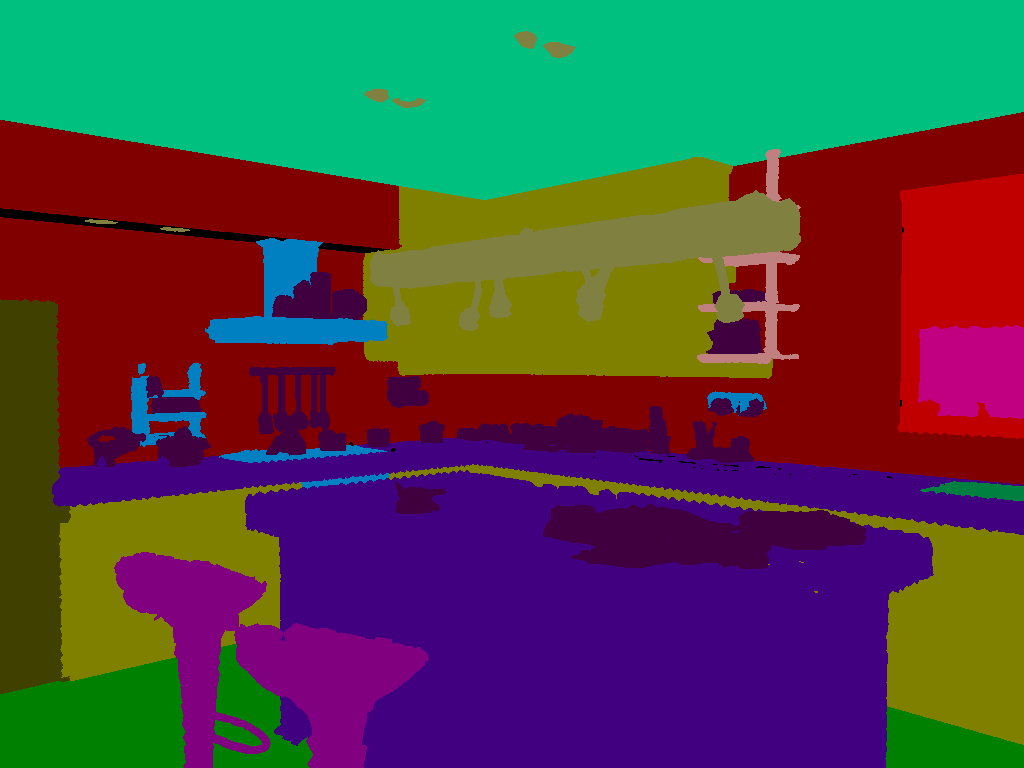}};%
        \node[right=2mm of gt_ndt_05_semantic_2d] (gt_ndt_05_instance_2d) {\includeimagedd{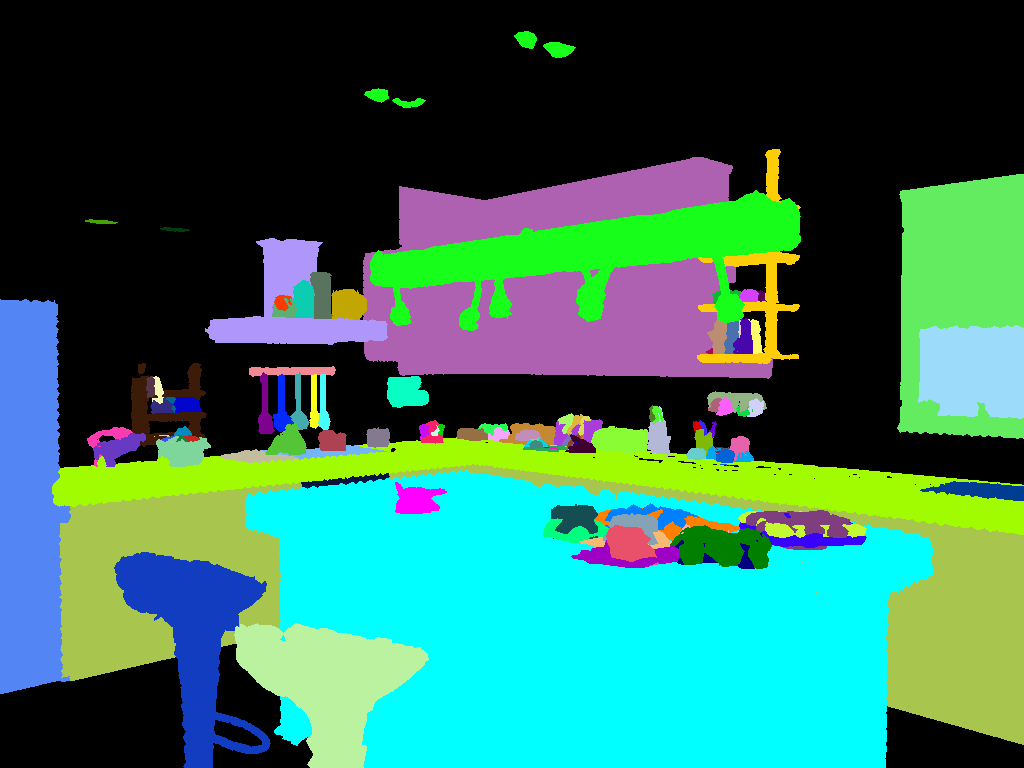}};%
        \node[below=2mm of gt_ndt_05_panoptic] (pred_ndt_05_panoptic) {\includeimageddd{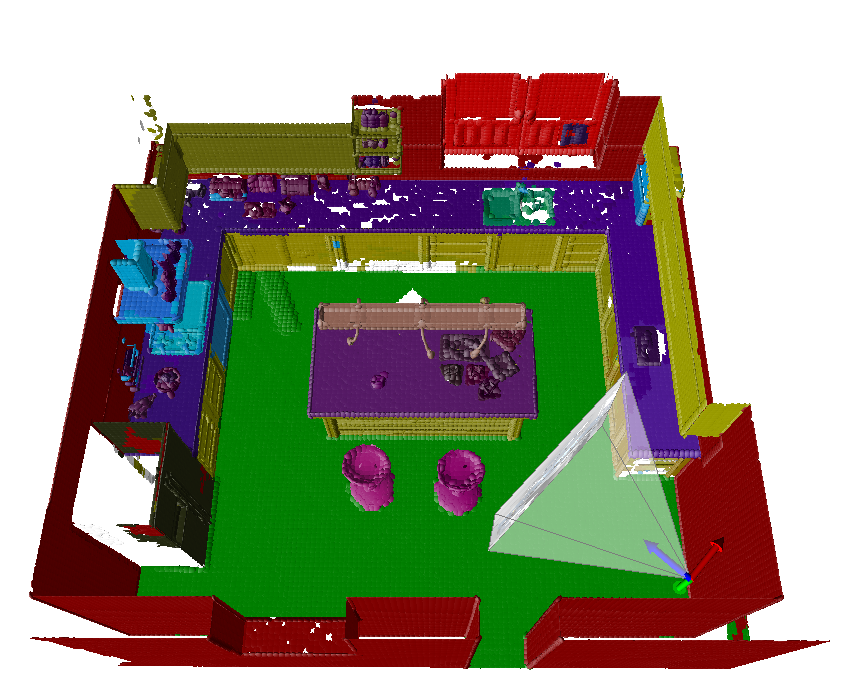}};%
        \node[right=-0.5mm of pred_ndt_05_panoptic] (pred_ndt_05_semantic) {\includeimageddd{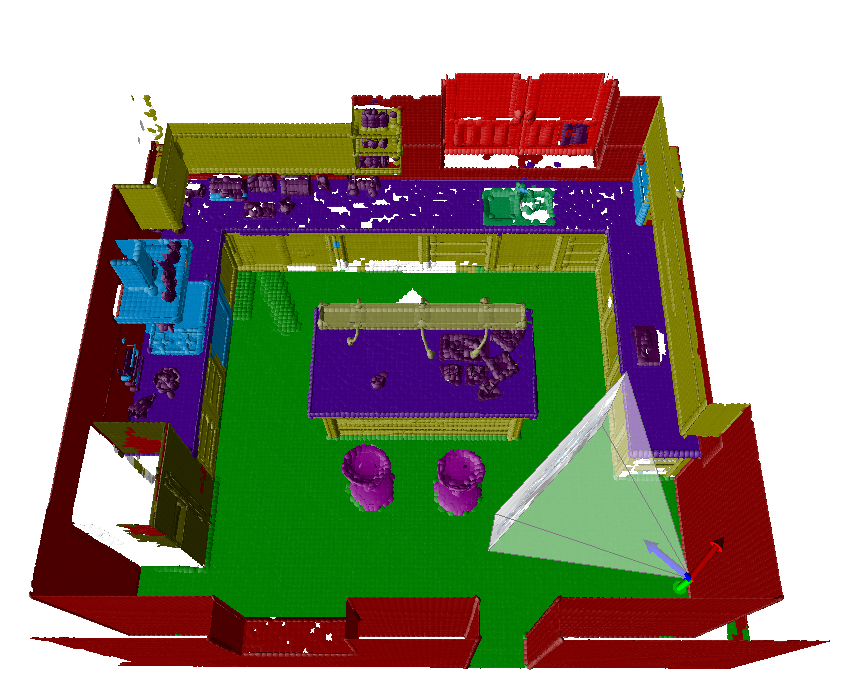}};%
        \node[right=-0.5mm of pred_ndt_05_semantic] (pred_ndt_05_instance) {\includeimageddd{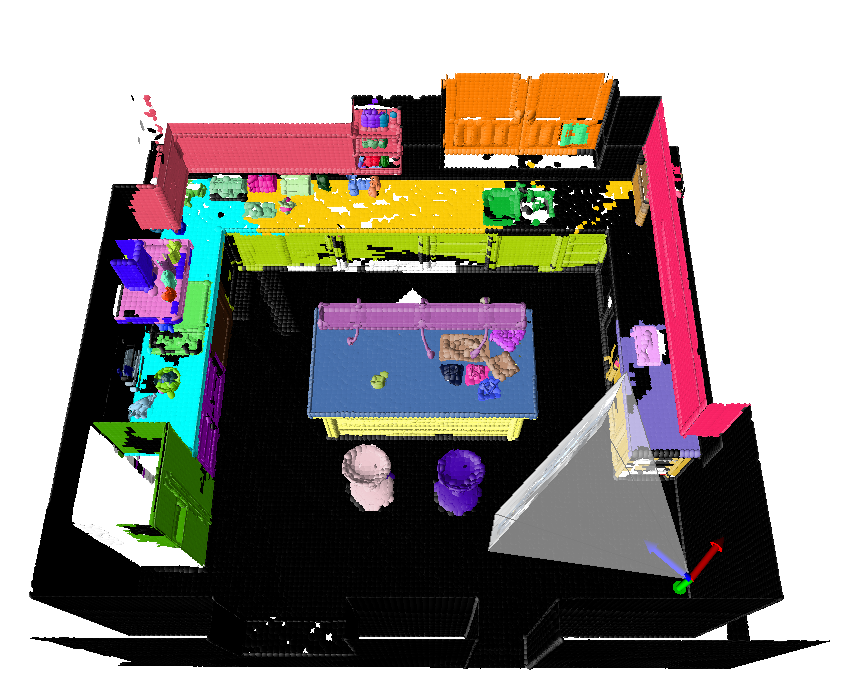}};%
        \node[right=4mm of pred_ndt_05_instance] (pred_ndt_05_panoptic_2d) {\includeimagedd{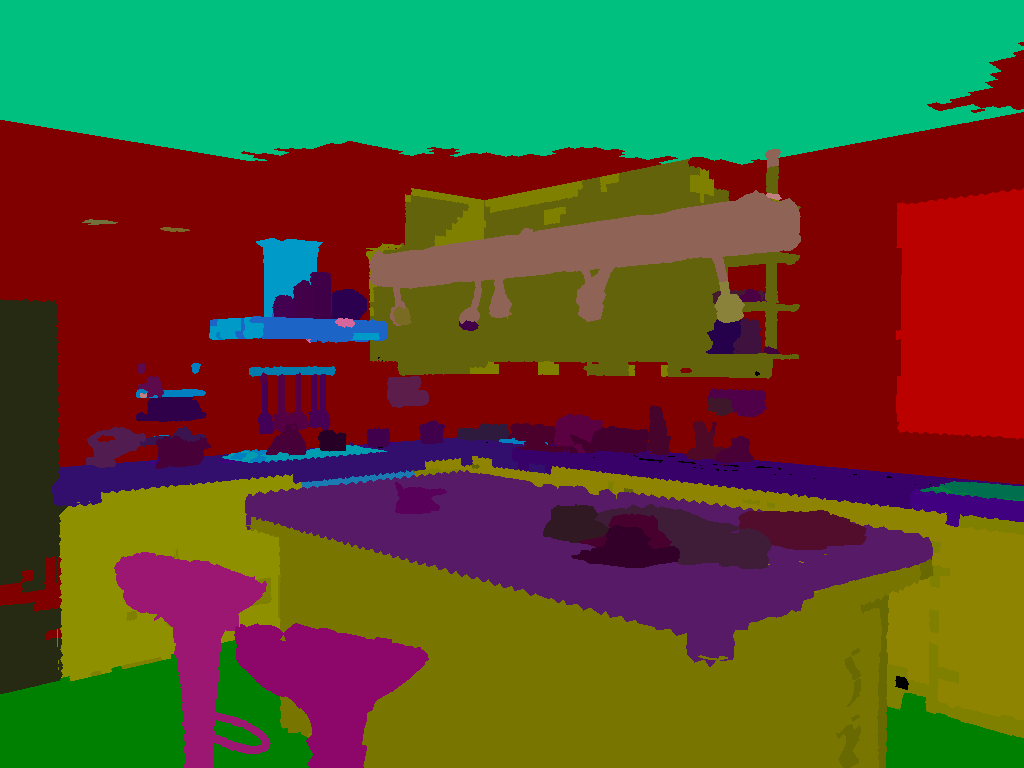}};%
        \node[right=2mm of pred_ndt_05_panoptic_2d] (pred_ndt_05_semantic_2d) {\includeimagedd{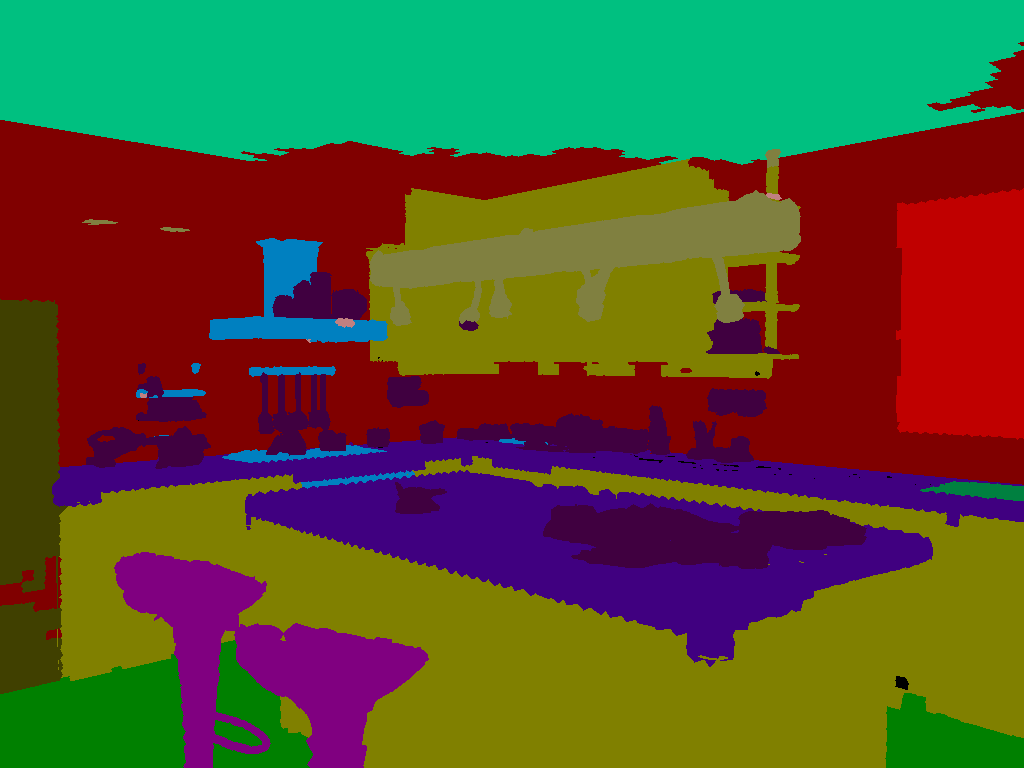}};%
        \node[right=2mm of pred_ndt_05_semantic_2d] (pred_ndt_05_instance_2d) {\includeimagedd{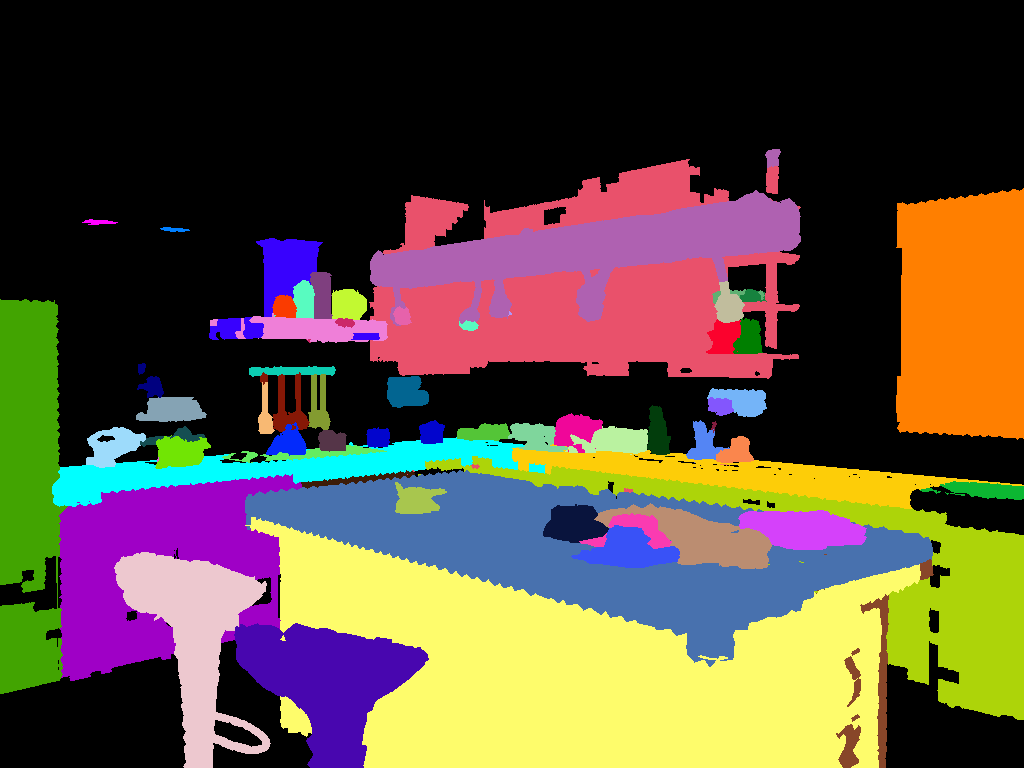}};%
        \node[below=5mm of pred_ndt_05_panoptic] (gt_ndt_10_panoptic) {\includeimageddd{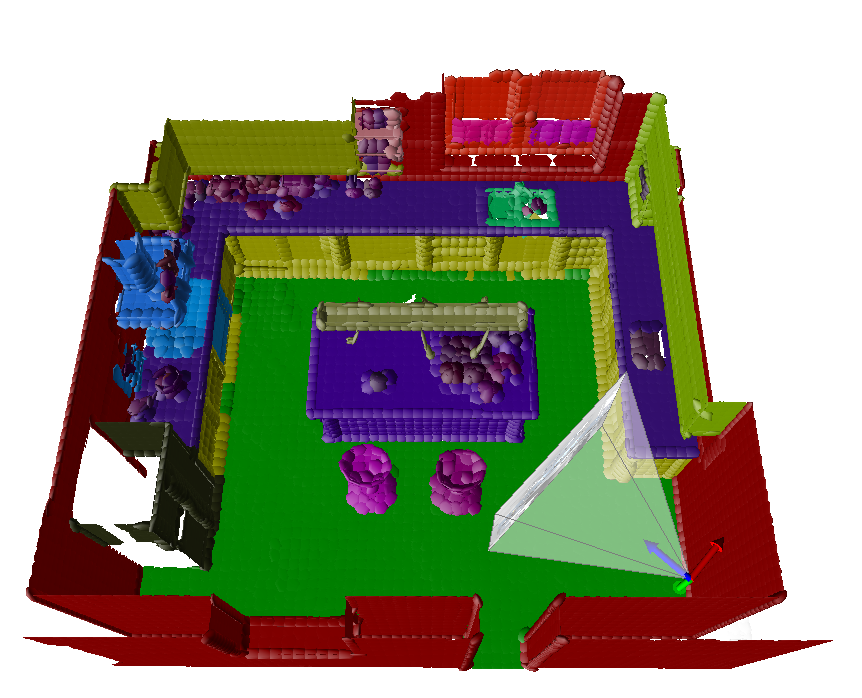}};%
        \node[right=-0.5mm of gt_ndt_10_panoptic] (gt_ndt_10_semantic) {\includeimageddd{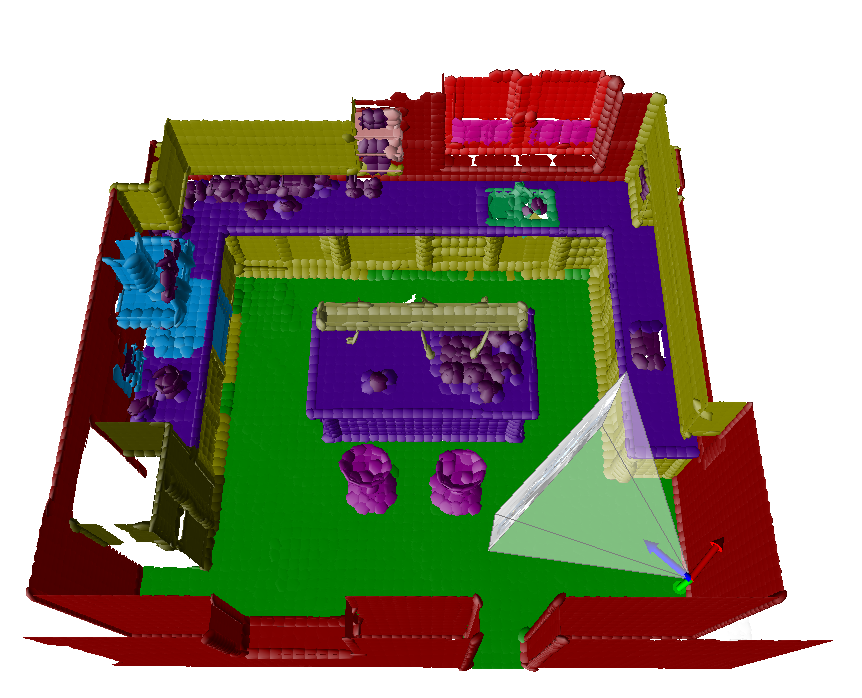}};%
        \node[right=-0.5mm of gt_ndt_10_semantic] (gt_ndt_10_instance) {\includeimageddd{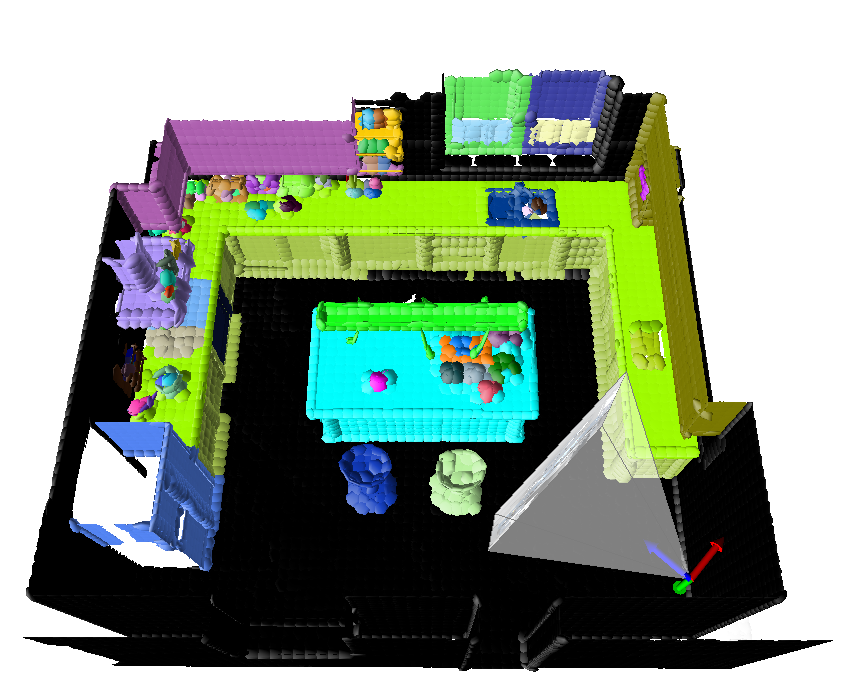}};%
        \node[right=4mm of gt_ndt_10_instance] (gt_ndt_10_panoptic_2d) {\includeimagedd{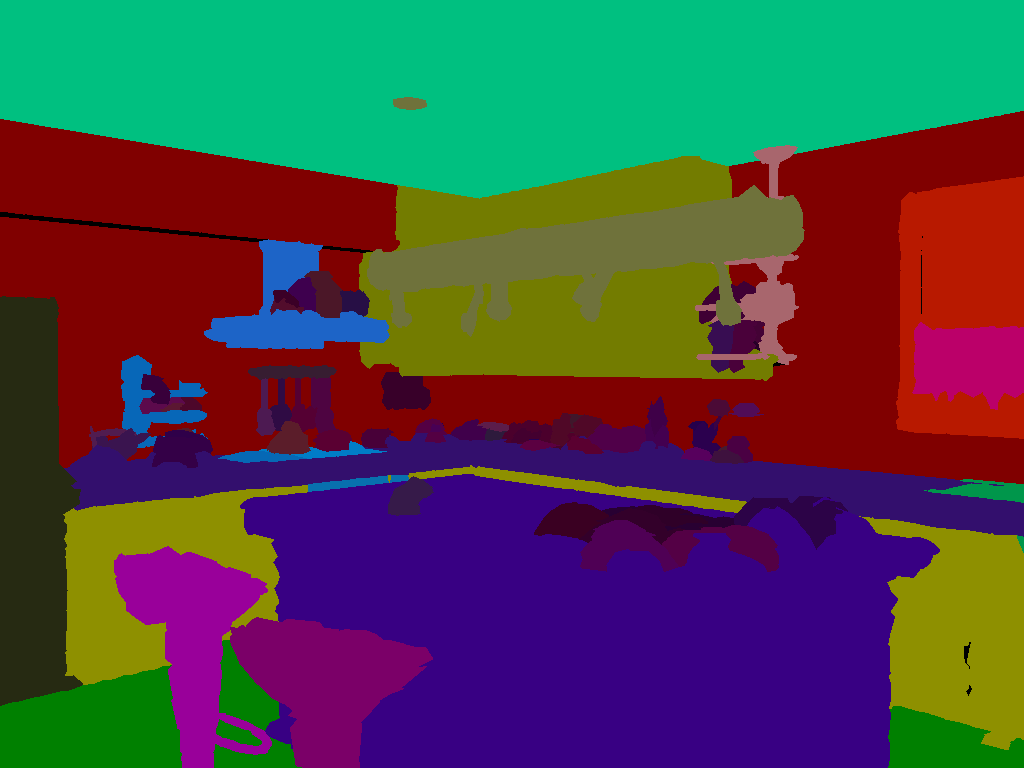}};%
        \node[right=2mm of gt_ndt_10_panoptic_2d] (gt_ndt_10_semantic_2d) {\includeimagedd{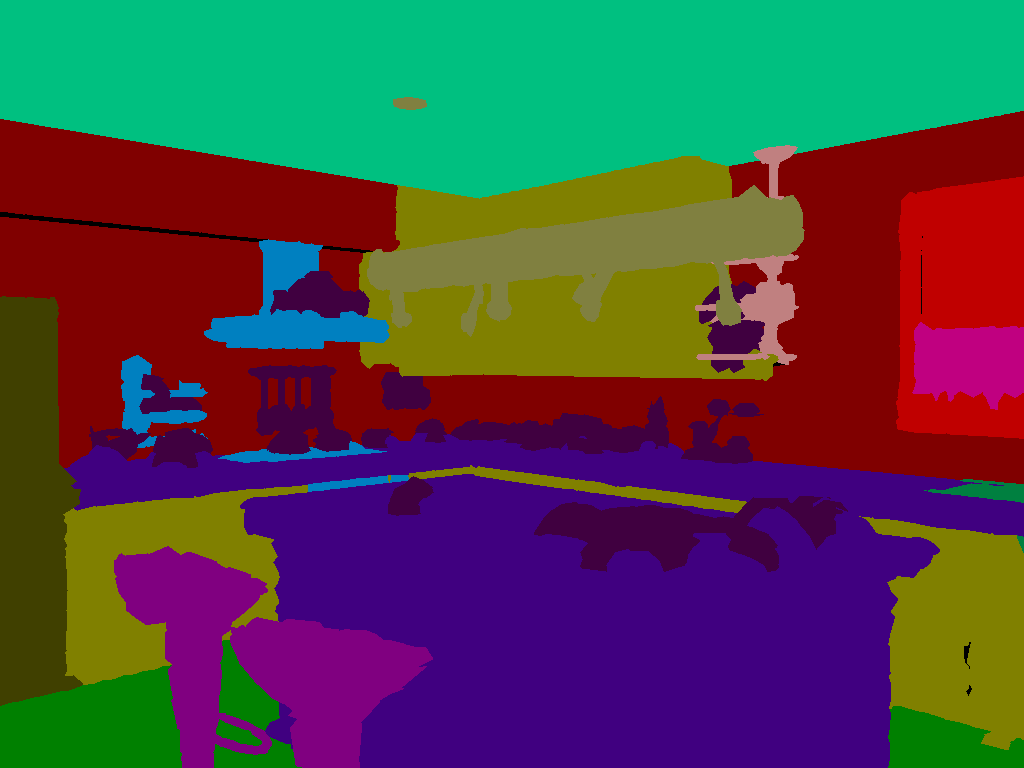}};%
        \node[right=2mm of gt_ndt_10_semantic_2d] (gt_ndt_10_instance_2d) {\includeimagedd{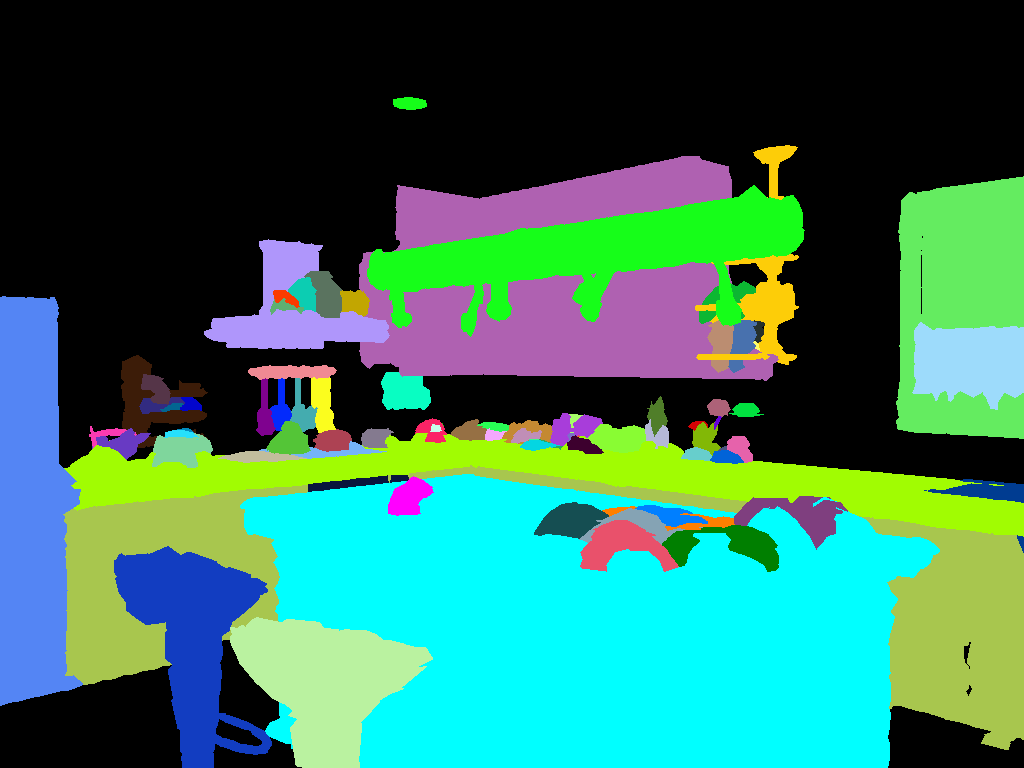}};%
        \node[below=2mm of gt_ndt_10_panoptic] (pred_ndt_10_panoptic) {\includeimageddd{img/examples/hypersim_example_0/pred_ndt_10_panoptic.png}};%
        \node[right=-0.5mm of pred_ndt_10_panoptic] (pred_ndt_10_semantic) {\includeimageddd{img/examples/hypersim_example_0/pred_ndt_10_semantic.png}};%
        \node[right=-0.5mm of pred_ndt_10_semantic] (pred_ndt_10_instance) {\includeimageddd{img/examples/hypersim_example_0/pred_ndt_10_instance.png}};%
        \node[right=4mm of pred_ndt_10_instance] (pred_ndt_10_panoptic_2d) {\includeimagedd{img/examples/hypersim_example_0/pred_ndt_10_backprojection_0000_panoptic.png}};%
        \node[right=2mm of pred_ndt_10_panoptic_2d] (pred_ndt_10_semantic_2d) {\includeimagedd{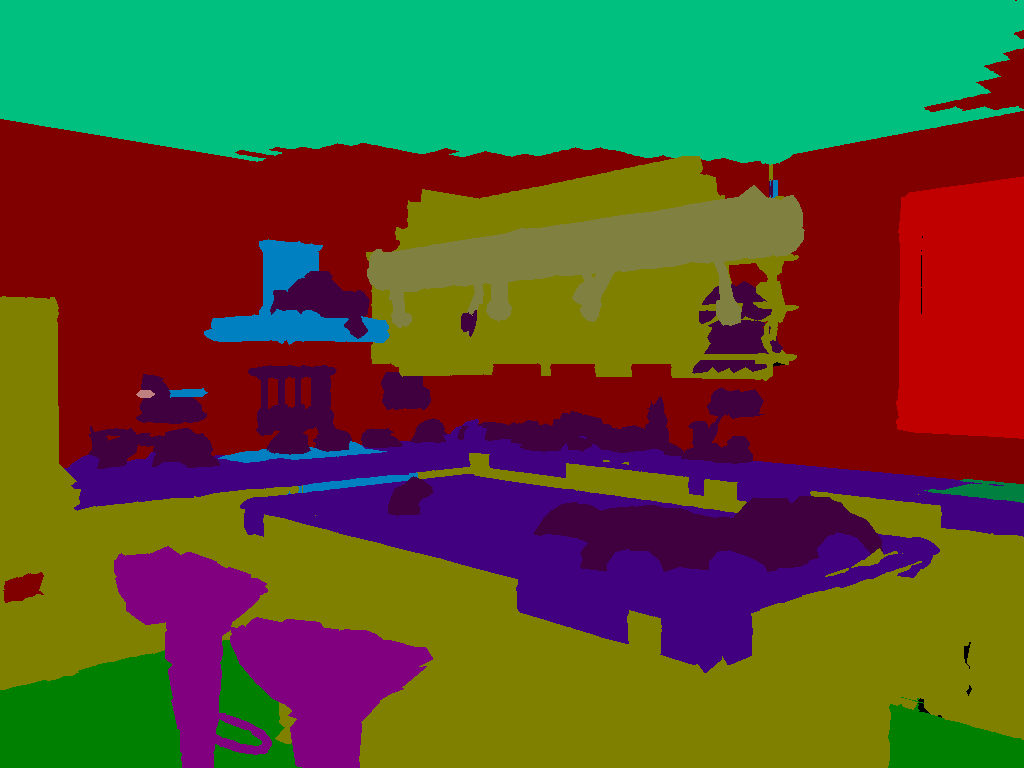}};%
        \node[right=2mm of pred_ndt_10_semantic_2d] (pred_ndt_10_instance_2d) {\includeimagedd{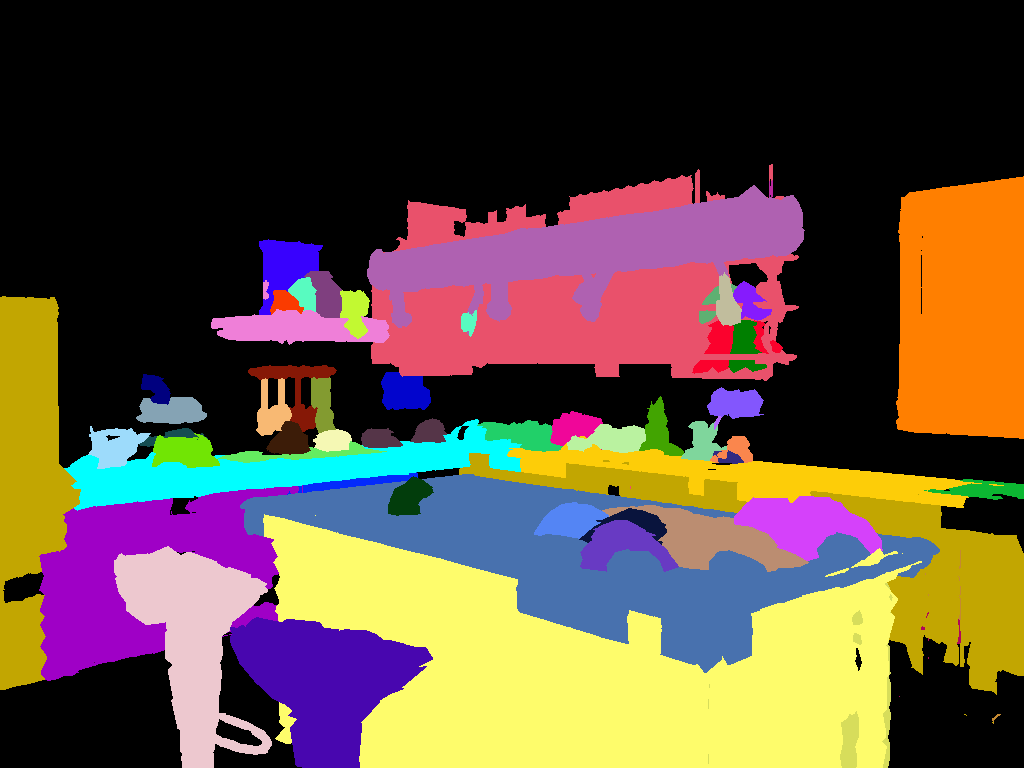}};%
        \node[below=5mm of pred_ndt_10_panoptic] (gt_pmtsdf_panoptic) {\includeimageddd{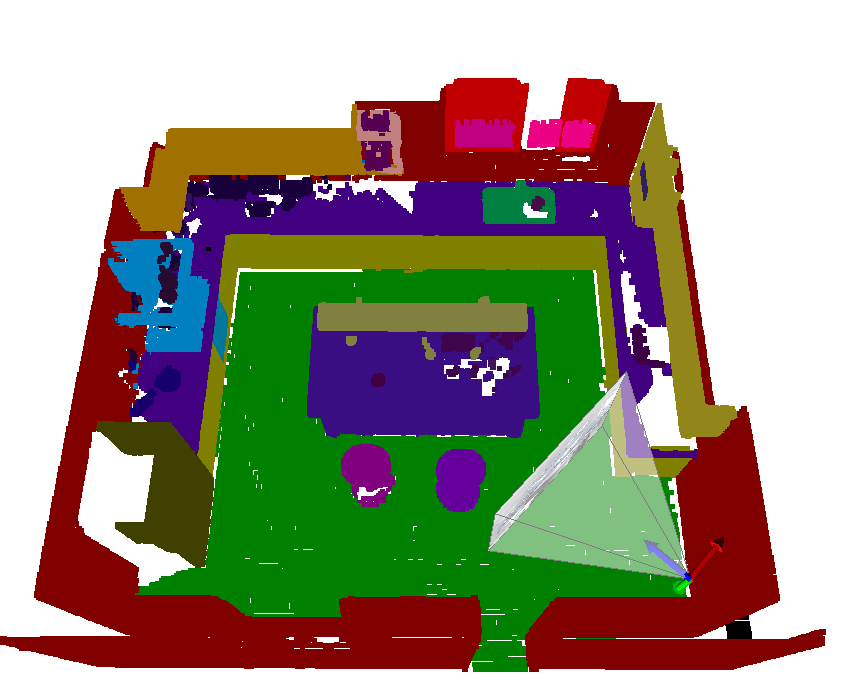}};%
        \node[right=-0.5mm of gt_pmtsdf_panoptic] (gt_pmtsdf_semantic) {\includeimageddd{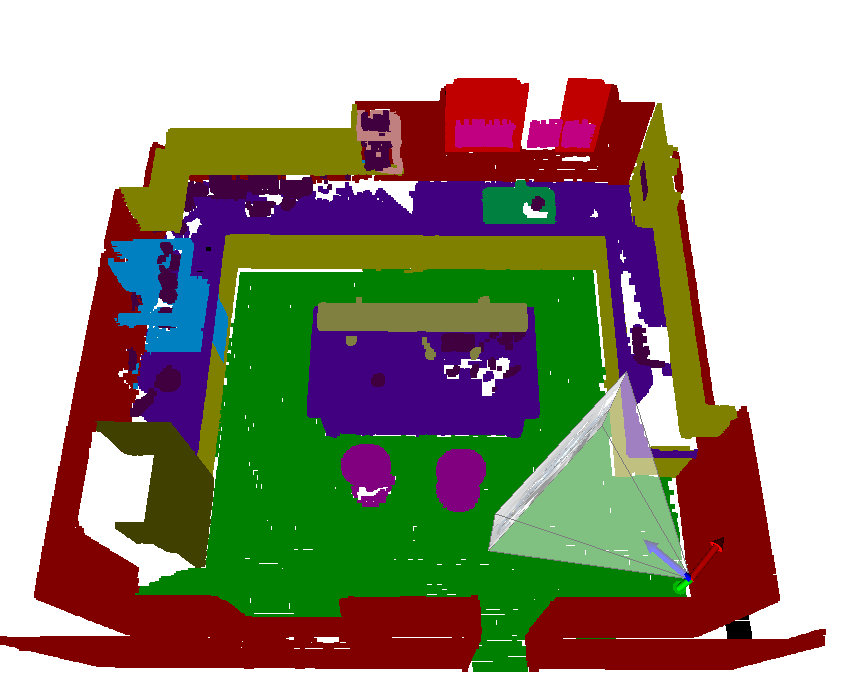}};%
        \node[right=-0.5mm of gt_pmtsdf_semantic] (gt_pmtsdf_instance) {\includeimageddd{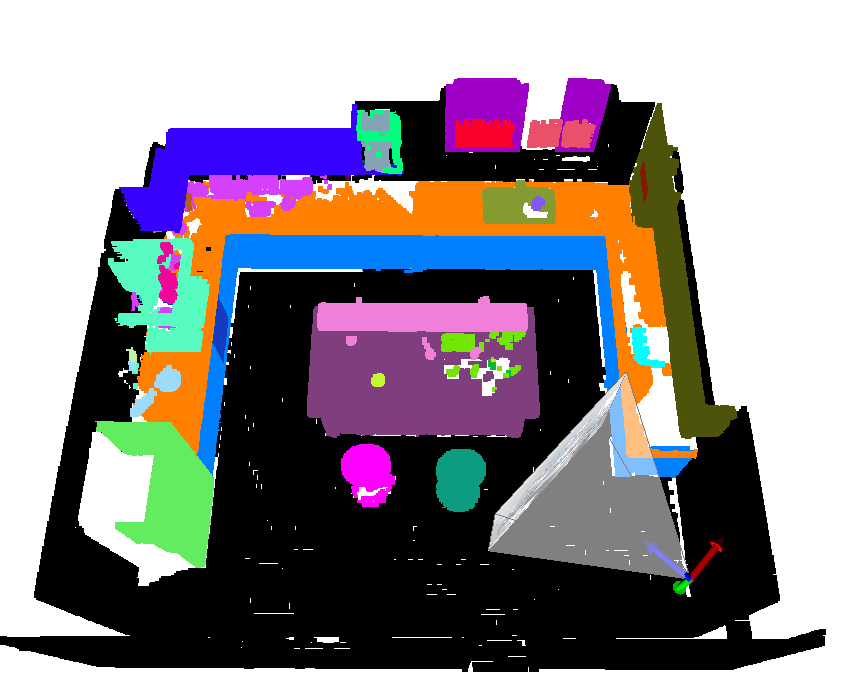}};%
        \node[right=4mm of gt_pmtsdf_instance] (gt_pmtsdf_panoptic_2d) {\includeimagedd{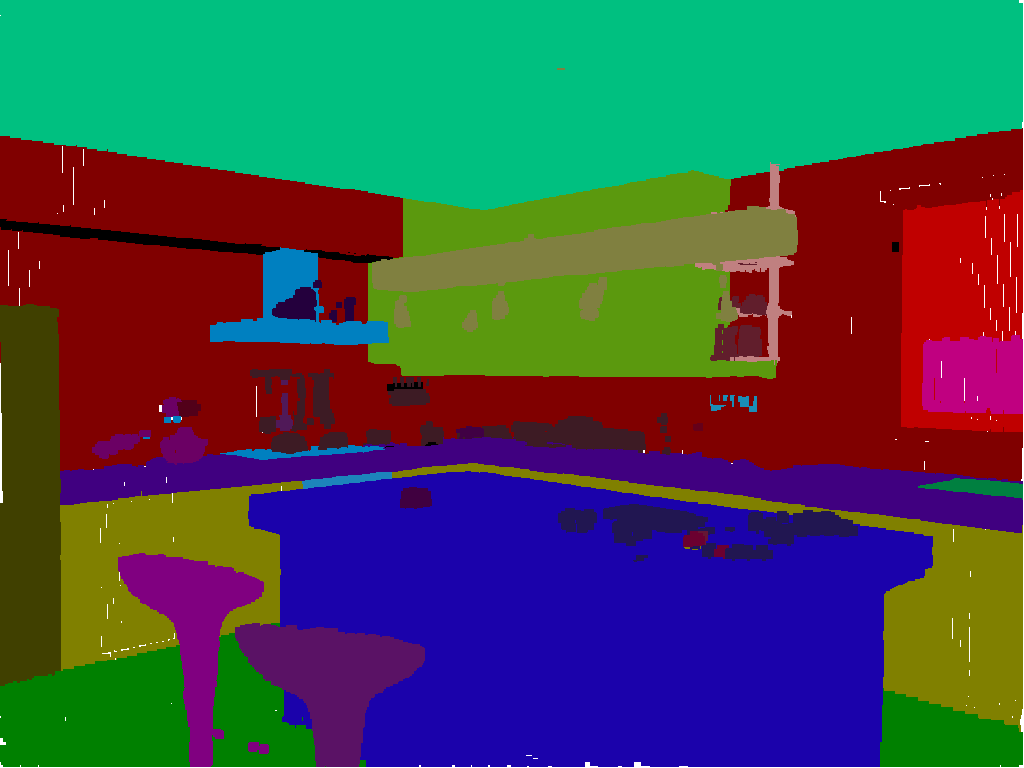}};%
        \node[right=2mm of gt_pmtsdf_panoptic_2d] (gt_pmtsdf_semantic_2d) {\includeimagedd{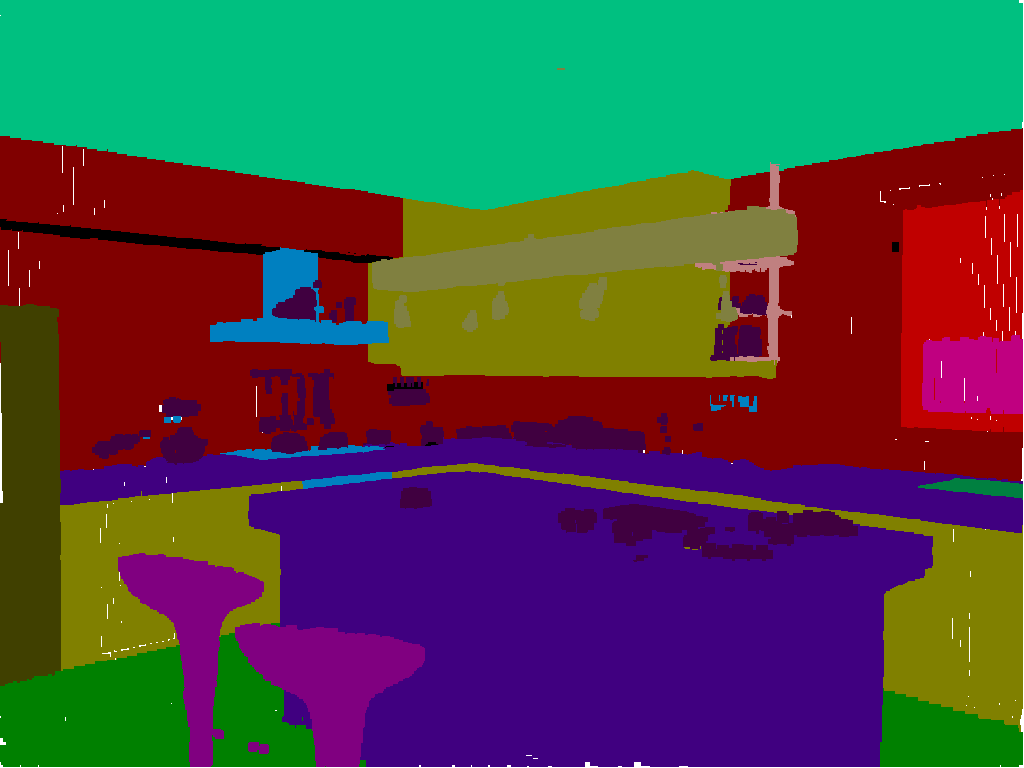}};%
        \node[right=2mm of gt_pmtsdf_semantic_2d] (gt_pmtsdf_instance_2d) {\includeimagedd{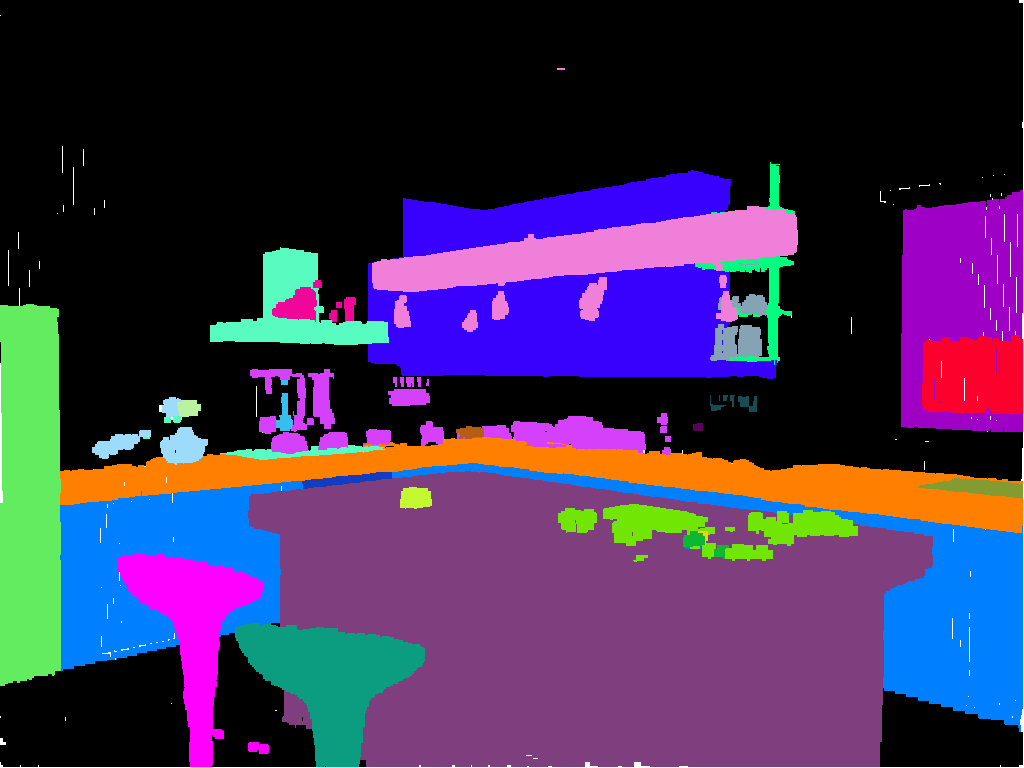}};%
        \node[below=2mm of gt_pmtsdf_panoptic] (pred_pmtsdf_panoptic) {\includeimageddd{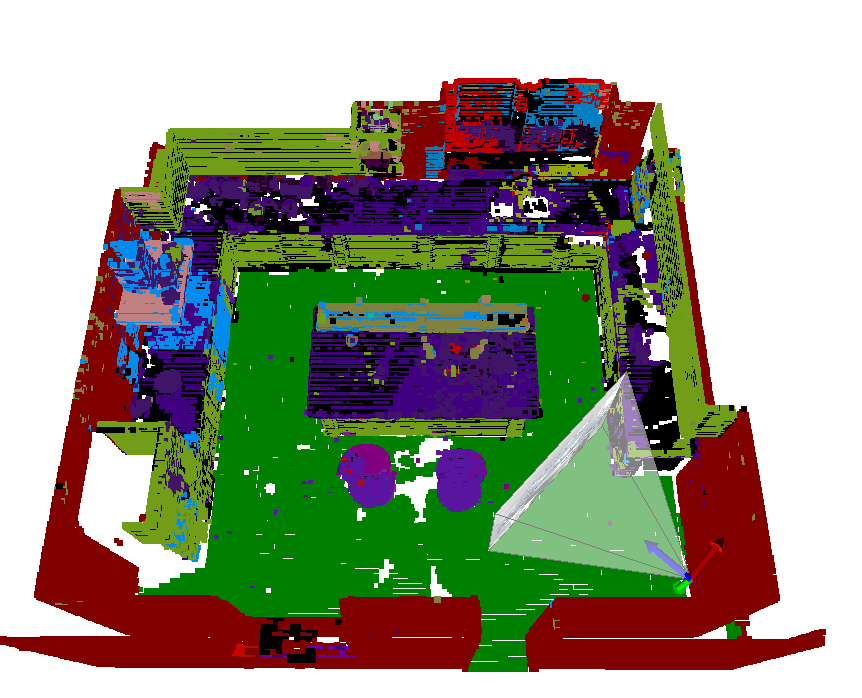}};%
        \node[right=-0.5mm of pred_pmtsdf_panoptic] (pred_pmtsdf_semantic) {\includeimageddd{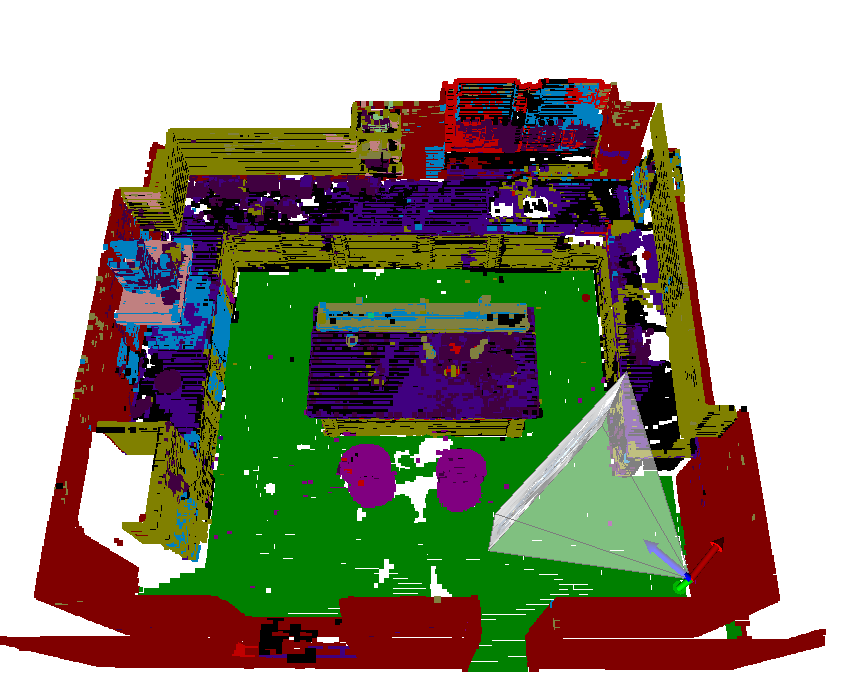}};%
        \node[right=-0.5mm of pred_pmtsdf_semantic] (pred_pmtsdf_instance) {\includeimageddd{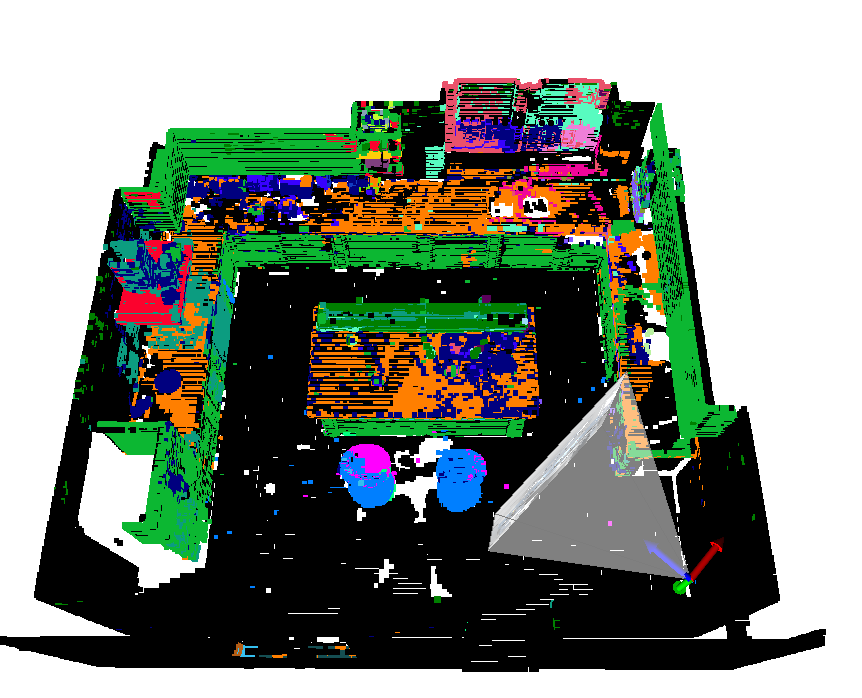}};%
        \node[right=4mm of pred_pmtsdf_instance] (pred_pmtsdf_panoptic_2d) {\includeimagedd{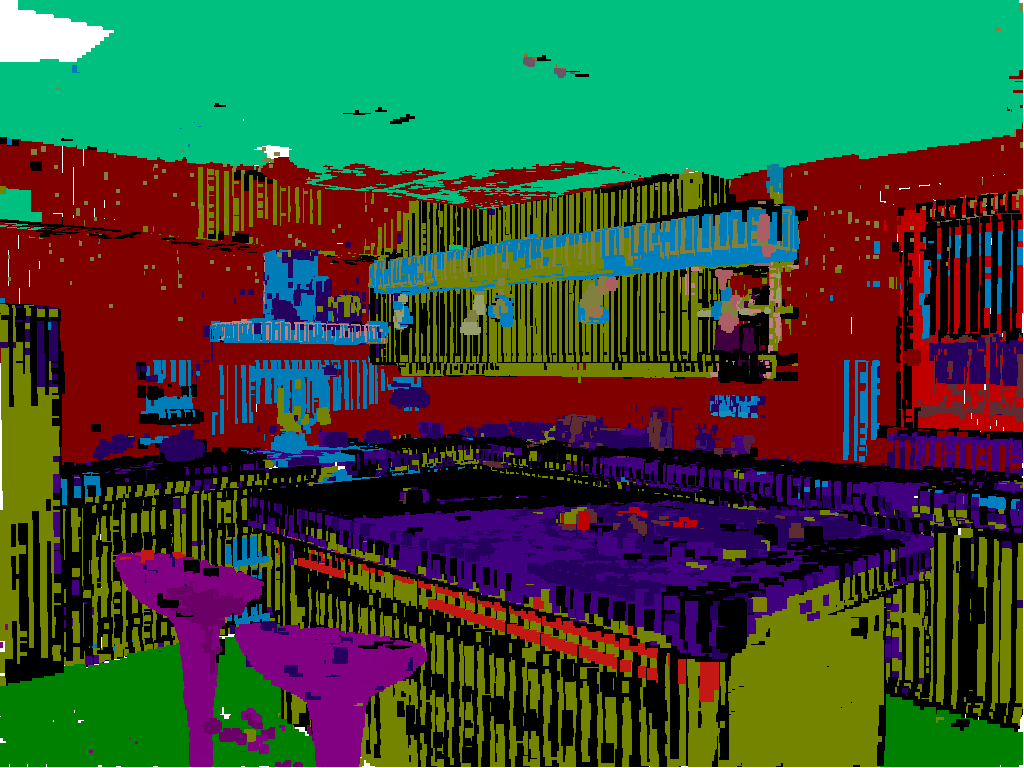}};%
        \node[right=2mm of pred_pmtsdf_panoptic_2d] (pred_pmtsdf_semantic_2d) {\includeimagedd{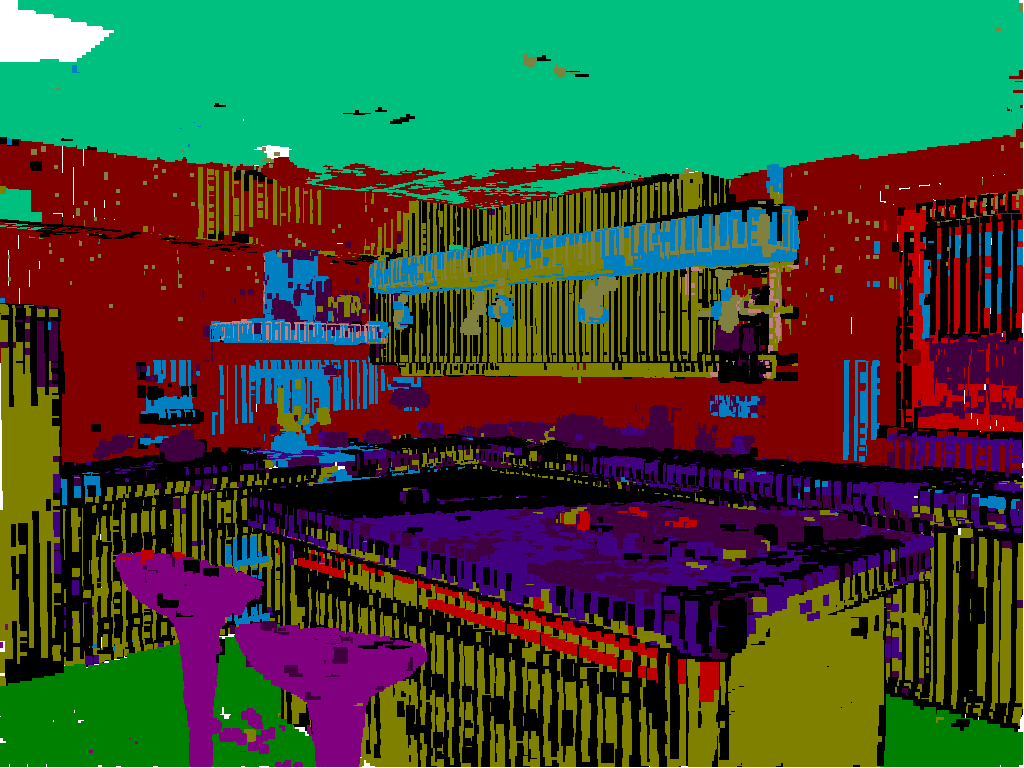}};%
        \node[right=2mm of pred_pmtsdf_semantic_2d] (pred_pmtsdf_instance_2d) {\includeimagedd{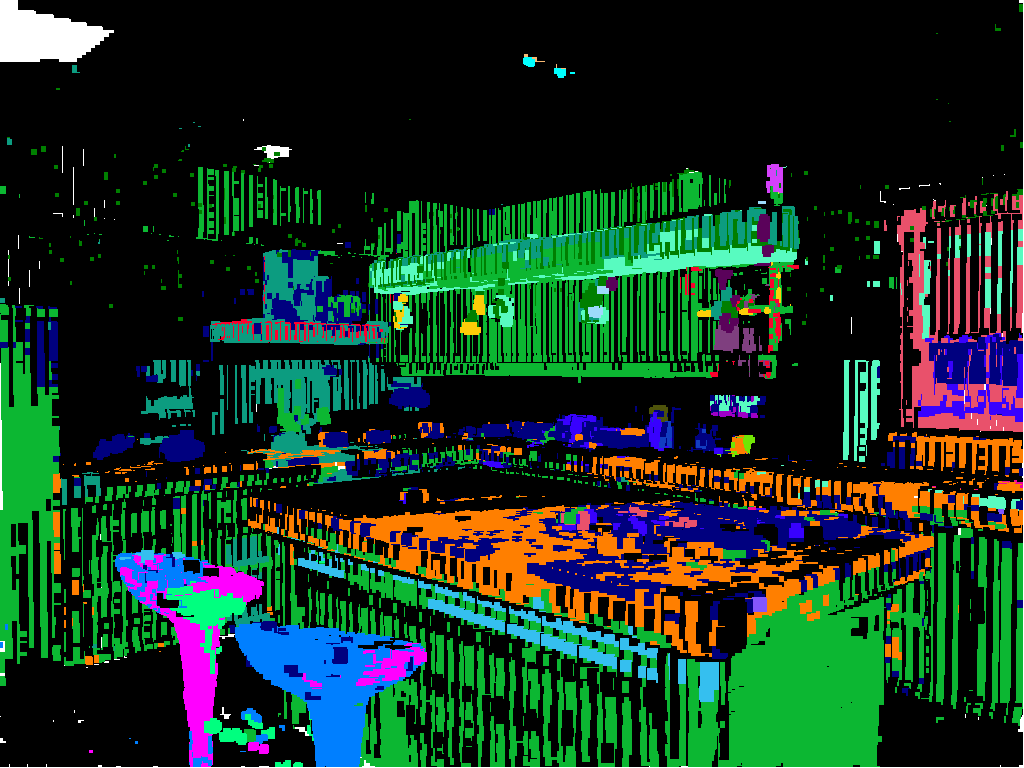}};%
        \node[above=5mm of gt_ndt_05_panoptic_2d] (emsanet_panoptic) {\includeimagedd{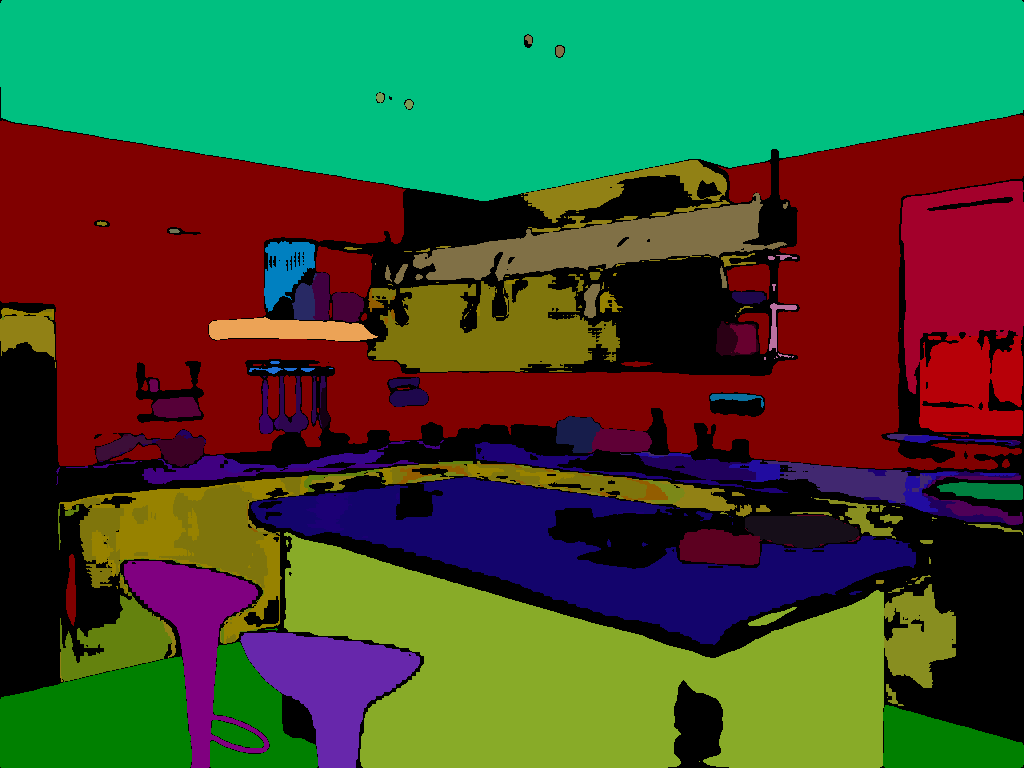}};%
        \node[above=5mm of gt_ndt_05_semantic_2d] (emsanet_semantic) {\includeimagedd{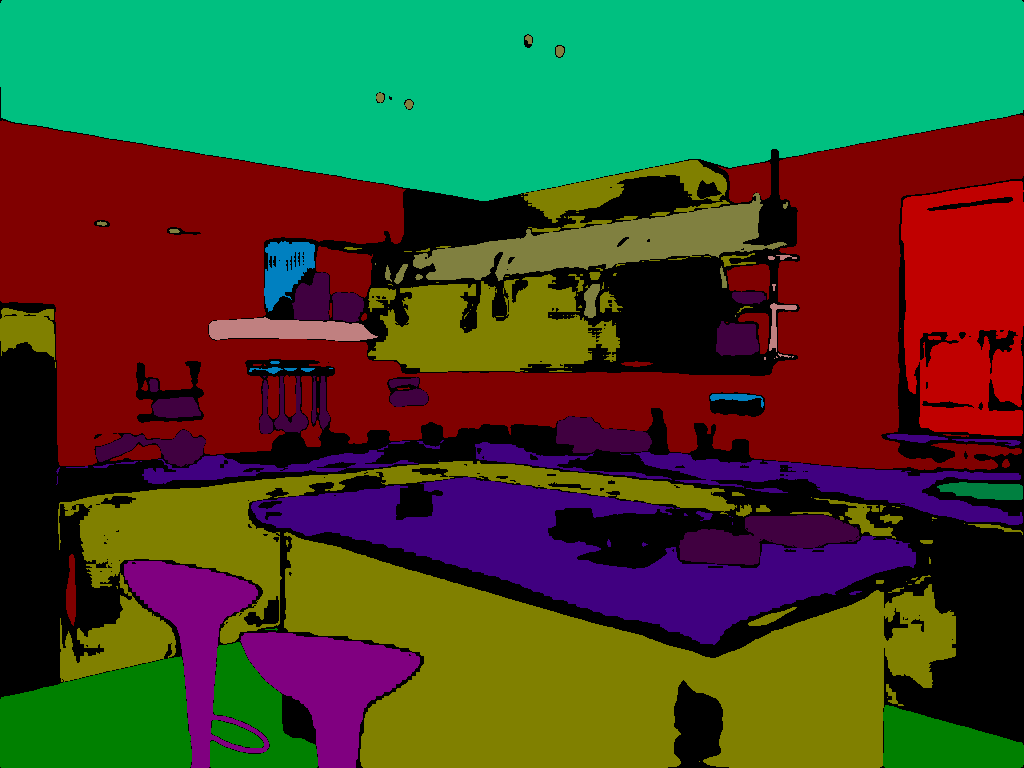}};%
        \node[above=5mm of gt_ndt_05_instance_2d] (emsanet_instance) {\includeimagedd{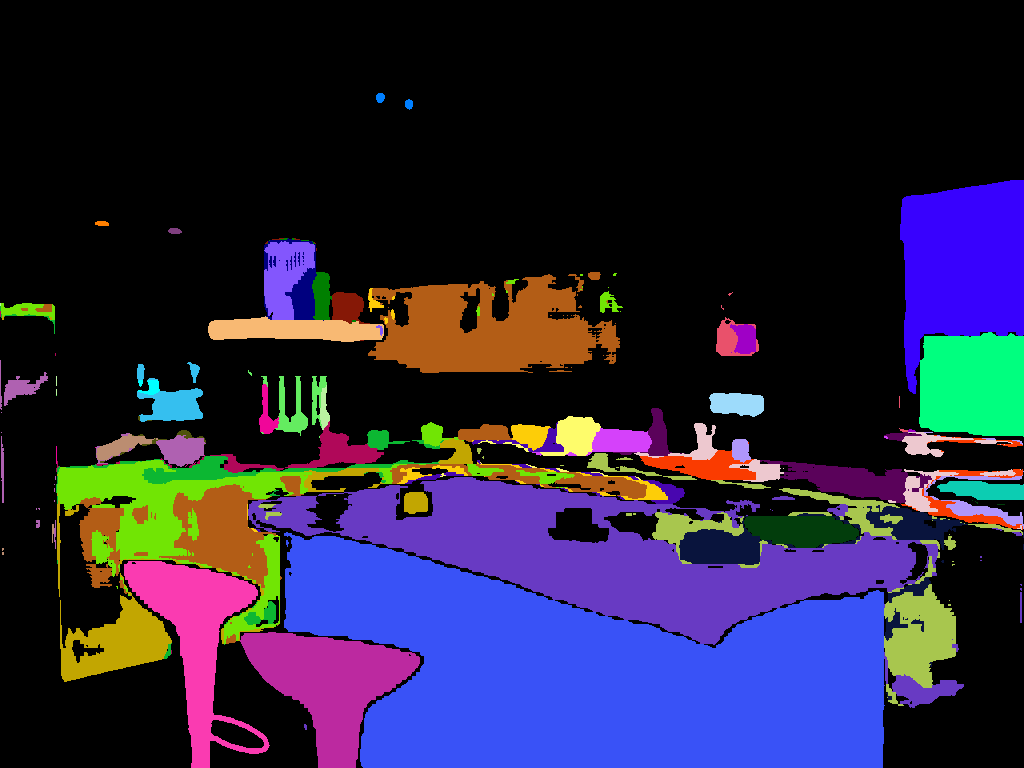}};%
        \node[above=2mm of emsanet_panoptic] (gt_panoptic) {\includeimagedd{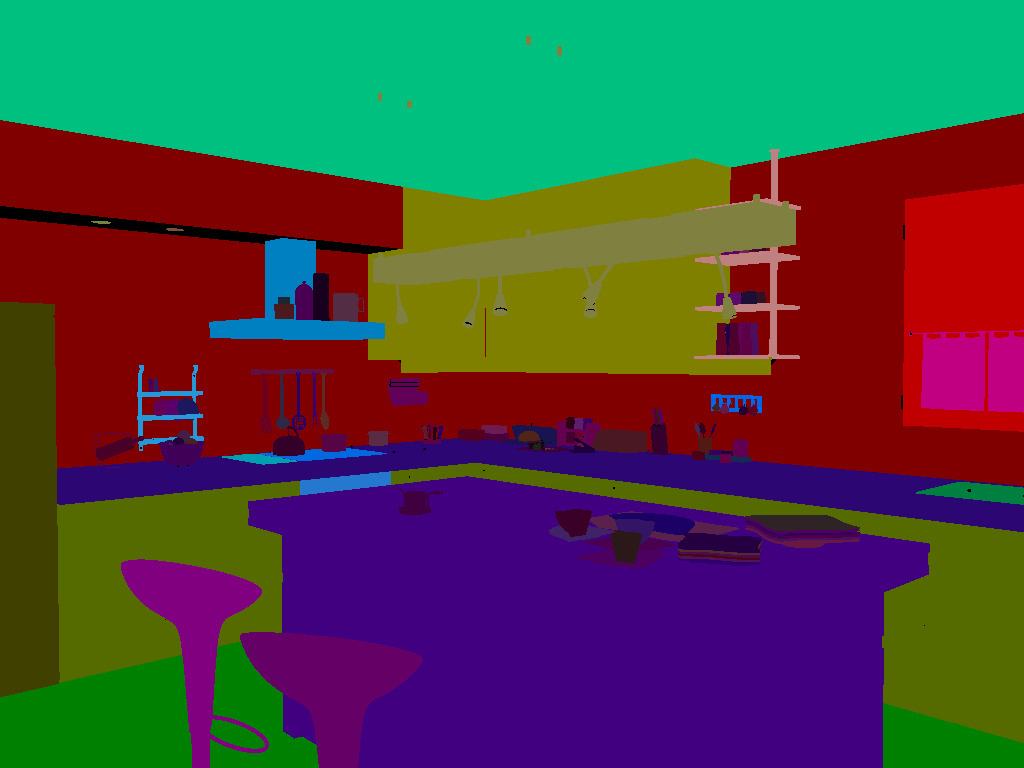}};%
        \node[above=2mm of emsanet_semantic] (gt_semantic) {\includeimagedd{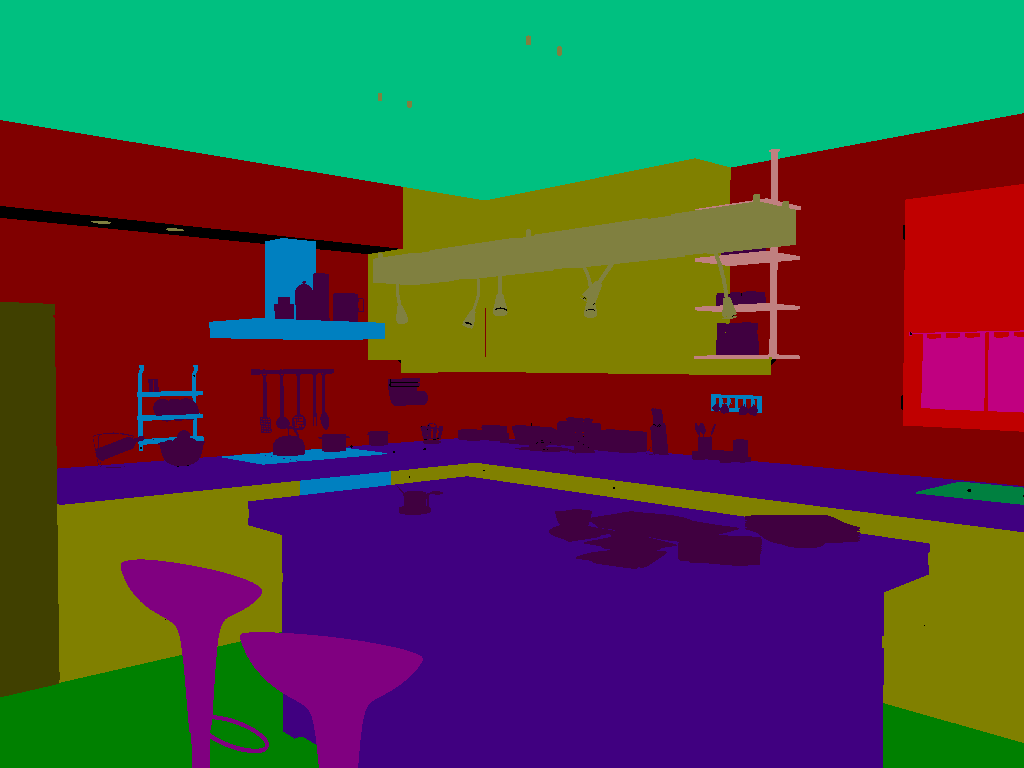}};%
        \node[above=2mm of emsanet_instance] (gt_instance) {\includeimagedd{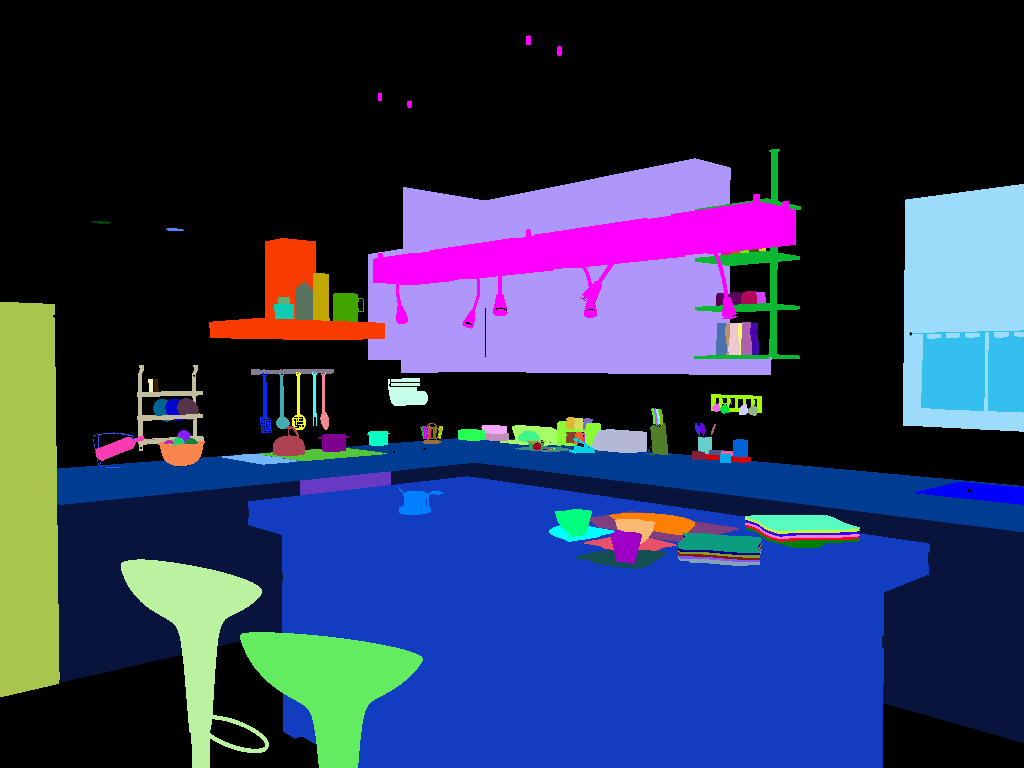}};%
        \node at (gt_ndt_05_panoptic |- gt_panoptic) (rgb) {\includeimagedd{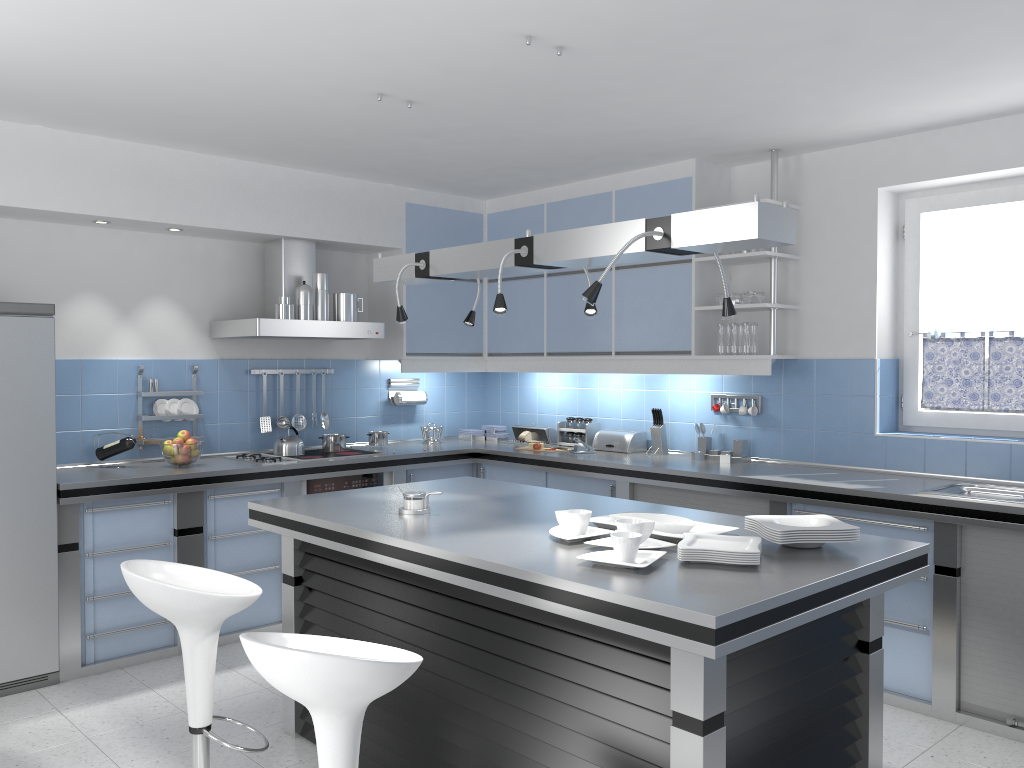}};%
        \node at (gt_ndt_05_semantic |- gt_panoptic) (depth) {\includeimagedd{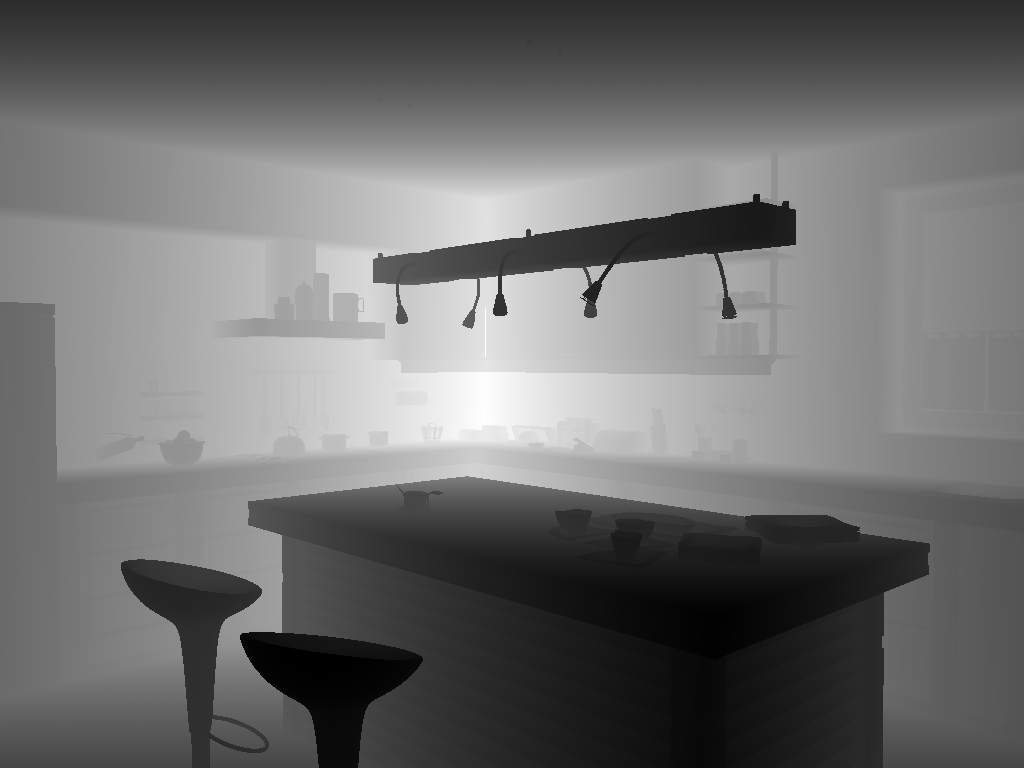}};%
        \node[below=2.5mm of pred_pmtsdf_instance_2d, anchor=mid] (l) {\scriptsize 2D instance};%
        \node[anchor=mid] at (l -| pred_pmtsdf_semantic_2d){\scriptsize 2D semantic};%
        \node[anchor=mid] at (l -| pred_pmtsdf_panoptic_2d){\scriptsize 2D panoptic};%
        \node[anchor=mid] at (l -| pred_pmtsdf_instance){\scriptsize 3D instance};%
        \node[anchor=mid] at (l -| pred_pmtsdf_semantic){\scriptsize 3D semantic};%
        \node[anchor=mid] at (l -| pred_pmtsdf_panoptic){\scriptsize 3D panoptic};%
        \node[below=2mm of rgb, anchor=mid] {\scriptsize RGB};%
        \node[below=2mm of depth, anchor=mid] {\scriptsize Depth};%
        \node[left=2.5mm of gt_panoptic, rotate=90, anchor=mid] {\scriptsize GT};%
        \node[left=2.5mm of emsanet_panoptic, rotate=90, anchor=mid] {\scriptsize EMSANet};%
        \node[left=0.5mm of gt_ndt_05_panoptic, rotate=90, anchor=mid] {\scriptsize GT};%
        \node[left=0.5mm of pred_ndt_05_panoptic, rotate=90, anchor=mid] {\scriptsize EMSANet};%
        \node[left=0.5mm of gt_ndt_10_panoptic, rotate=90, anchor=mid] {\scriptsize GT};%
        \node[left=0.5mm of pred_ndt_10_panoptic, rotate=90, anchor=mid] {\scriptsize EMSANet};%
        \node[left=0.5mm of gt_pmtsdf_panoptic, rotate=90, anchor=mid] {\scriptsize GT};%
        \node[left=0.5mm of pred_pmtsdf_panoptic, rotate=90, anchor=mid] {\scriptsize EMSANet};%
        \node[rotate=90, anchor=mid] at ([xshift=-5.5mm]$(gt_ndt_05_panoptic.west)!0.5!(pred_ndt_05_panoptic.west)$) (l) {\footnotesize PanopticNDT~(5cm)};%
        \node[rotate=90, anchor=mid] at ([xshift=-5.5mm]$(gt_ndt_10_panoptic.west)!0.5!(pred_ndt_10_panoptic.west)$) {\footnotesize\bf PanopticNDT~(10cm)};%
        \node[rotate=90, anchor=mid] at ([xshift=-5.5mm]$(gt_pmtsdf_panoptic.west)!0.5!(pred_pmtsdf_panoptic.west)$) {\footnotesize Panoptic Multi-TSDFs};%
        \coordinate (p) at ($(gt_panoptic)!0.5!(emsanet_panoptic)$);%
        \node[rotate=90, anchor=mid] at (l |- p) (lsf) {\footnotesize Single frame (given camera pose)};%
        \coordinate (p) at ([yshift=-2.5mm]lsf.north west |- emsanet_panoptic.south west);%
        \draw[gray!80, dashed, thick] (p) -- (p -| emsanet_instance.south east);%
        \draw[gray!80, dashed, thick] ([yshift=-2.5mm]pred_ndt_05_panoptic.south west) -- ([yshift=-2.5mm]pred_ndt_05_instance_2d.south east);%
        \draw[gray!80, dashed, thick] ([yshift=-2.5mm]pred_ndt_10_panoptic.south west) -- ([yshift=-2.5mm]pred_ndt_10_instance_2d.south east);%
    \end{tikzpicture}%
    \vspace{-3mm}%
    \caption{%
        Qualitative results for scene \emph{ai\_001\_10} of the Hypersim test split. %
        The upper part shows the dataset's ground truth as well as the thresholded predictions of EMSANet~\cite{emsanet2022ijcnn}~(see Sec.~\ref{sec:experiments:implementation}). %
        The lower part compares our proposed PanopticNDT with voxel sizes 5\si{\centi\meter} and 10\si{\centi\meter} to Panoptic Multi-TSDFs~\cite{panoptic-multi-tsdf-2022-icra}. %
        For each mapping approach, results are visualized for both when mapping with ground truth~(top) and when mapping with predicted segmentation of EMSANet~(bottom).
        Best viewed in color at 300\%. %
        Black indicates \emph{void/no\_instance}, for the semantic colors, we refer to Fig.~\ref{fig:experiments:hypersim_radar_chart}. %
        Panoptic labels are visualized by small color differences based on the semantic color.%
    }        
    \label{fig:appendix:qualitative_results_hypersim}
\end{figure*}
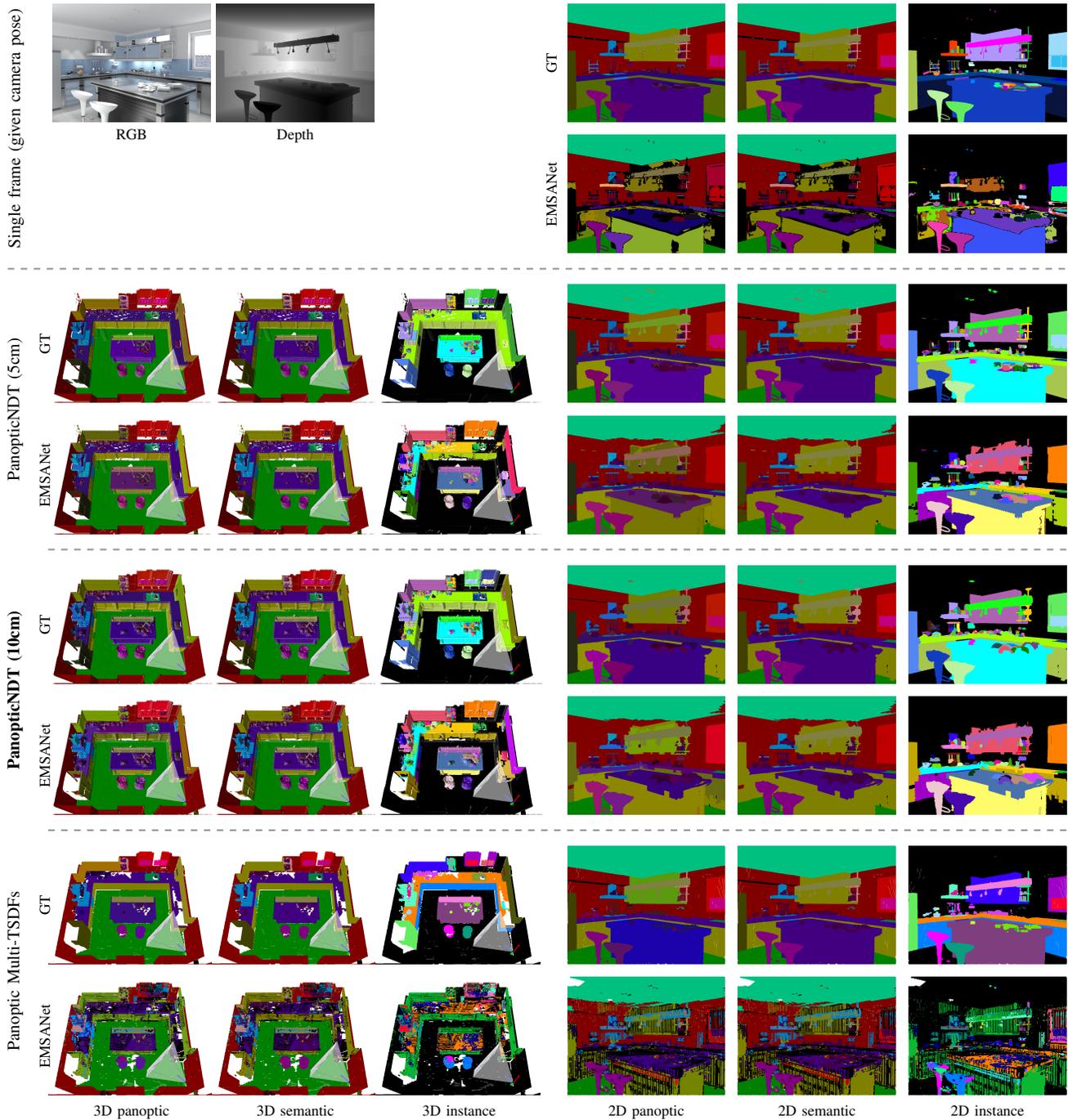
The lower part of Tab.~\ref{tab:emsanet_application} further presents results when fine-tuning the application network on the individual datasets.
Again, especially the results for the real-world datasets NYUv2 and SUNRGB-D indicate that the fine-tuned network has much stronger generalization capabilities. 
For Hypersim and ScanNetV2, the results are close to those listed in Tab.~\ref{tab:results_hypersim} and Tab.~\ref{tab:results_scannet} in the main part. 
We share the weights for all networks in our GitHub repository.

The hyperparameters for PanopticNDT in real-world applications follow the values described in Sec.~\ref{sec:experiments:implementation} for the ScanNetV2 experiments.
The only exception are the IoU thresholds for instance matching~(see Eq.~\ref{eq:main:matching}). 
For application, we use slightly higher values for $\theta^{\text{M}}$ and~$\theta^{\text{N}}$ and set them to 0.3 and 0.2, respectively.
For the environment shown in the video, mapping runs at 6.89\si{\hertz} on average~(the actual map update rate slightly varies depending on the captured part of a scene) on the hardware of our mobile 
robot~(Intel NUC11PHKi7C~(Intel i7-1165G7)). 
Incorporating panoptic information almost halves the throughput compared to SemanticNDT~\cite{semanticmapping2022icra}~(12.29\si{\hertz}) and almost cuts the throughput into thirds compared to plain NDT of~\cite{Einhorn-ECMR-2013-NDT, Einhorn-PHD-2019-NDT}~(18.08\si{\hertz}).
Due to the voxel size of 10\si{\centi\meter}, the resulting panoptic NDT map for the entire flat shown in the video is only 13.5\si{\mega\byte}. 
Storing panoptic information increases the map size by \textasciitilde53\% and \textasciitilde255\% compared to semantic NDT maps of~\cite{semanticmapping2022icra}~(8.8\si{\mega\byte}) and plain NDT maps of~\cite{Einhorn-ECMR-2013-NDT, Einhorn-PHD-2019-NDT}~(3.8\si{\mega\byte}).%

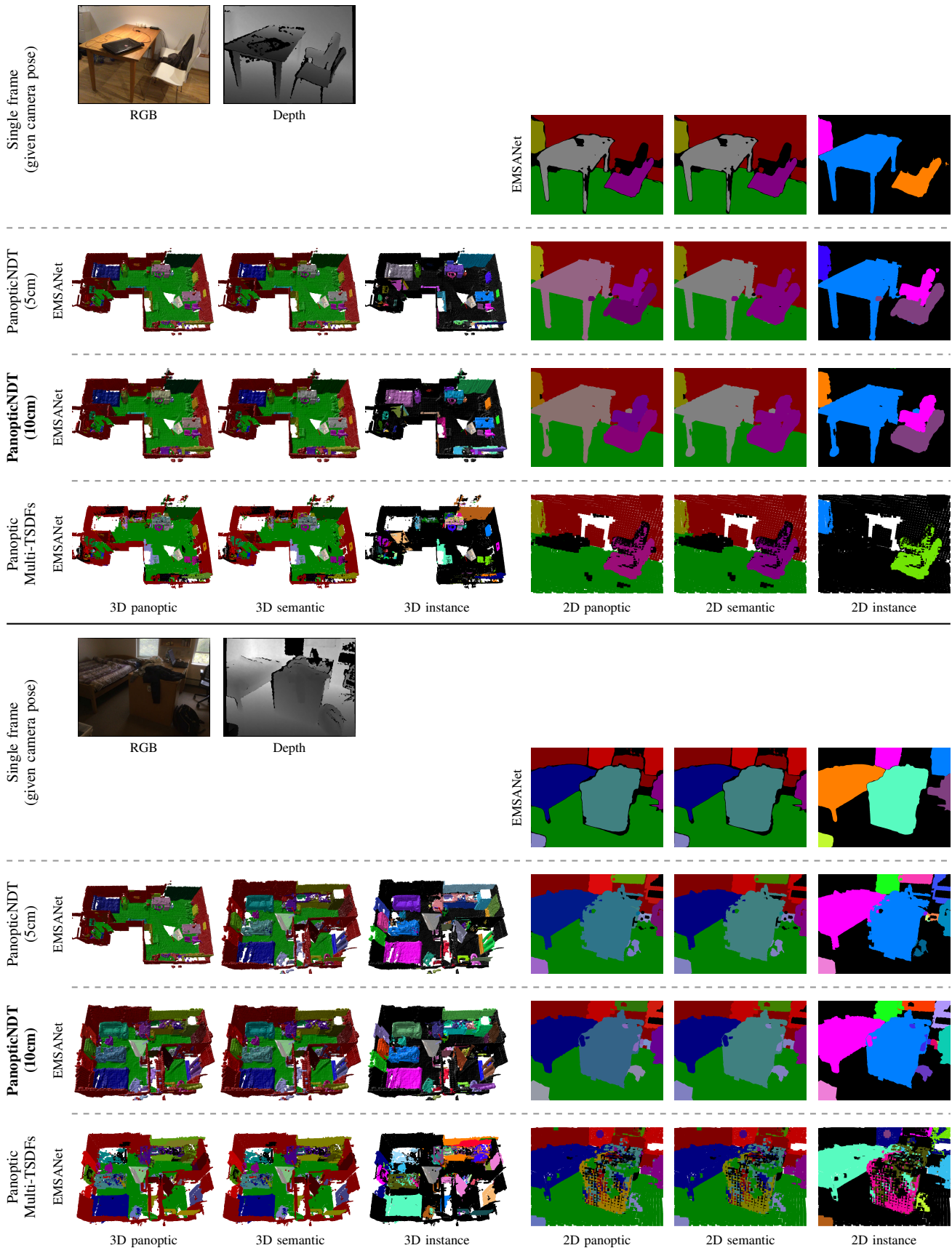
\begin{figure*}[!b]
    \centering 
    \begin{tikzpicture}[inner sep=0pt]     
        \newcommand{\includeimageddd}[1]{\includegraphics[height=1.8cm, trim={0 0.5cm 0 3.2cm}, clip]{#1}}%
        \newcommand{\includeimagedd}[1]{\includegraphics[height=1.8cm]{#1}}%
        \node (pred_ndt_05_panoptic) {\includeimageddd{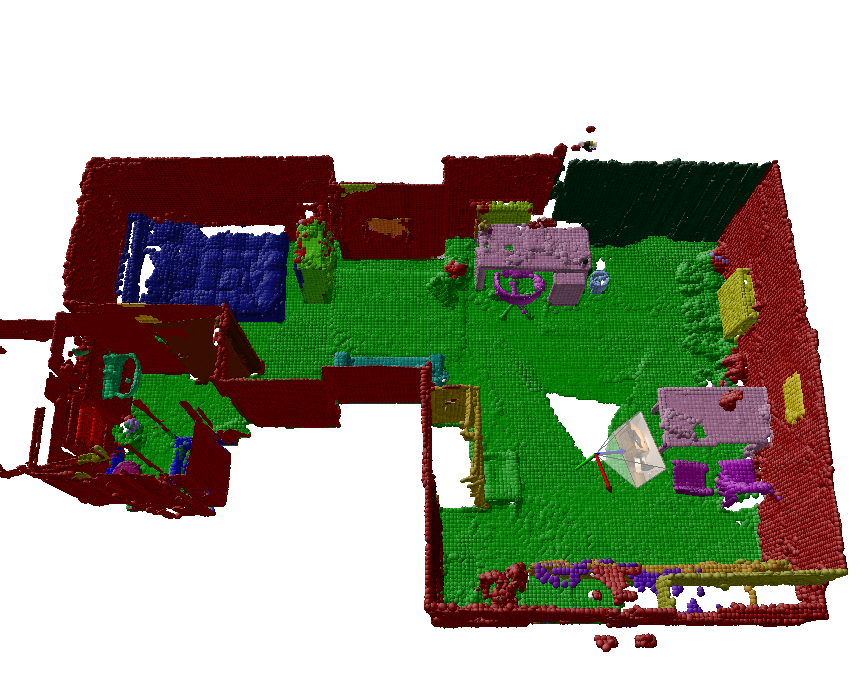}};%
        \node[right=0.5mm of pred_ndt_05_panoptic] (pred_ndt_05_semantic) {\includeimageddd{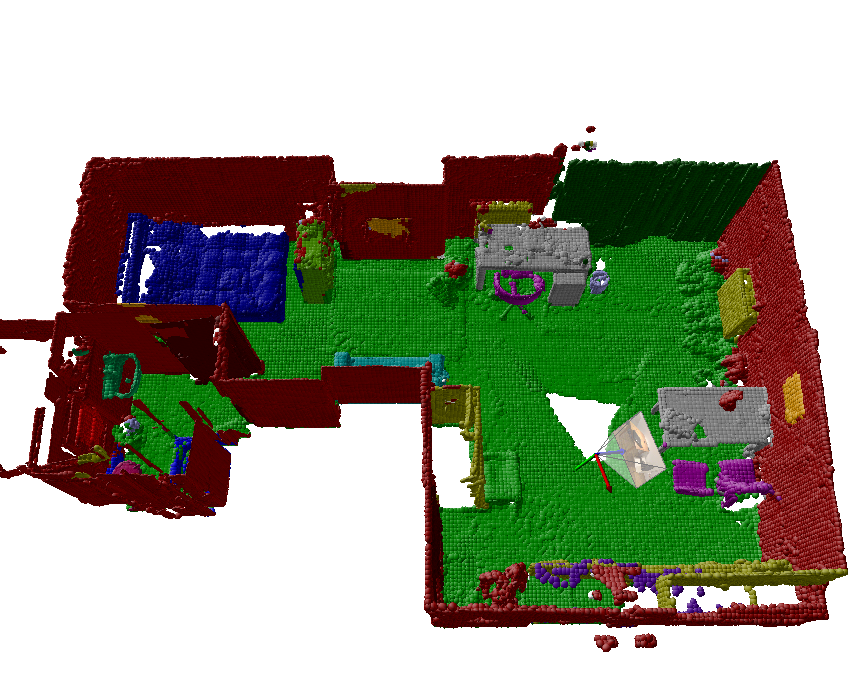}};%
        \node[right=0.5mm of pred_ndt_05_semantic] (pred_ndt_05_instance) {\includeimageddd{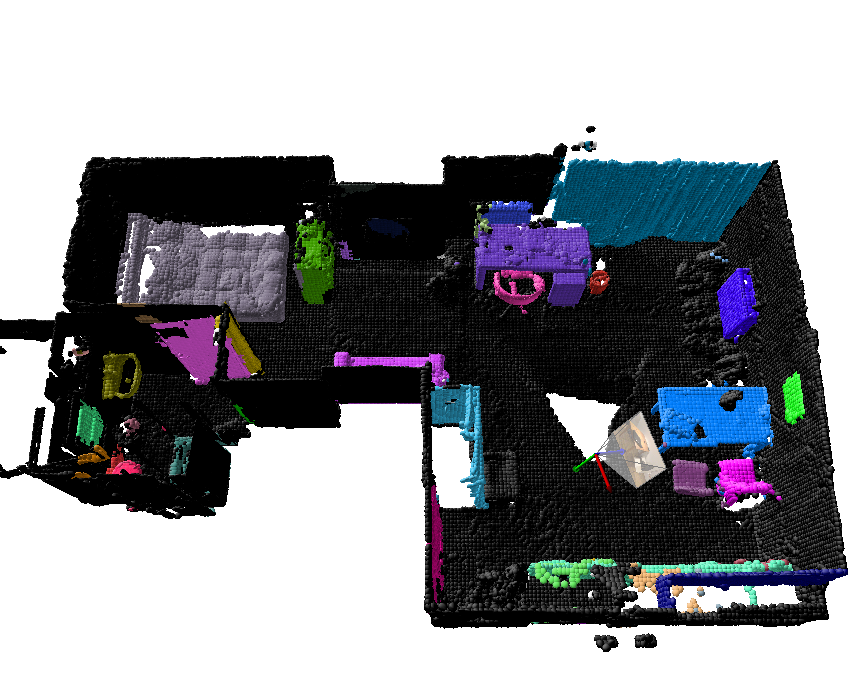}};%
        \node[right=4mm of pred_ndt_05_instance] (pred_ndt_05_panoptic_2d) {\includeimagedd{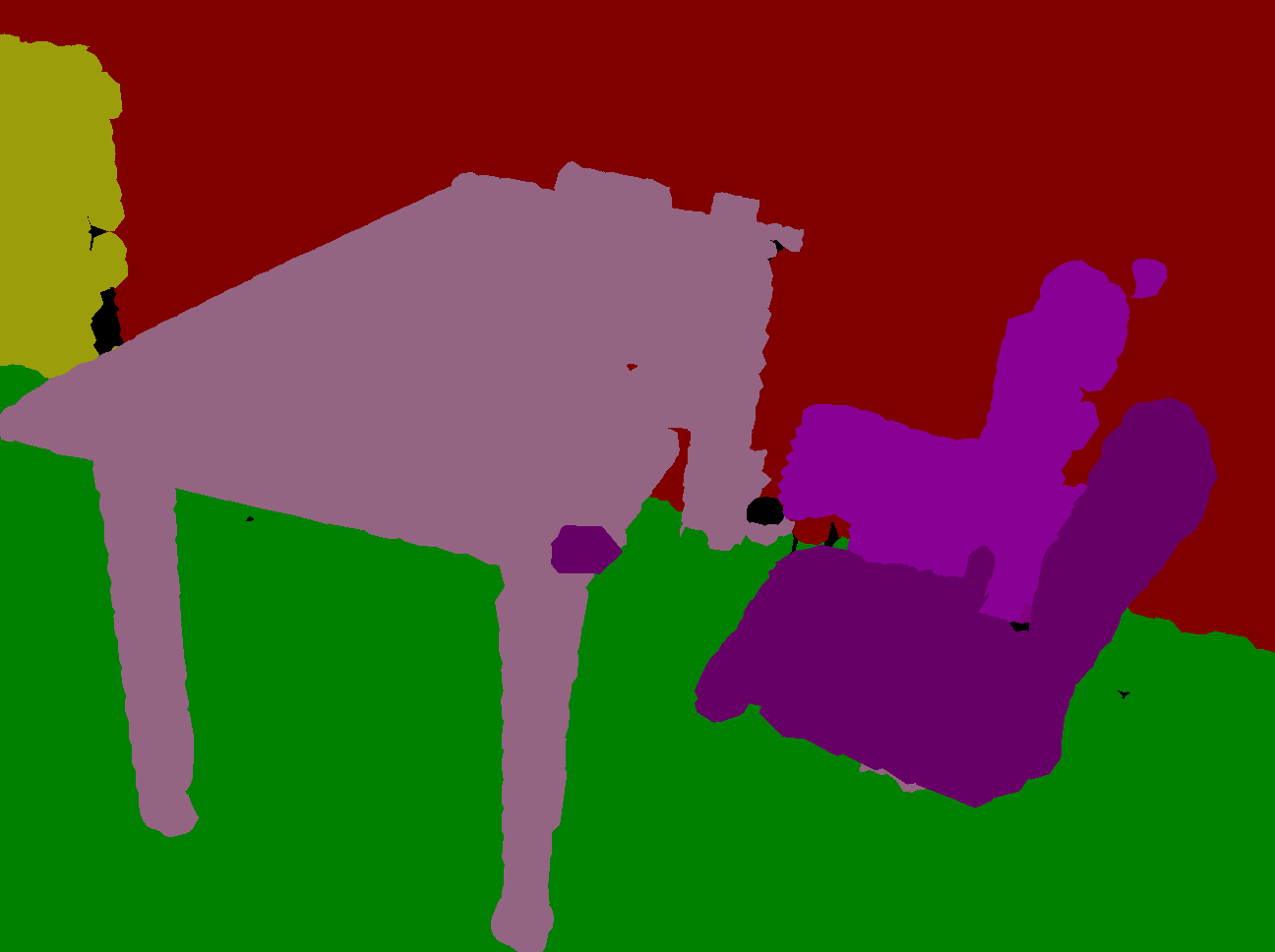}};%
        \node[right=2mm of pred_ndt_05_panoptic_2d] (pred_ndt_05_semantic_2d) {\includeimagedd{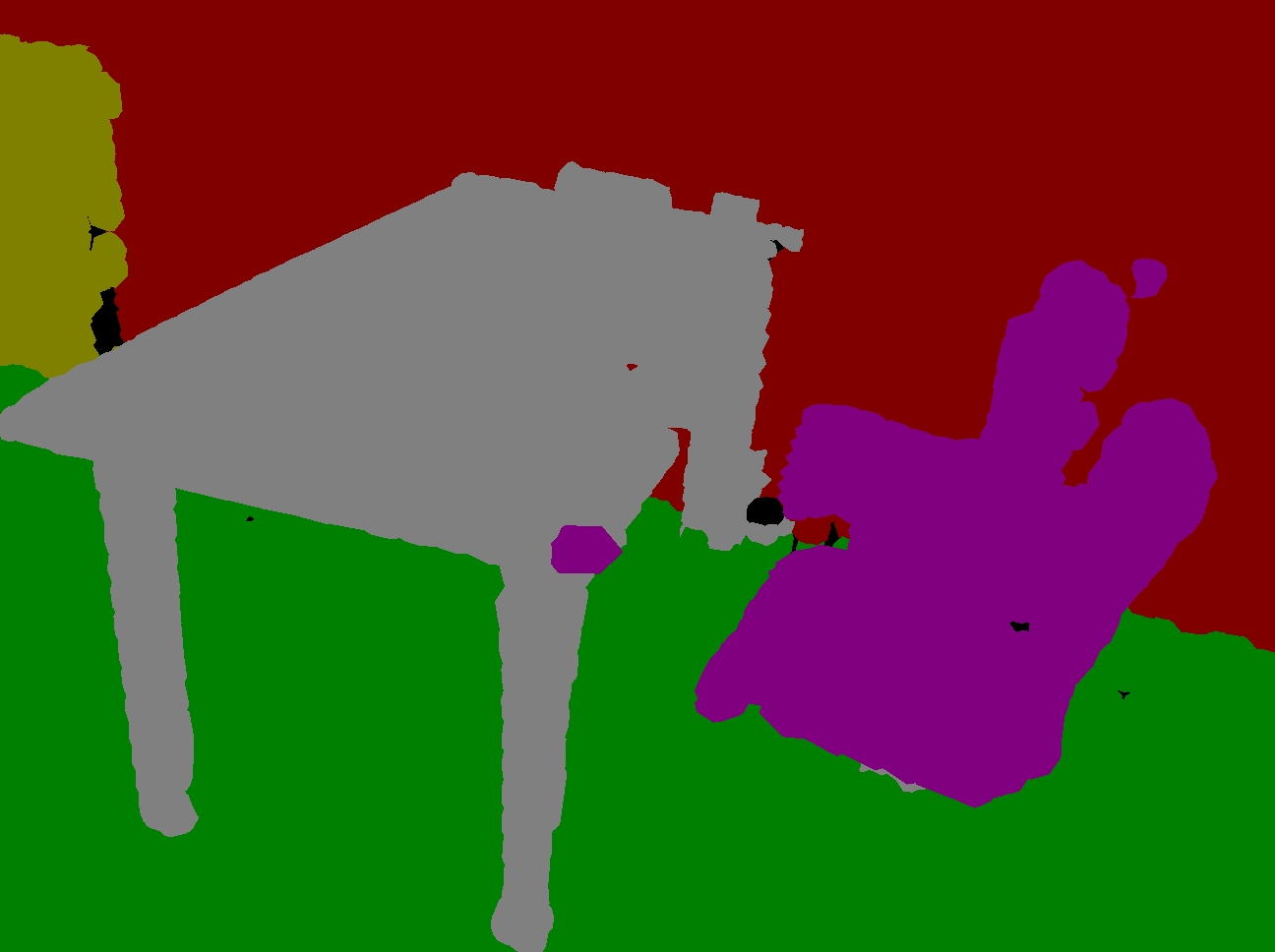}};%
        \node[right=2mm of pred_ndt_05_semantic_2d] (pred_ndt_05_instance_2d) {\includeimagedd{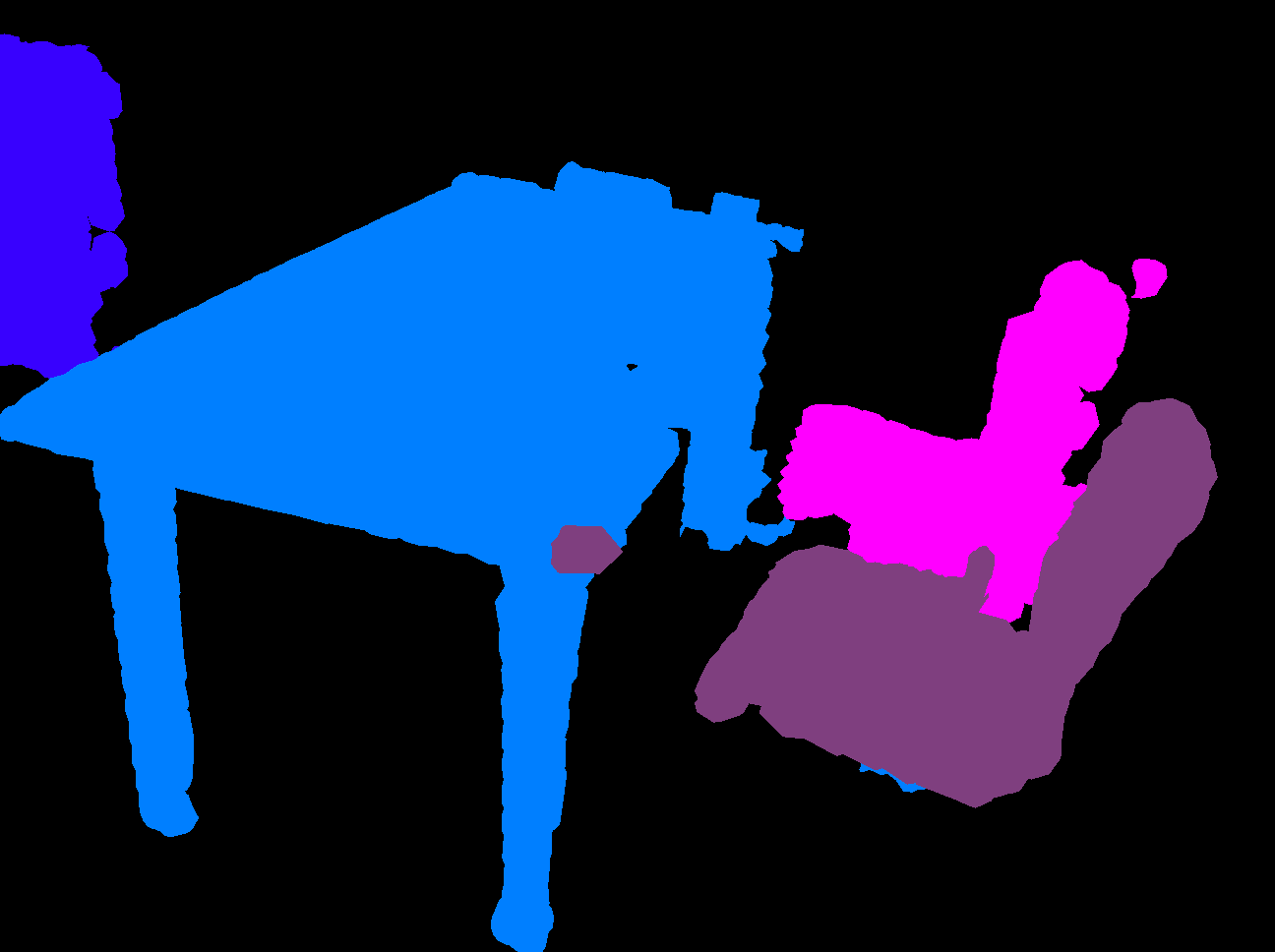}};%
        \node[below=5mm of pred_ndt_05_panoptic] (pred_ndt_10_panoptic) {\includeimageddd{img/examples/scannet_example_0/pred_ndt_10_panoptic.png}};%
        \node[right=0.5mm of pred_ndt_10_panoptic] (pred_ndt_10_semantic) {\includeimageddd{img/examples/scannet_example_0/pred_ndt_10_semantic.png}};%
        \node[right=0.5mm of pred_ndt_10_semantic] (pred_ndt_10_instance) {\includeimageddd{img/examples/scannet_example_0/pred_ndt_10_instance.png}};%
        \node[right=4mm of pred_ndt_10_instance] (pred_ndt_10_panoptic_2d) {\includeimagedd{img/examples/scannet_example_0/pred_ndt_10_backprojection_00435_panoptic.png}};%
        \node[right=2mm of pred_ndt_10_panoptic_2d] (pred_ndt_10_semantic_2d) {\includeimagedd{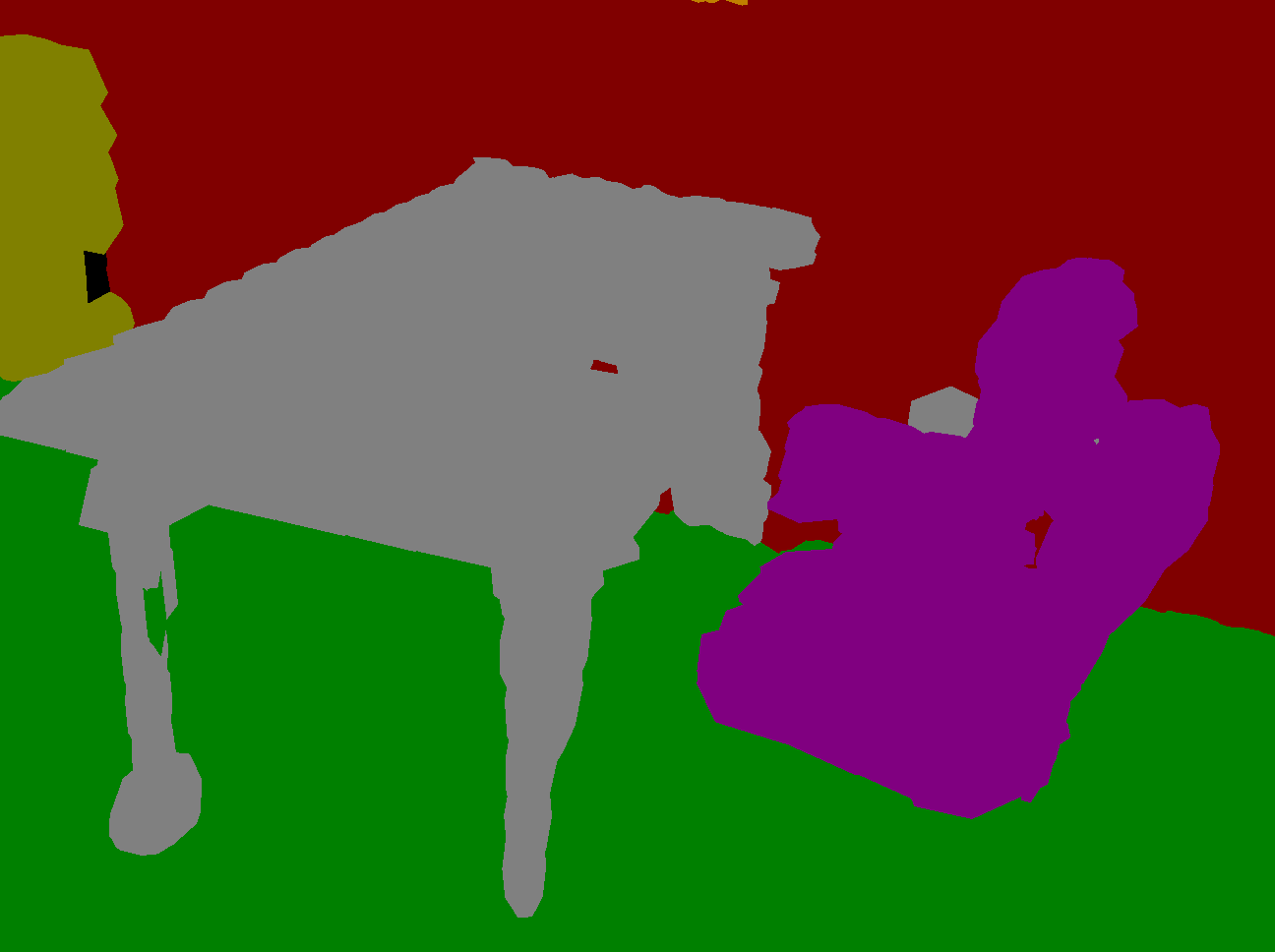}};%
        \node[right=2mm of pred_ndt_10_semantic_2d] (pred_ndt_10_instance_2d) {\includeimagedd{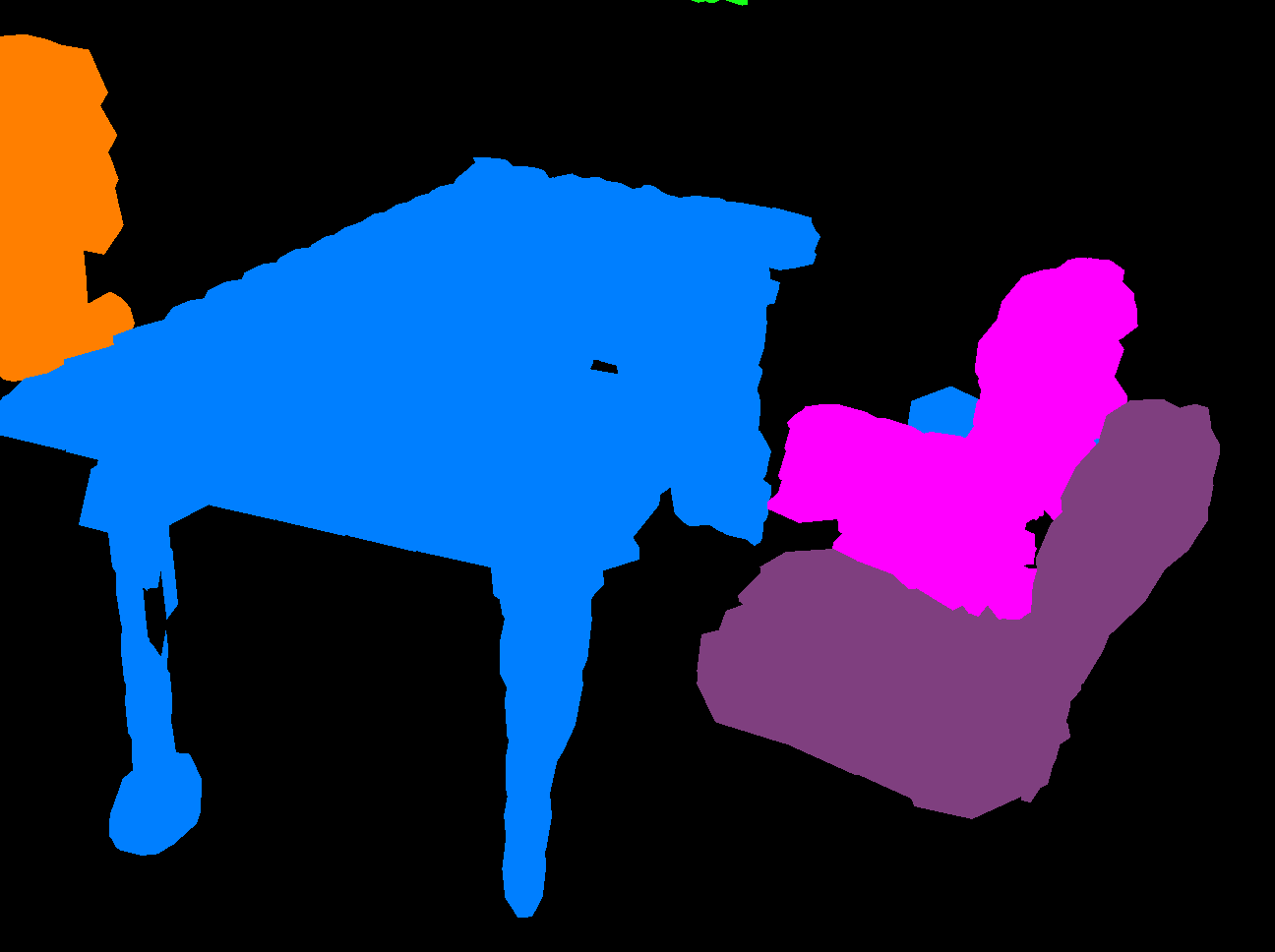}};%
        \node[below=5mm of pred_ndt_10_panoptic] (pred_pmtsdf_panoptic) {\includeimageddd{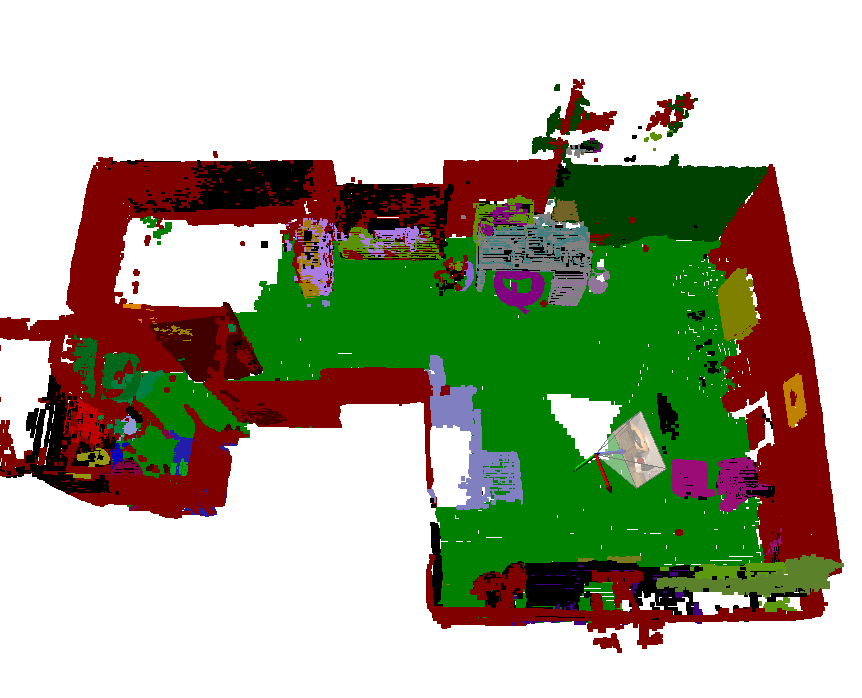}};%
        \node[right=0.5mm of pred_pmtsdf_panoptic] (pred_pmtsdf_semantic) {\includeimageddd{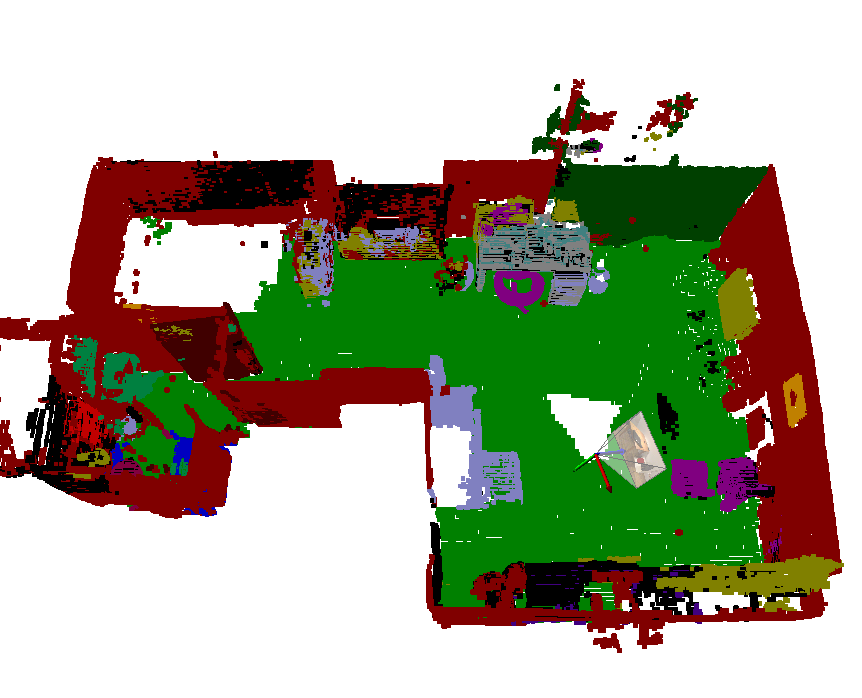}};%
        \node[right=0.5mm of pred_pmtsdf_semantic] (pred_pmtsdf_instance) {\includeimageddd{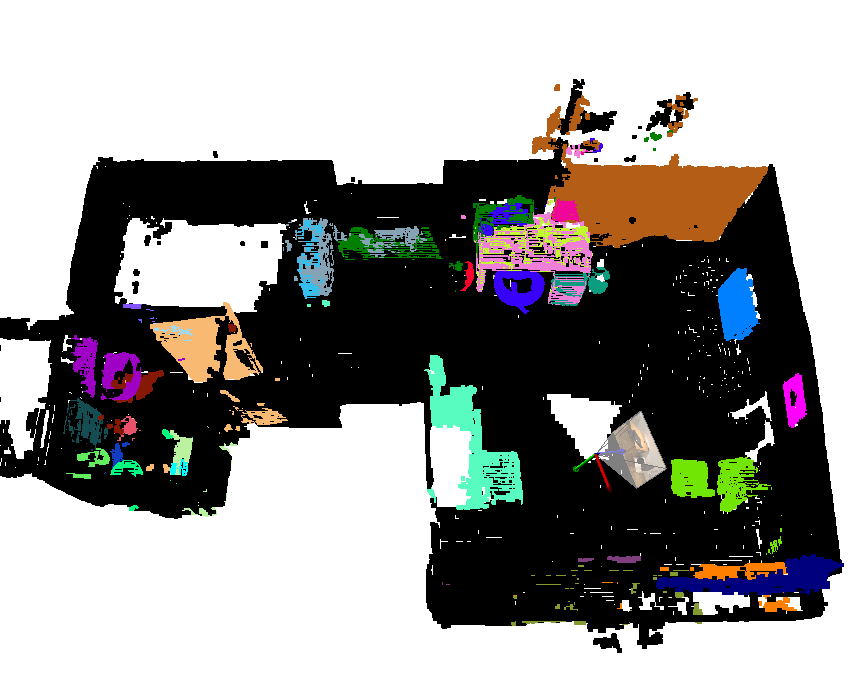}};%
        \node[right=4mm of pred_pmtsdf_instance] (pred_pmtsdf_panoptic_2d) {\includeimagedd{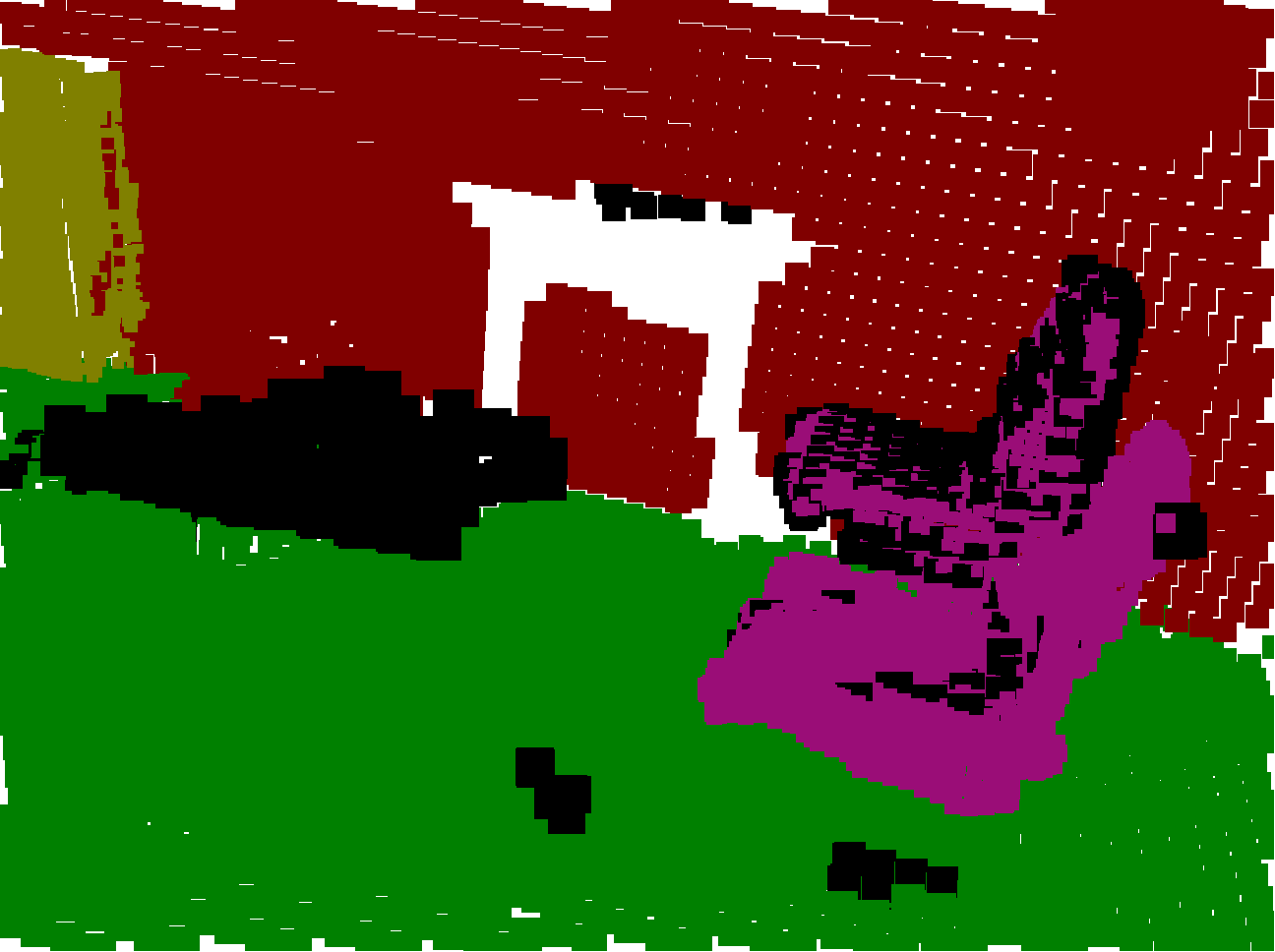}};%
        \node[right=2mm of pred_pmtsdf_panoptic_2d] (pred_pmtsdf_semantic_2d) {\includeimagedd{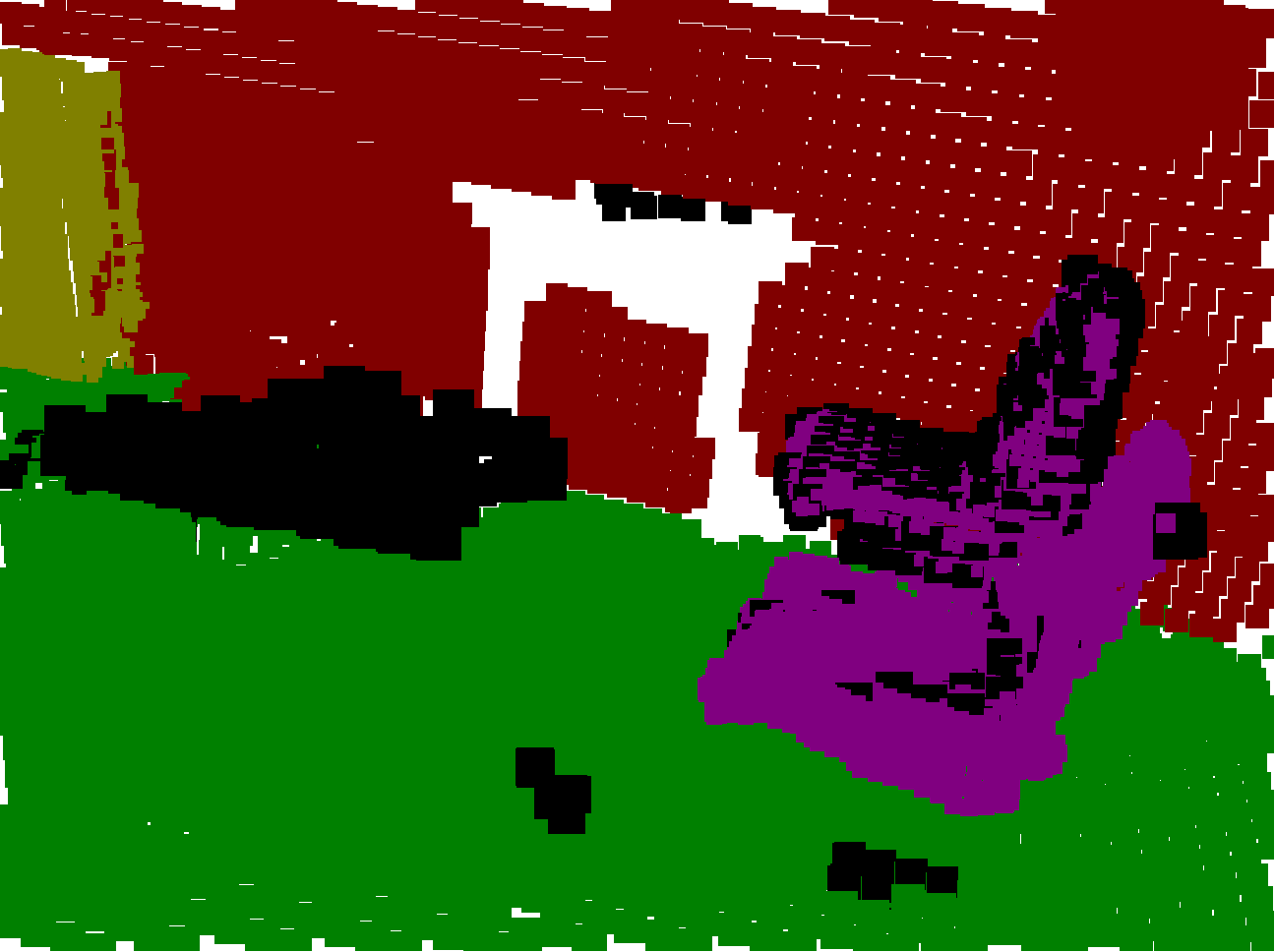}};%
        \node[right=2mm of pred_pmtsdf_semantic_2d] (pred_pmtsdf_instance_2d) {\includeimagedd{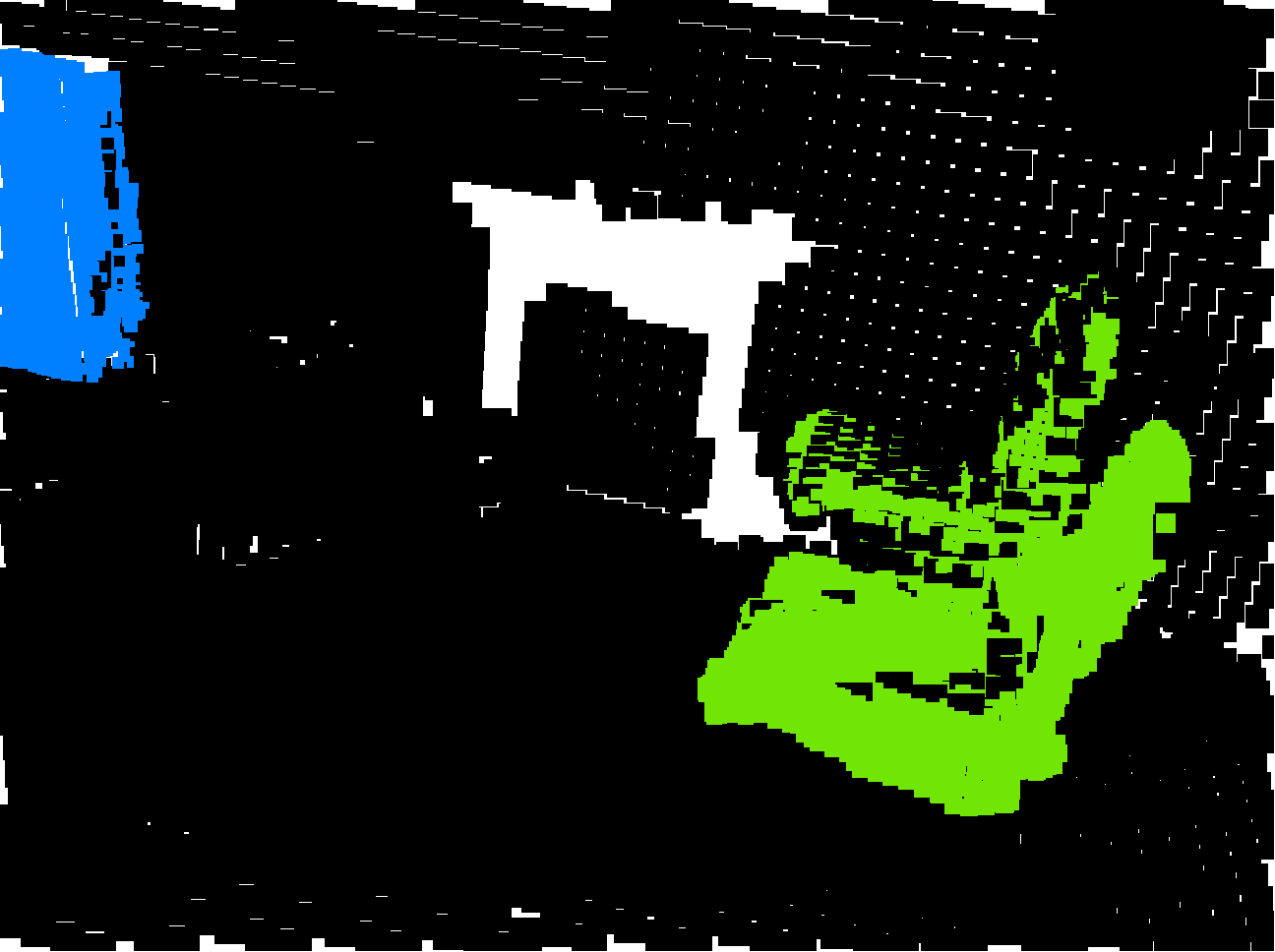}};%
        \node[above=5mm of pred_ndt_05_panoptic_2d] (emsanet_panoptic) {\includeimagedd{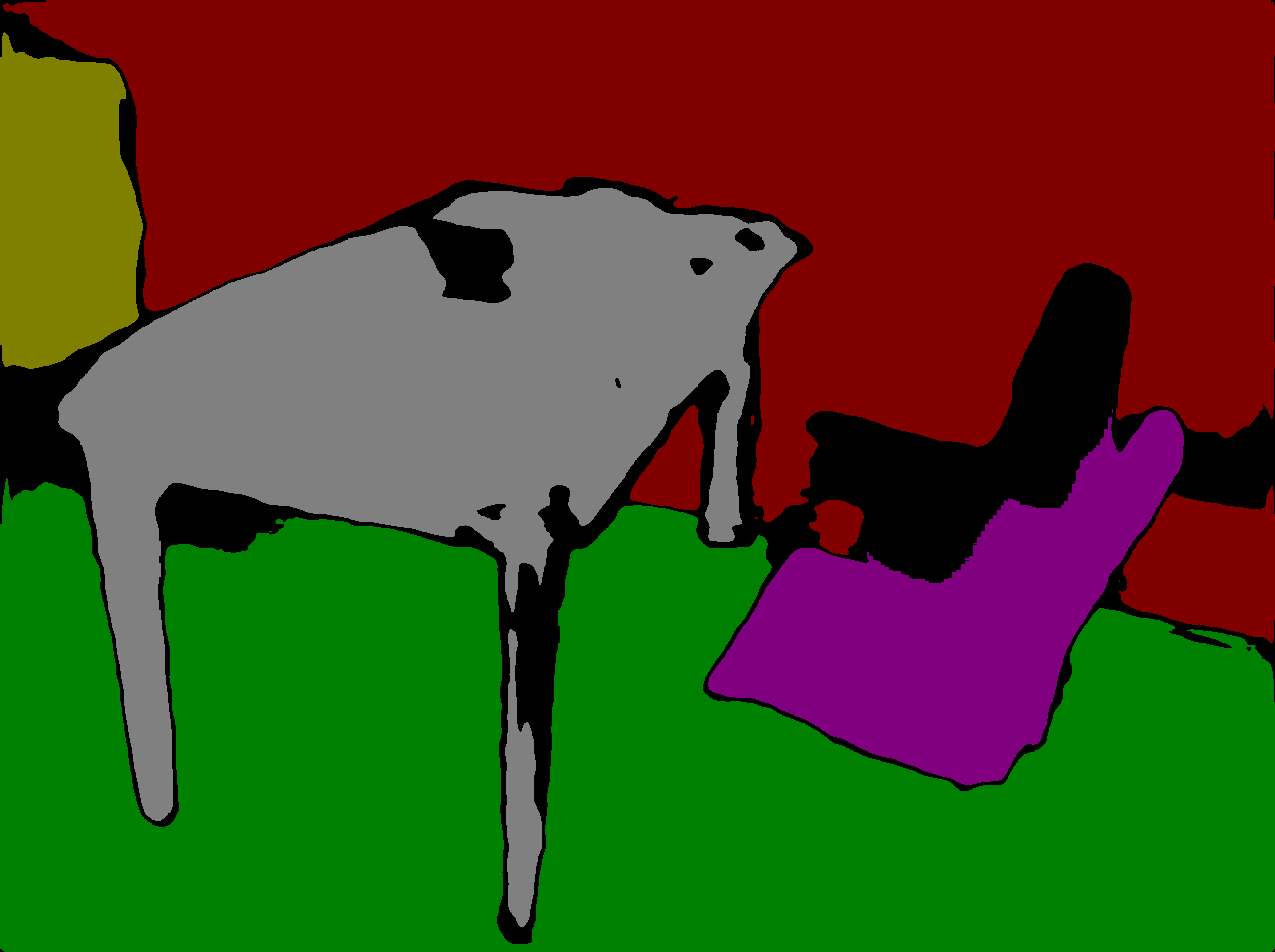}};%
        \node[above=5mm of pred_ndt_05_semantic_2d] (emsanet_semantic) {\includeimagedd{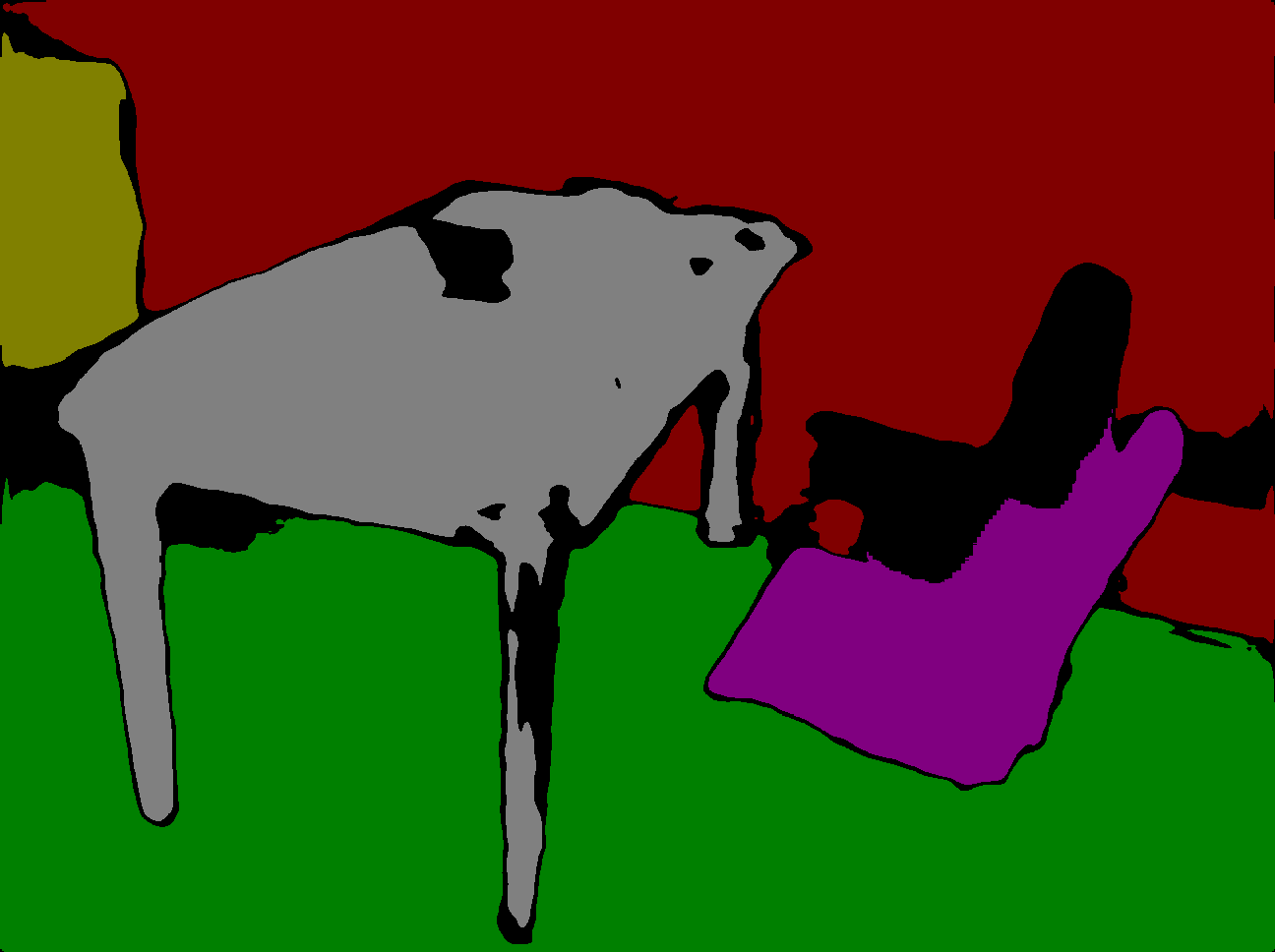}};%
        \node[above=5mm of pred_ndt_05_instance_2d] (emsanet_instance) {\includeimagedd{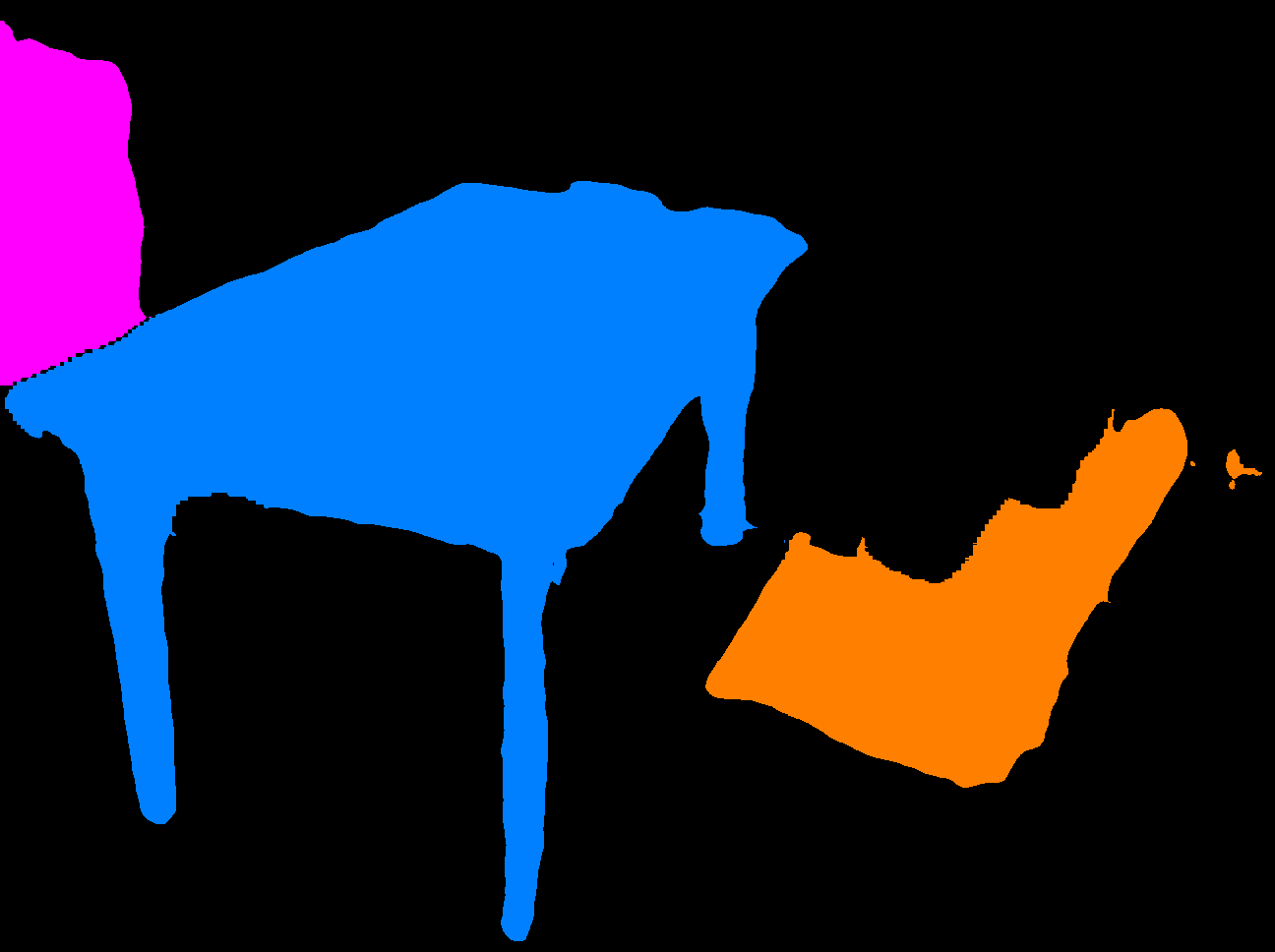}};%
        \node[above=2mm of emsanet_panoptic] (gt_panoptic) {\phantom{\includeimagedd{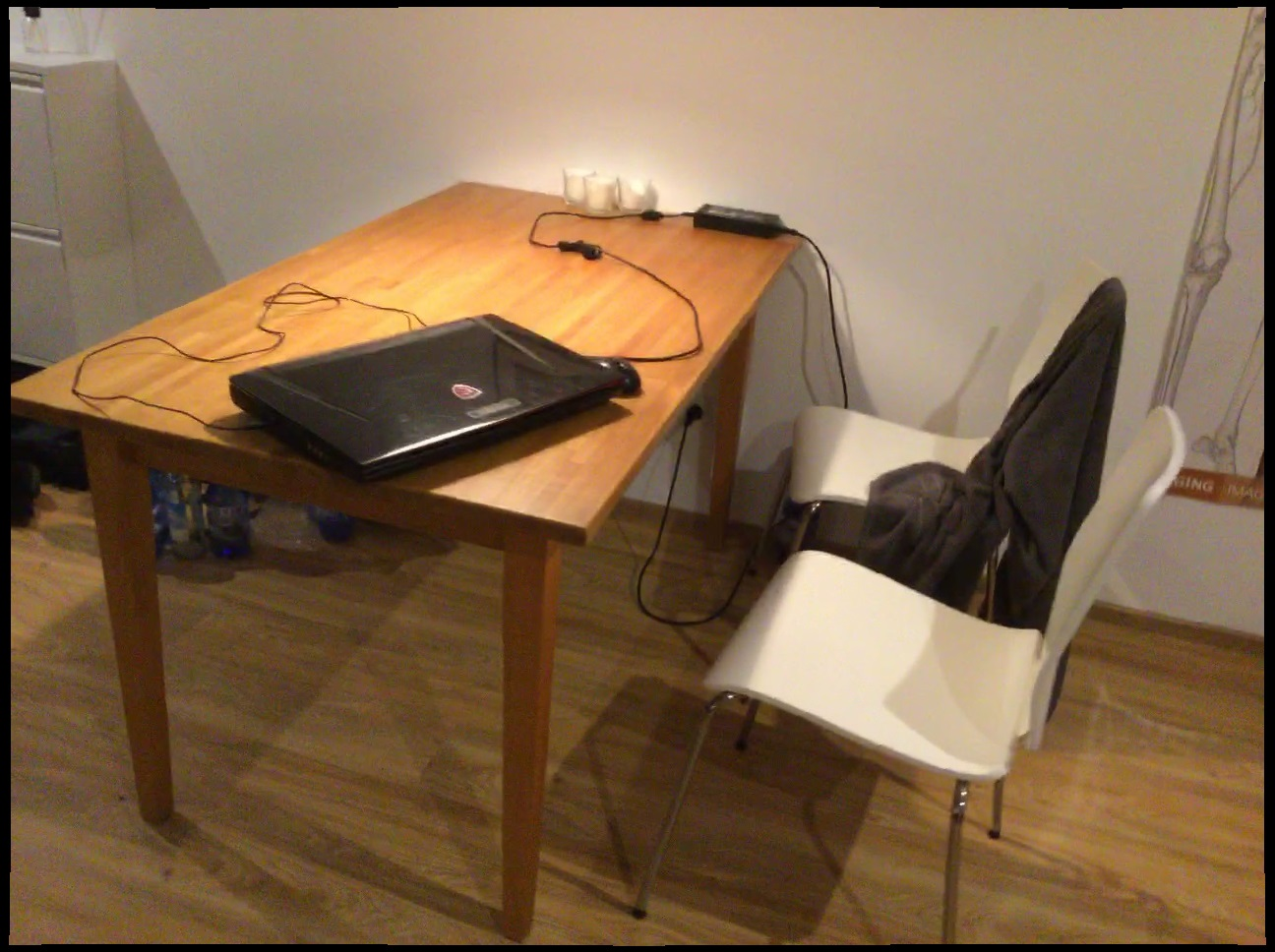}}};%
        \node at (gt_ndt_05_panoptic |- gt_panoptic) (rgb) {\includeimagedd{img/examples/scannet_example_0/rgb.png}};%
        \node at (gt_ndt_05_semantic |- gt_panoptic) (depth) {\includeimagedd{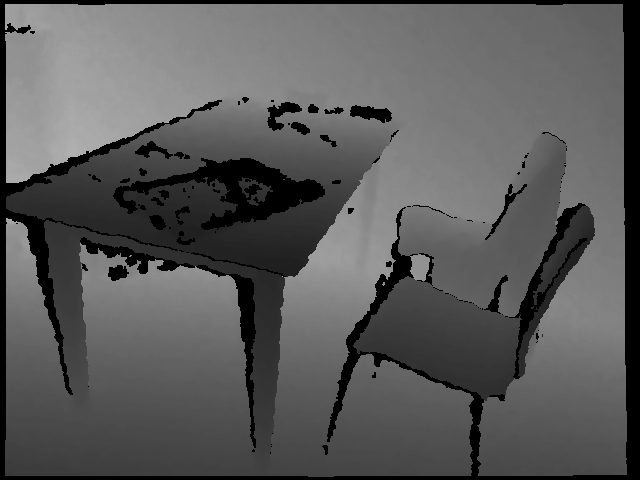}};%
        \node[below=2.5mm of pred_pmtsdf_instance_2d, anchor=mid] (l) {\scriptsize 2D instance};%
        \node[anchor=mid] at (l -| pred_pmtsdf_semantic_2d){\scriptsize 2D semantic};%
        \node[anchor=mid] at (l -| pred_pmtsdf_panoptic_2d){\scriptsize 2D panoptic};%
        \node[anchor=mid] at (l -| pred_pmtsdf_instance){\scriptsize 3D instance};%
        \node[anchor=mid] at (l -| pred_pmtsdf_semantic){\scriptsize 3D semantic};%
        \node[anchor=mid] at (l -| pred_pmtsdf_panoptic){\scriptsize 3D panoptic};%
        \node[below=2mm of rgb, anchor=mid] {\scriptsize RGB};%
        \node[below=2mm of depth, anchor=mid] {\scriptsize Depth};%
        \node[left=2.5mm of emsanet_panoptic, rotate=90, anchor=mid] {\scriptsize EMSANet};%
        \node[left=2.5mm of pred_ndt_05_panoptic, rotate=90, anchor=mid] {\scriptsize EMSANet};%
        \node[left=2.5mm of pred_ndt_10_panoptic, rotate=90, anchor=mid] {\scriptsize EMSANet};%
        \node[left=2.5mm of pred_pmtsdf_panoptic, rotate=90, anchor=mid] {\scriptsize EMSANet};%
        \node[left=7.5mm of pred_ndt_05_panoptic, rotate=90, anchor=mid, align=center] (l) {\footnotesize PanopticNDT\\[-1mm]\footnotesize(5cm)};%
        \node[left=7.5mm of pred_ndt_10_panoptic, rotate=90, anchor=mid, align=center] (l) {\footnotesize \bf PanopticNDT\\[-1mm]\footnotesize\bf (10cm)};%
        \node[left=7.5mm of pred_pmtsdf_panoptic, rotate=90, anchor=mid, align=center] (l) {\footnotesize Panoptic\\[-1mm]\footnotesize Multi-TSDFs};%
        \coordinate (p) at ($(gt_panoptic)!0.5!(emsanet_panoptic)$);%
        \node[rotate=90, anchor=mid, align=center] at (l.mid |- p) (lsf) {\footnotesize Single frame\\[-1mm]\footnotesize(given camera pose)};%
        \coordinate (p) at ([yshift=-2.5mm]lsf.north west |- emsanet_panoptic.south west);%
        \draw[gray!80, dashed, thick] (p) -- (p -| emsanet_instance.south east);%
        \draw[gray!80, dashed, thick] ([yshift=-2.5mm]pred_ndt_05_panoptic.south west) -- ([yshift=-2.5mm]pred_ndt_05_instance_2d.south east);%
        \draw[gray!80, dashed, thick] ([yshift=-2.5mm]pred_ndt_10_panoptic.south west) -- ([yshift=-2.5mm]pred_ndt_10_instance_2d.south east);%
        \coordinate (p) at ([yshift=-5.5mm]lsf.north west |- pred_pmtsdf_panoptic.south west);%
        \draw[thick] (p) -- (p -| pred_pmtsdf_instance_2d.south east);%
    \end{tikzpicture}\\[2.5mm]
    \begin{tikzpicture}[inner sep=0pt]     
        \newcommand{\includeimageddd}[1]{\includegraphics[height=1.8cm, trim={0 0.5cm 0 3.2cm}, clip]{#1}}%
        \newcommand{\includeimagedd}[1]{\includegraphics[height=1.8cm]{#1}}%
        \node (pred_ndt_05_panoptic) {\includeimageddd{img/examples/scannet_example_0/pred_ndt_05_panoptic.png}};%
        \node[right=0.5mm of pred_ndt_05_panoptic] (pred_ndt_05_semantic) {\includeimageddd{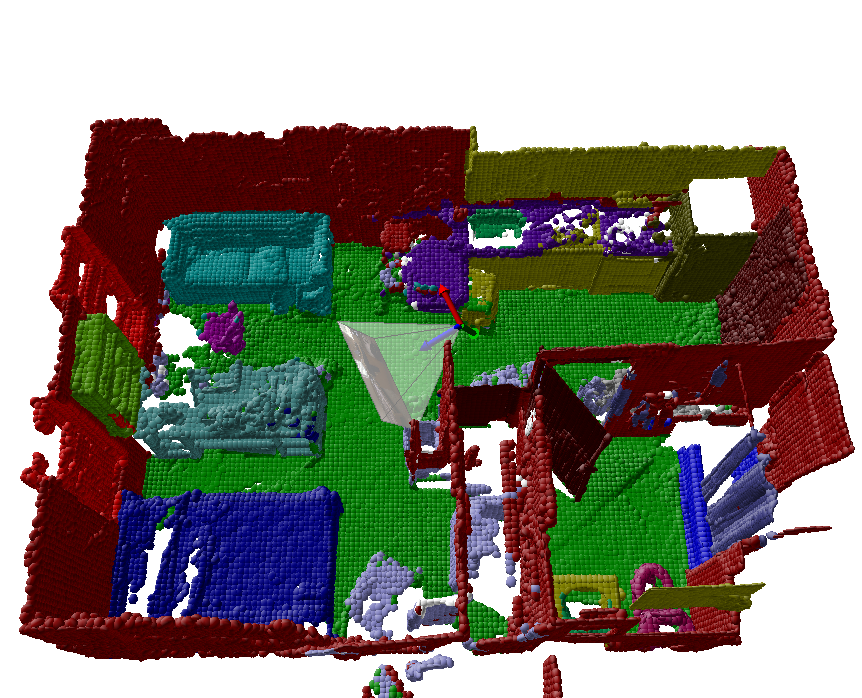}};%
        \node[right=0.5mm of pred_ndt_05_semantic] (pred_ndt_05_instance) {\includeimageddd{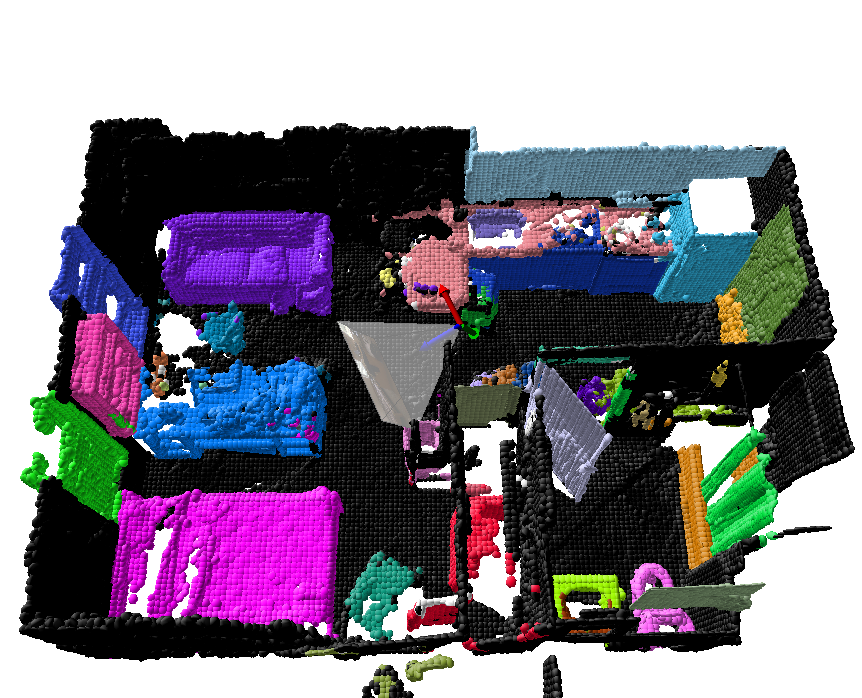}};%
        \node[right=4mm of pred_ndt_05_instance] (pred_ndt_05_panoptic_2d) {\includeimagedd{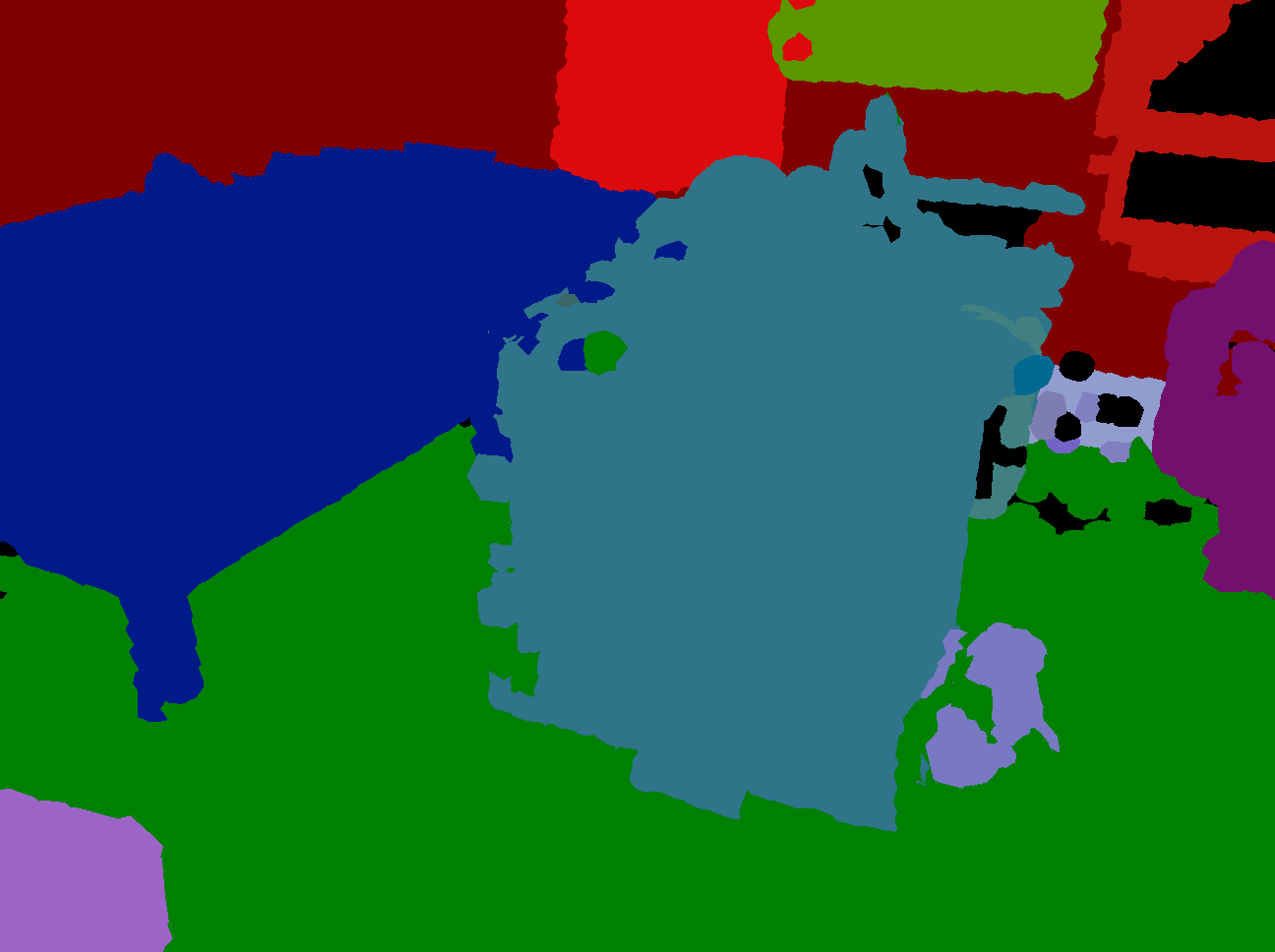}};%
        \node[right=2mm of pred_ndt_05_panoptic_2d] (pred_ndt_05_semantic_2d) {\includeimagedd{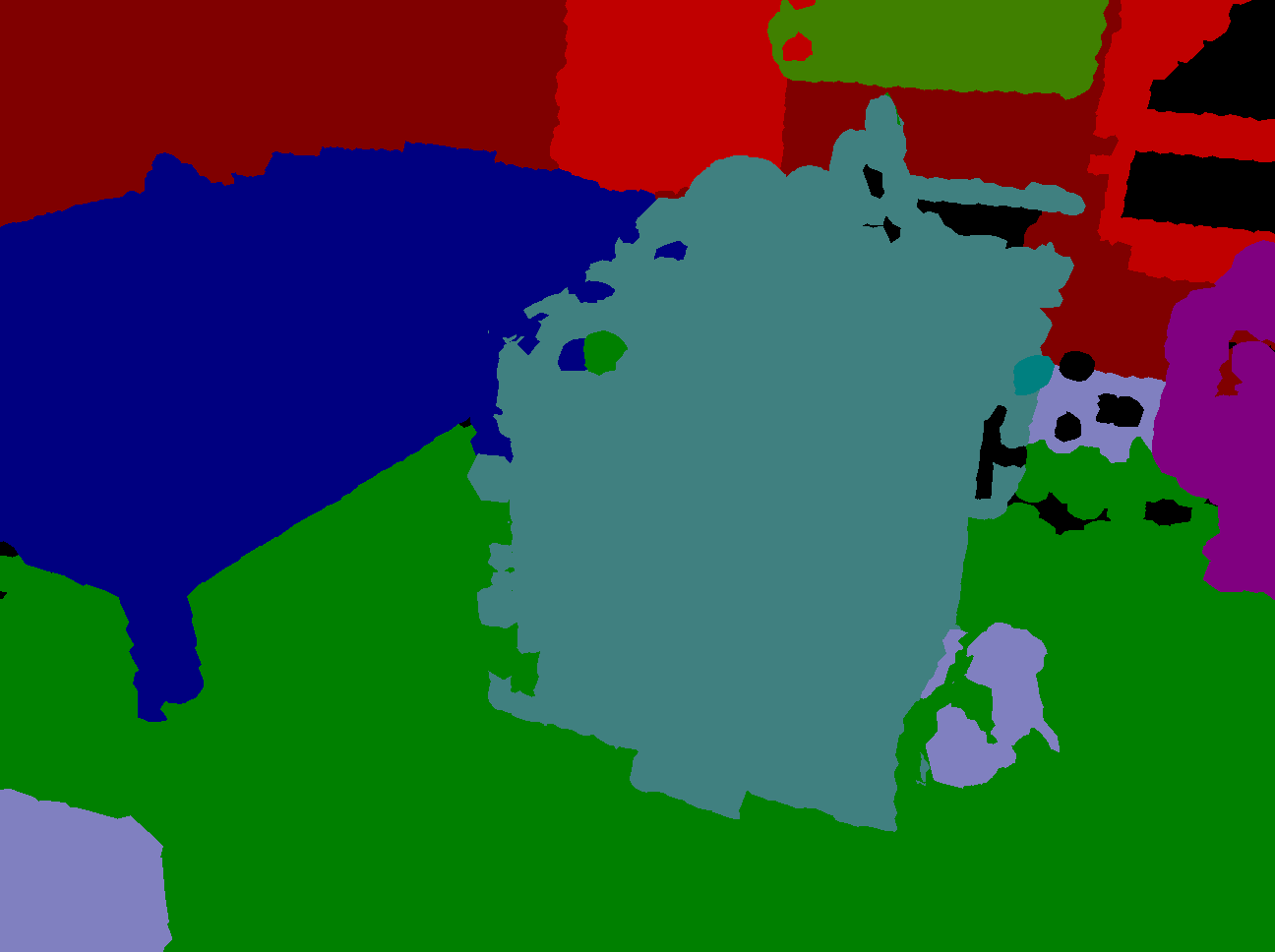}};%
        \node[right=2mm of pred_ndt_05_semantic_2d] (pred_ndt_05_instance_2d) {\includeimagedd{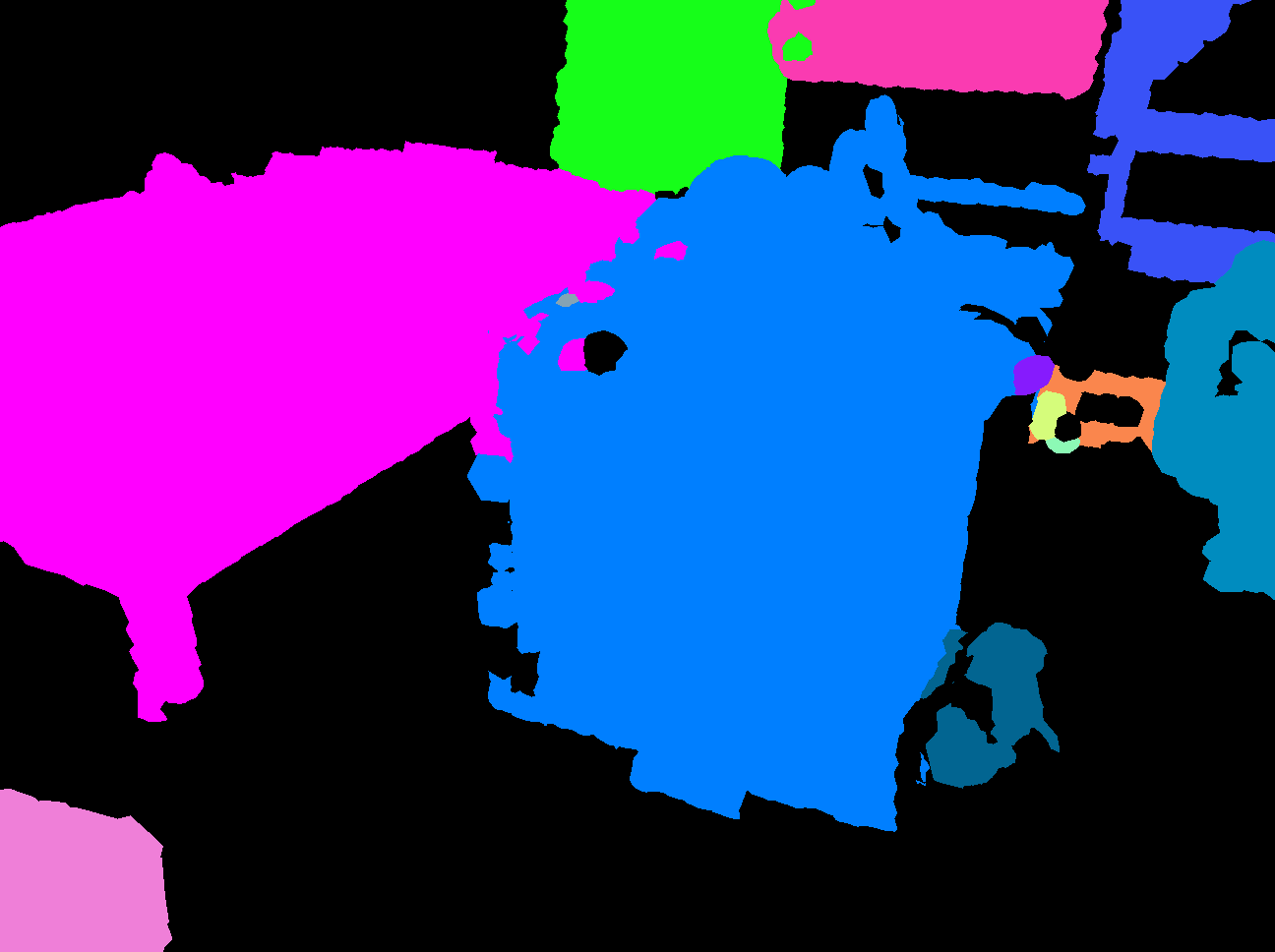}};%
        \node[below=5mm of pred_ndt_05_panoptic] (pred_ndt_10_panoptic) {\includeimageddd{img/examples/scannet_example_1/pred_ndt_10_panoptic.png}};%
        \node[right=0.5mm of pred_ndt_10_panoptic] (pred_ndt_10_semantic) {\includeimageddd{img/examples/scannet_example_1/pred_ndt_10_semantic.png}};%
        \node[right=0.5mm of pred_ndt_10_semantic] (pred_ndt_10_instance) {\includeimageddd{img/examples/scannet_example_1/pred_ndt_10_instance.png}};%
        \node[right=4mm of pred_ndt_10_instance] (pred_ndt_10_panoptic_2d) {\includeimagedd{img/examples/scannet_example_1/pred_ndt_10_backprojection_05180_panoptic.png}};%
        \node[right=2mm of pred_ndt_10_panoptic_2d] (pred_ndt_10_semantic_2d) {\includeimagedd{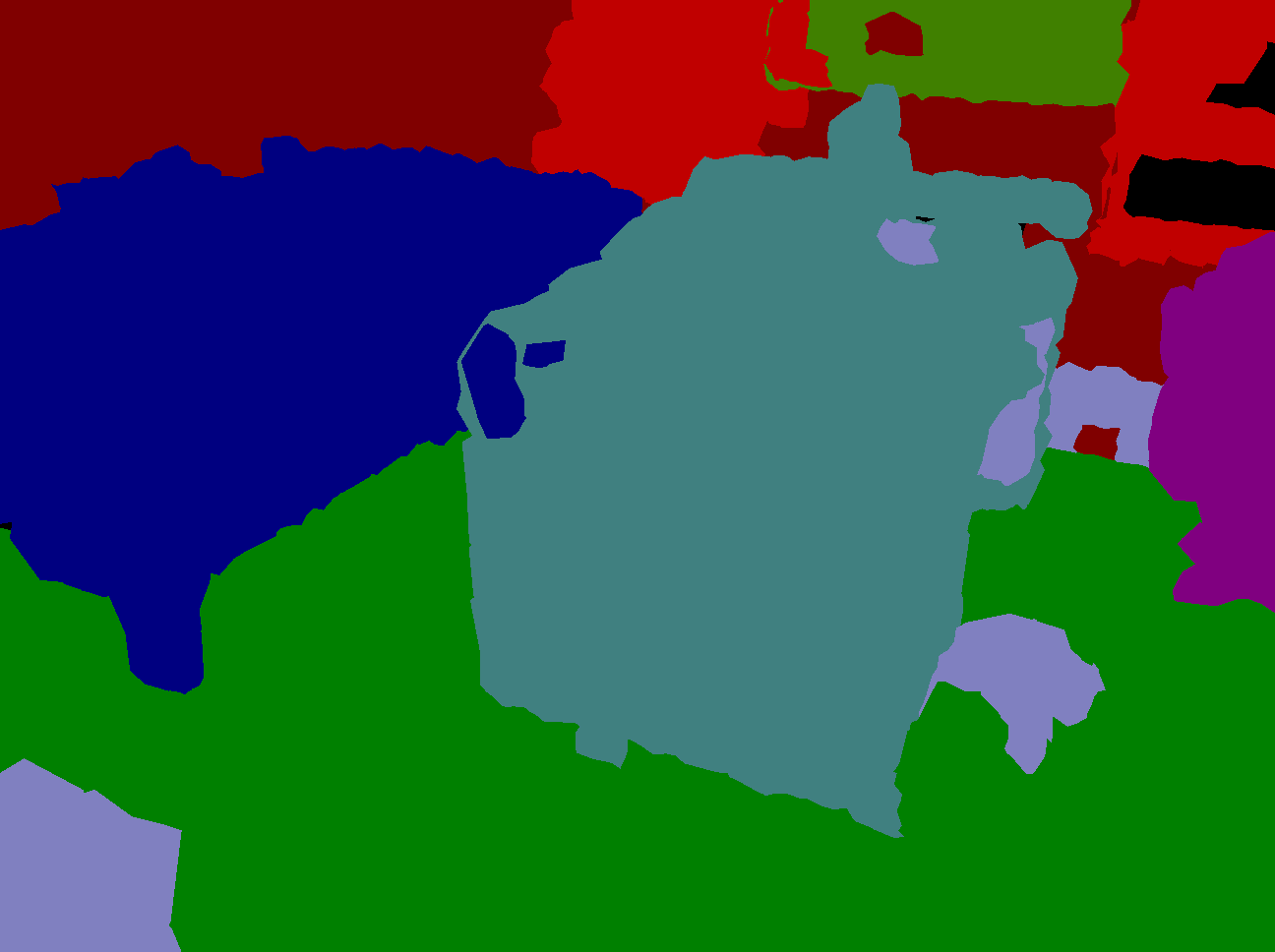}};%
        \node[right=2mm of pred_ndt_10_semantic_2d] (pred_ndt_10_instance_2d) {\includeimagedd{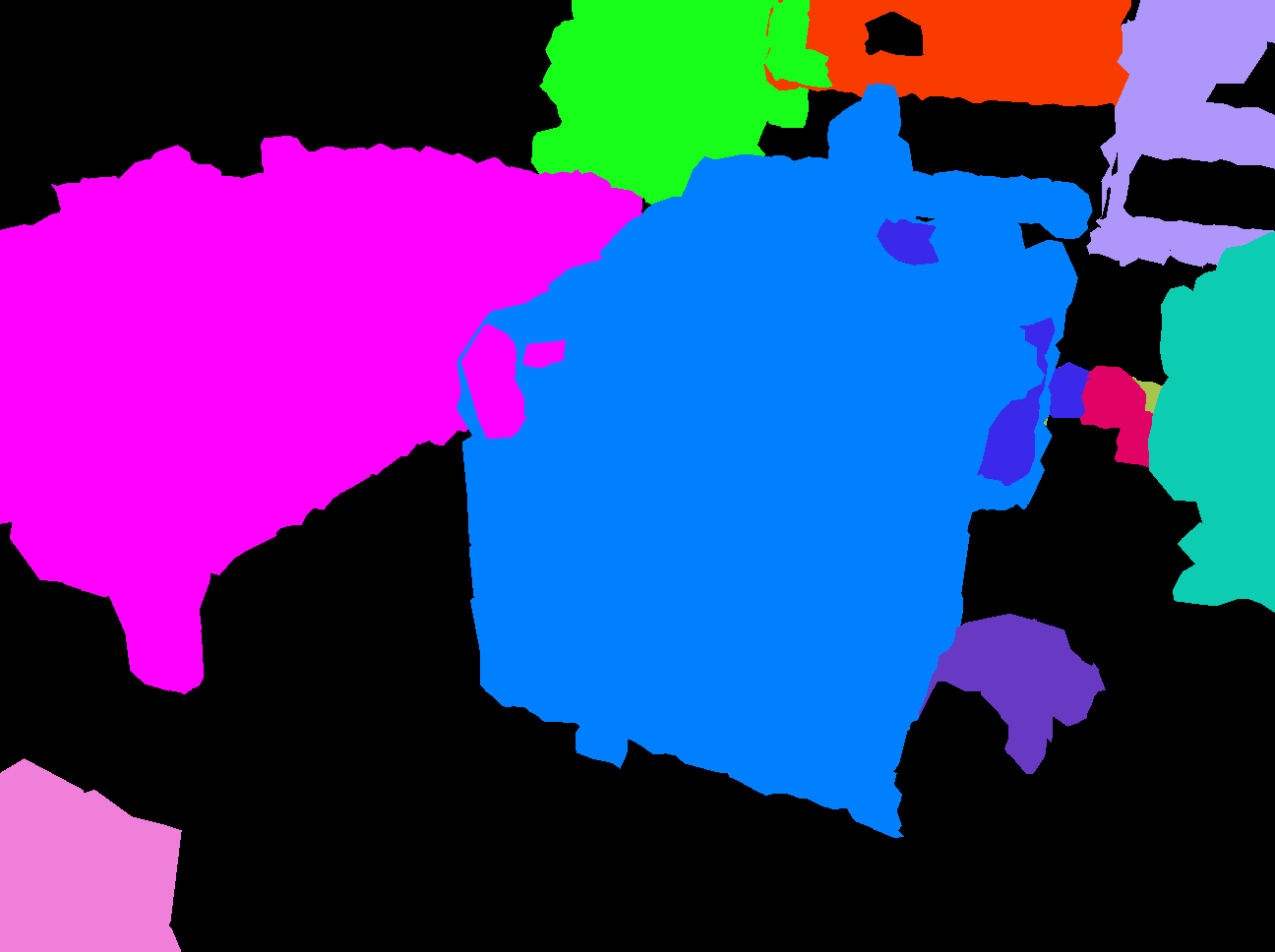}};%
        \node[below=5mm of pred_ndt_10_panoptic] (pred_pmtsdf_panoptic) {\includeimageddd{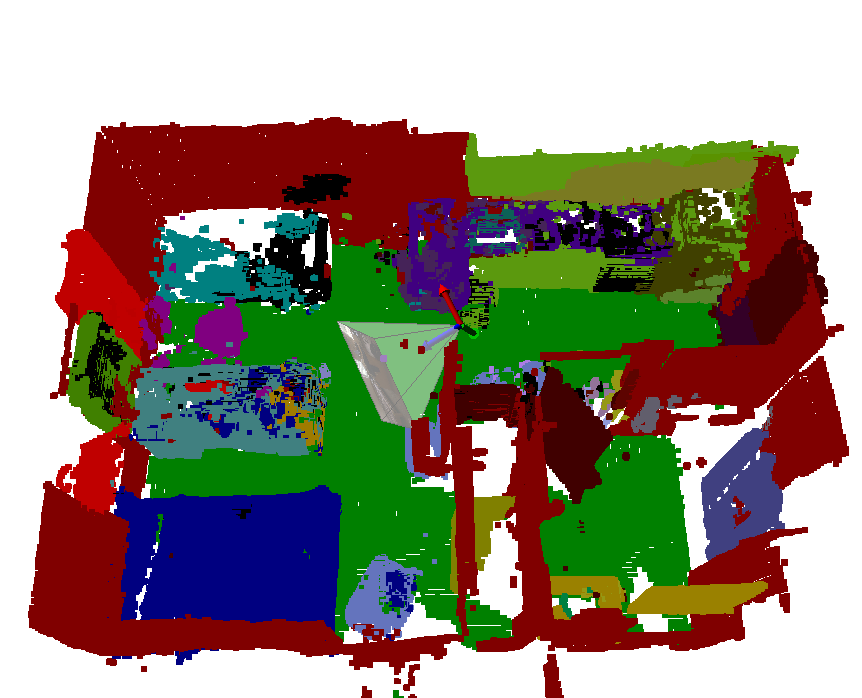}};%
        \node[right=0.5mm of pred_pmtsdf_panoptic] (pred_pmtsdf_semantic) {\includeimageddd{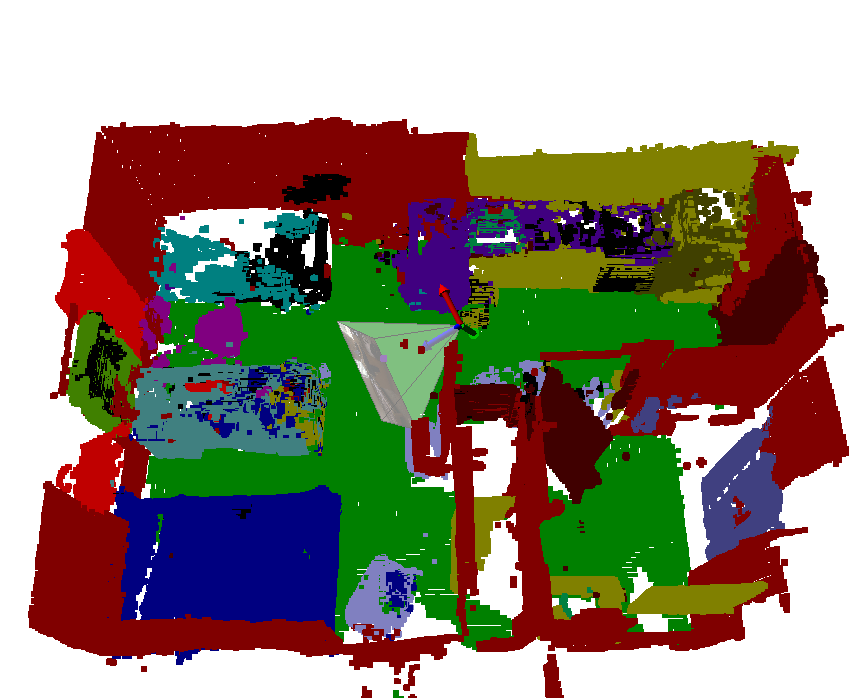}};%
        \node[right=0.5mm of pred_pmtsdf_semantic] (pred_pmtsdf_instance) {\includeimageddd{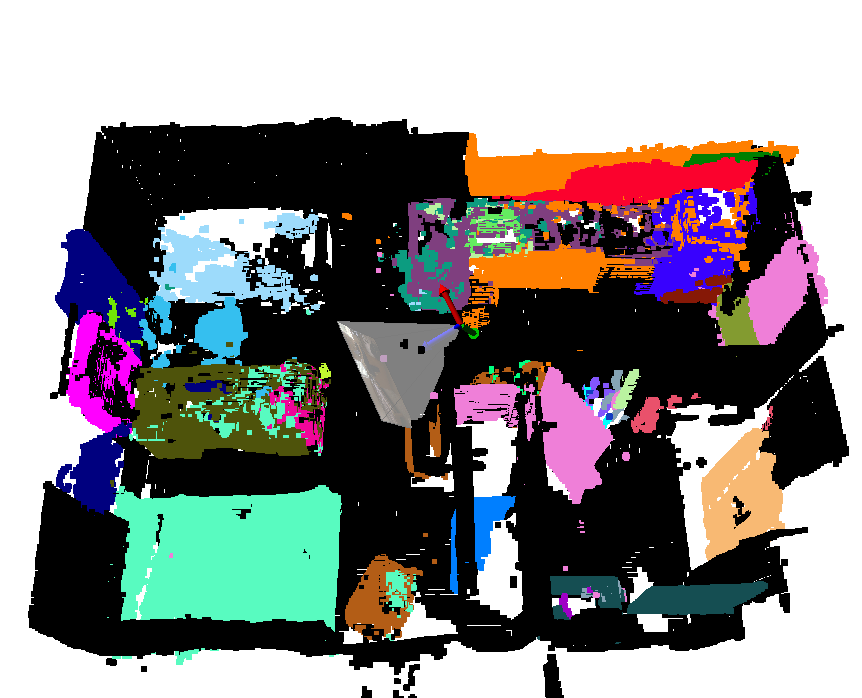}};%
        \node[right=4mm of pred_pmtsdf_instance] (pred_pmtsdf_panoptic_2d) {\includeimagedd{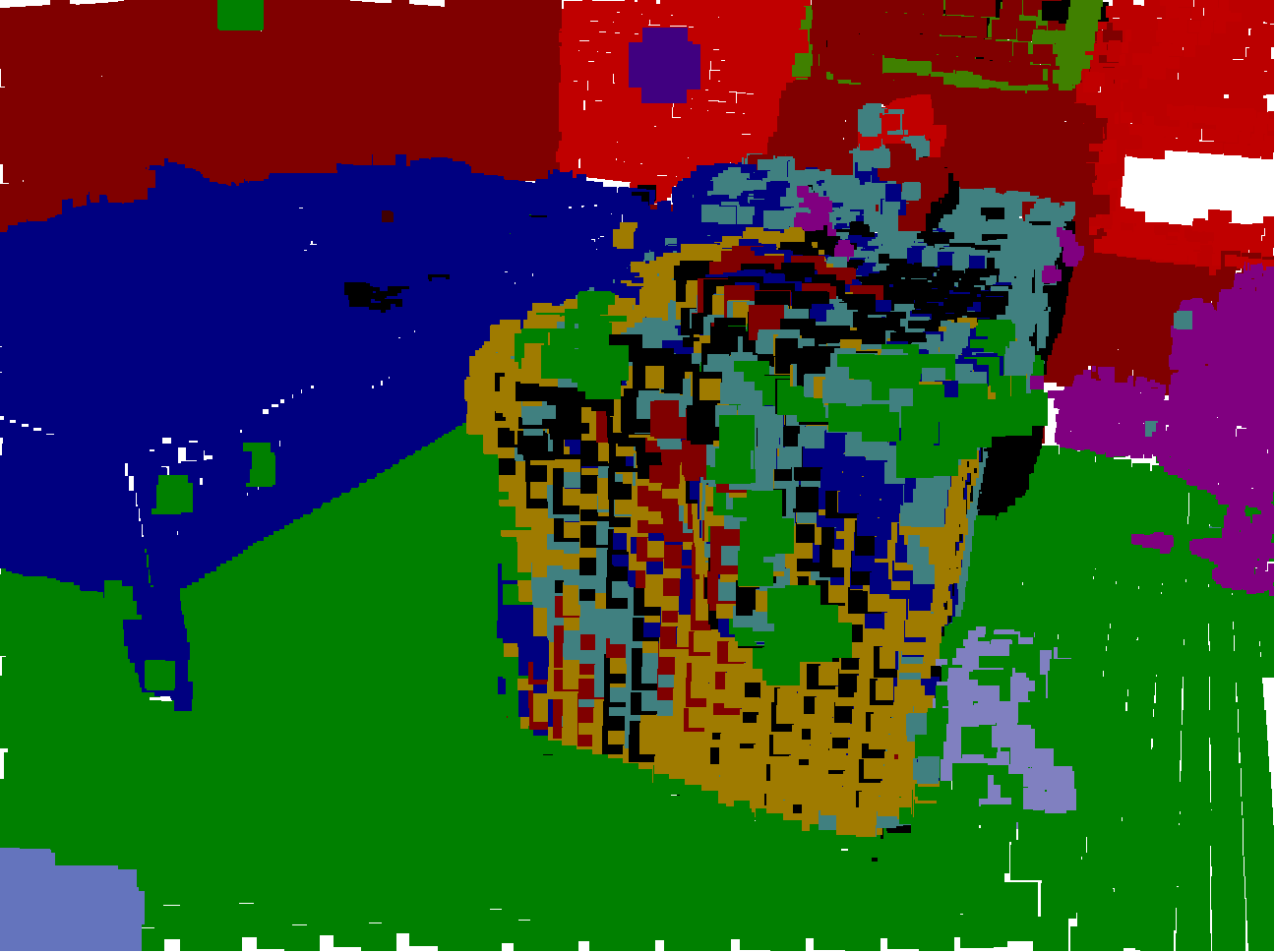}};%
        \node[right=2mm of pred_pmtsdf_panoptic_2d] (pred_pmtsdf_semantic_2d) {\includeimagedd{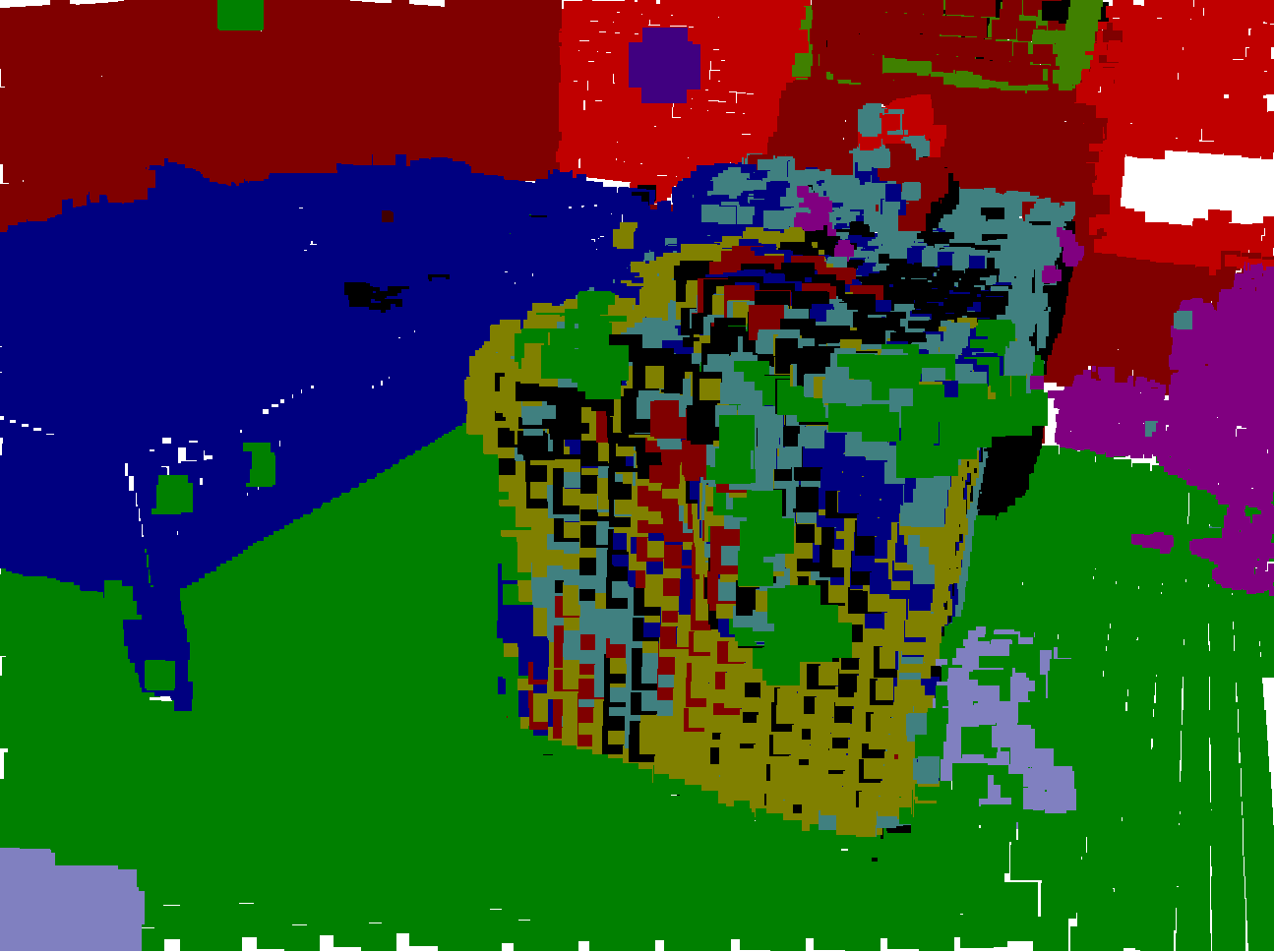}};%
        \node[right=2mm of pred_pmtsdf_semantic_2d] (pred_pmtsdf_instance_2d) {\includeimagedd{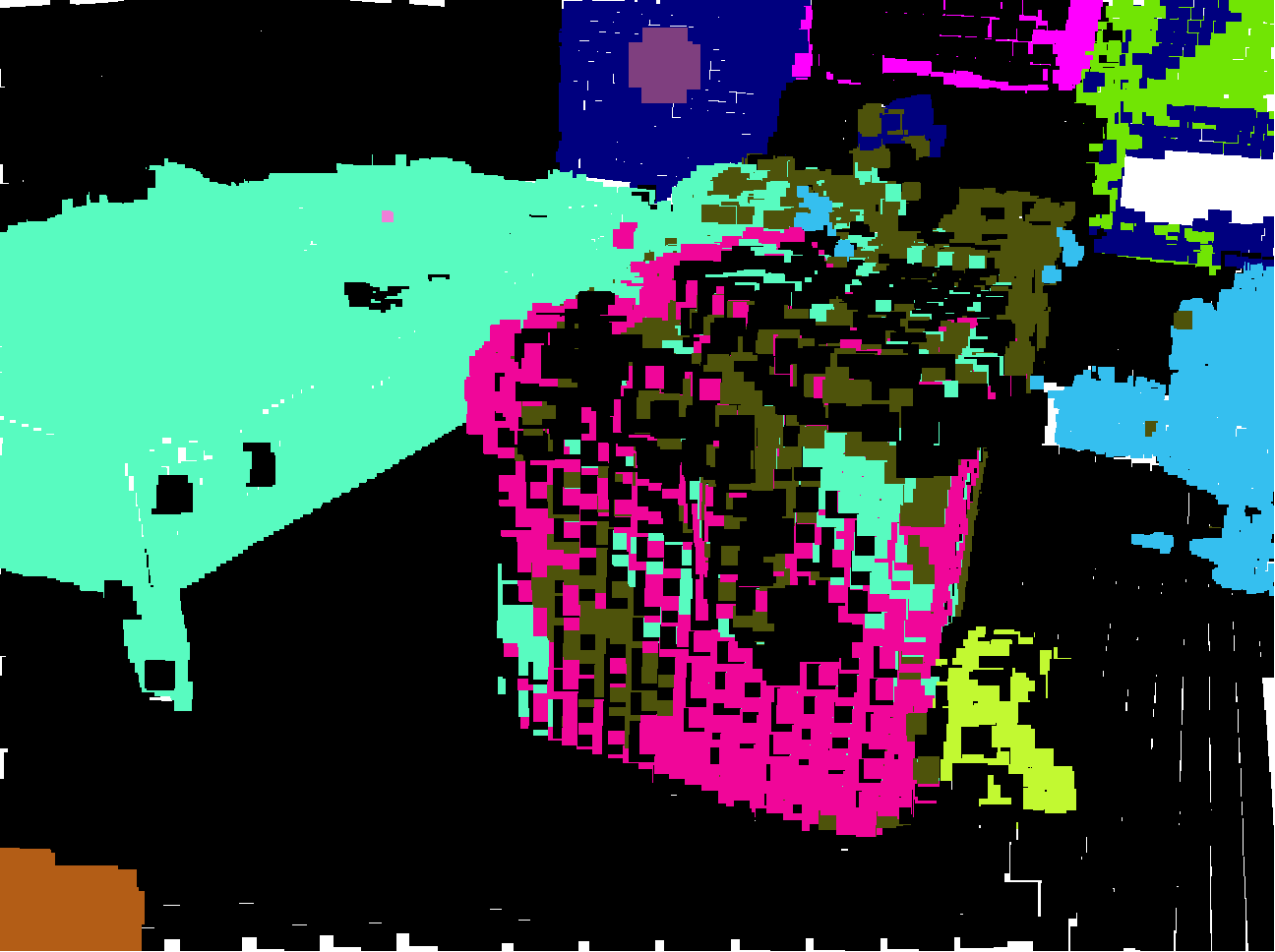}};%
        \node[above=5mm of pred_ndt_05_panoptic_2d] (emsanet_panoptic) {\includeimagedd{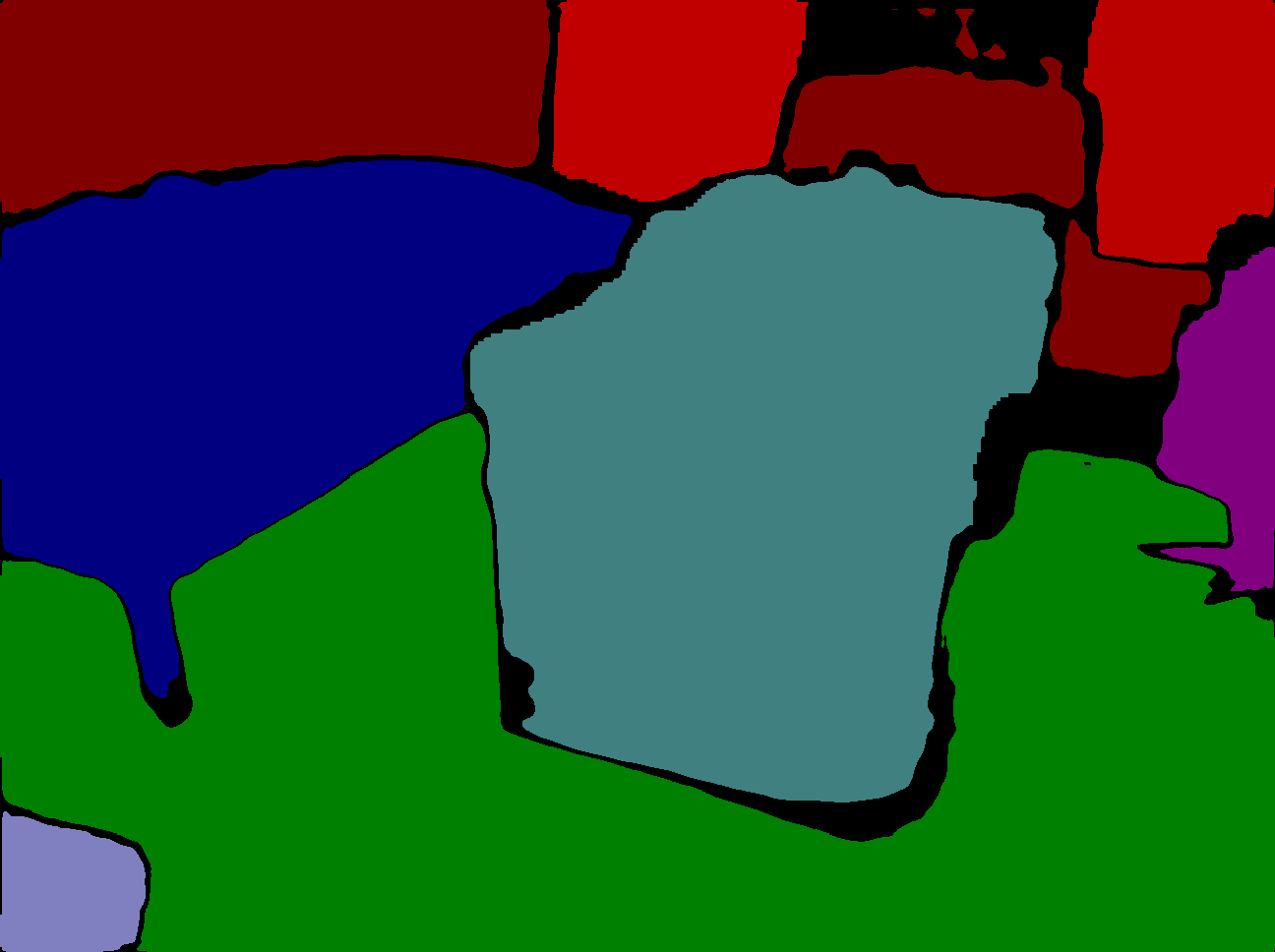}};%
        \node[above=5mm of pred_ndt_05_semantic_2d] (emsanet_semantic) {\includeimagedd{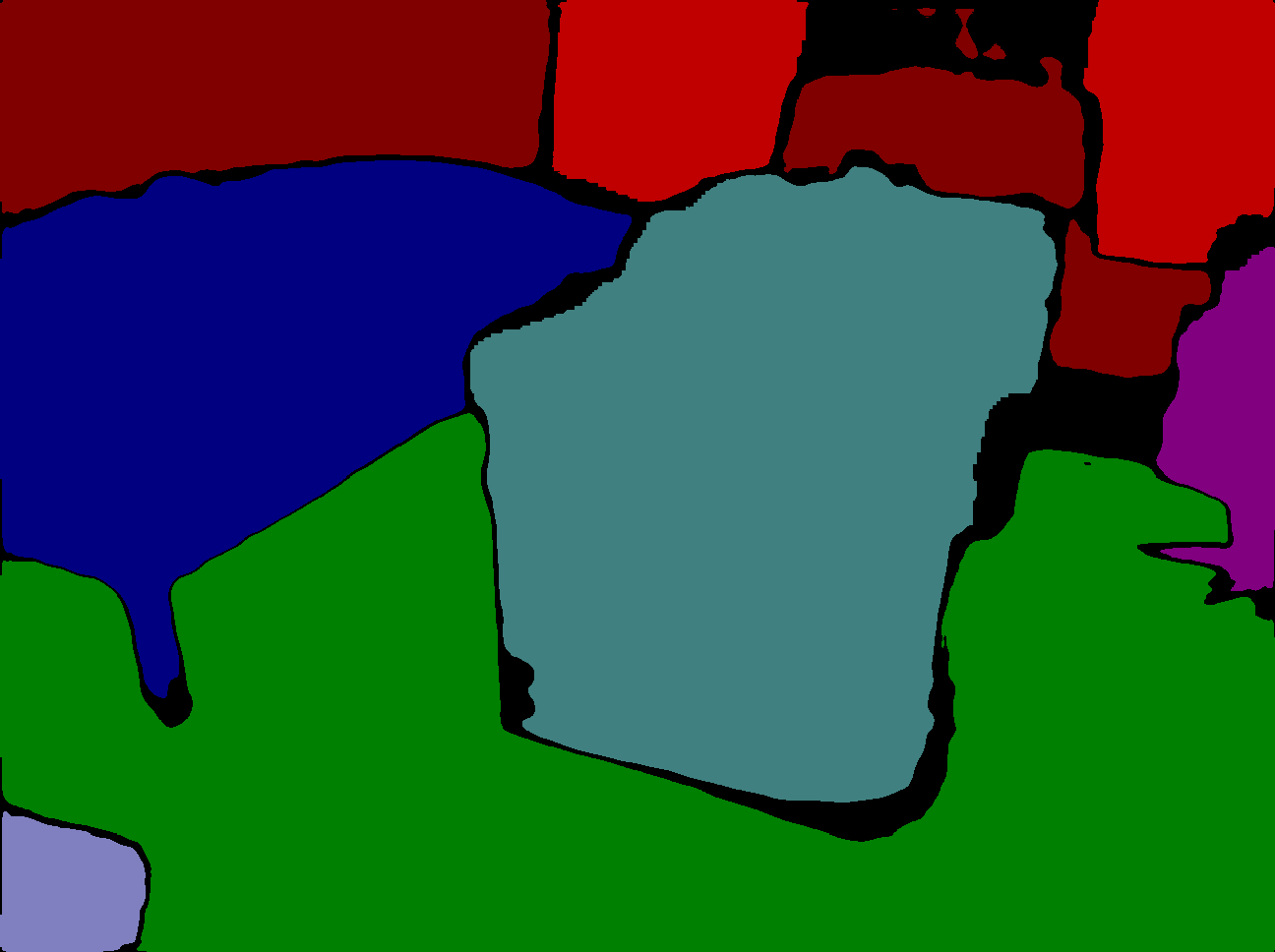}};%
        \node[above=5mm of pred_ndt_05_instance_2d] (emsanet_instance) {\includeimagedd{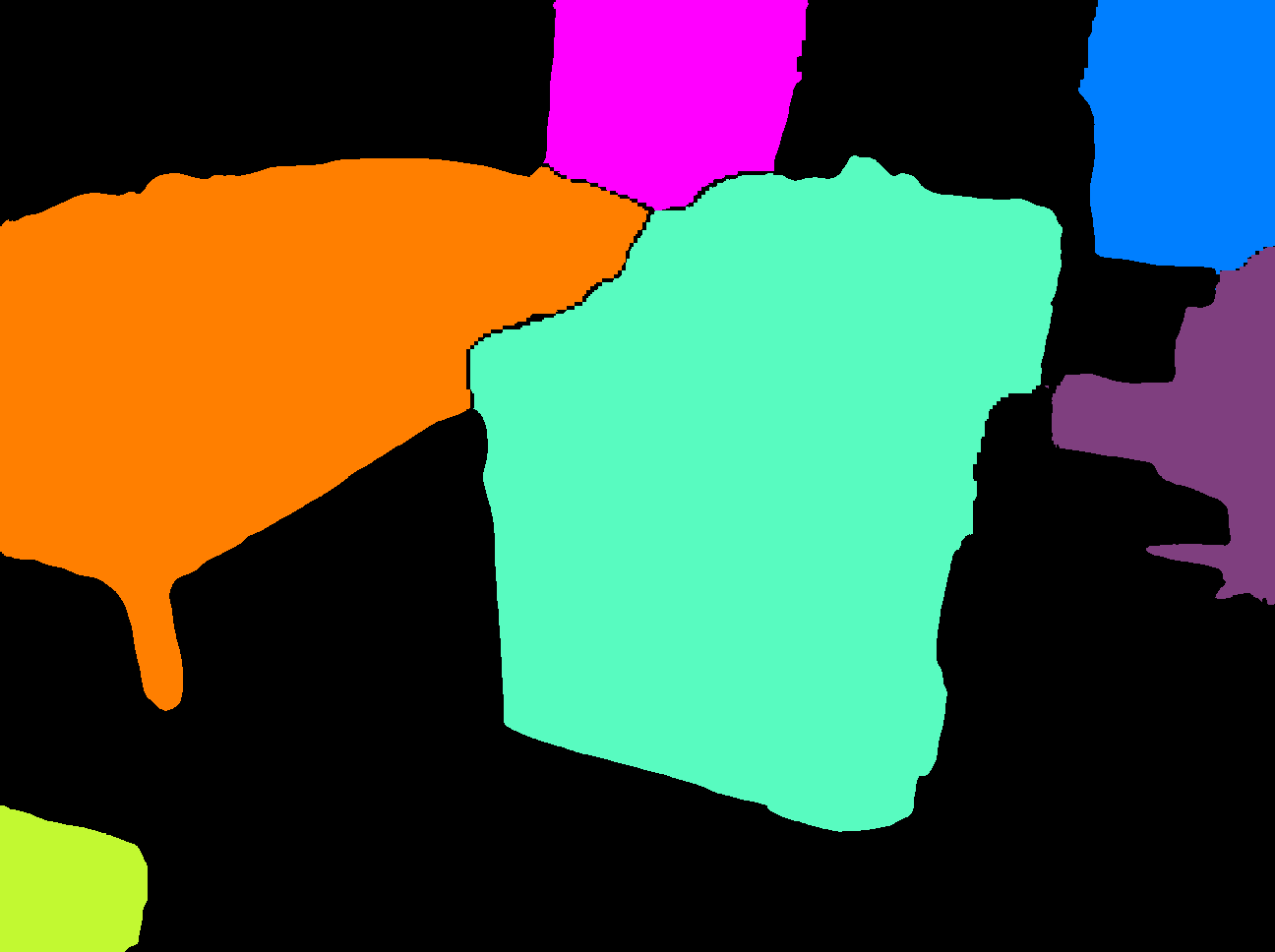}};%
        \node[above=2mm of emsanet_panoptic] (gt_panoptic) {\phantom{\includeimagedd{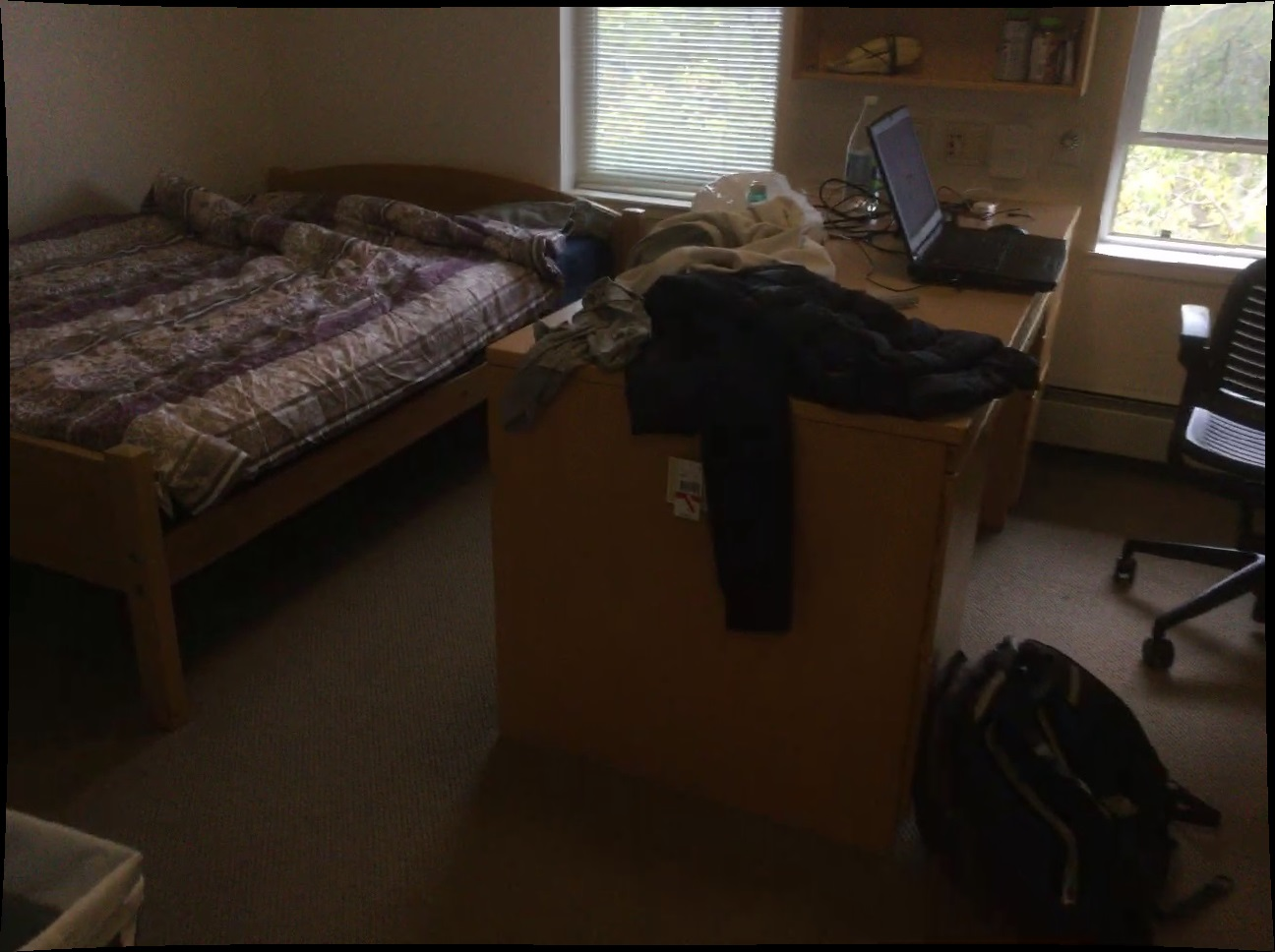}}};%
        \node at (gt_ndt_05_panoptic |- gt_panoptic) (rgb) {\includeimagedd{img/examples/scannet_example_1/rgb.png}};%
        \node at (gt_ndt_05_semantic |- gt_panoptic) (depth) {\includeimagedd{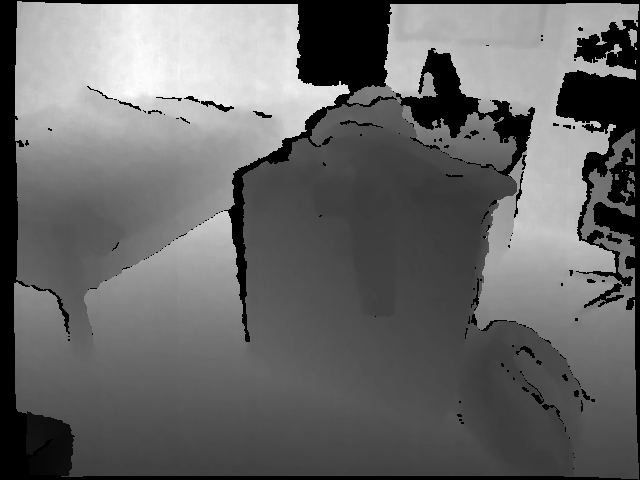}};%
        \node[below=2.5mm of pred_pmtsdf_instance_2d, anchor=mid] (l) {\scriptsize 2D instance};%
        \node[anchor=mid] at (l -| pred_pmtsdf_semantic_2d){\scriptsize 2D semantic};%
        \node[anchor=mid] at (l -| pred_pmtsdf_panoptic_2d){\scriptsize 2D panoptic};%
        \node[anchor=mid] at (l -| pred_pmtsdf_instance){\scriptsize 3D instance};%
        \node[anchor=mid] at (l -| pred_pmtsdf_semantic){\scriptsize 3D semantic};%
        \node[anchor=mid] at (l -| pred_pmtsdf_panoptic){\scriptsize 3D panoptic};%
        \node[below=2mm of rgb, anchor=mid] {\scriptsize RGB};%
        \node[below=2mm of depth, anchor=mid] {\scriptsize Depth};%
        \node[left=2.5mm of emsanet_panoptic, rotate=90, anchor=mid] {\scriptsize EMSANet};%
        \node[left=2.5mm of pred_ndt_05_panoptic, rotate=90, anchor=mid] {\scriptsize EMSANet};%
        \node[left=2.5mm of pred_ndt_10_panoptic, rotate=90, anchor=mid] {\scriptsize EMSANet};%
        \node[left=2.5mm of pred_pmtsdf_panoptic, rotate=90, anchor=mid] {\scriptsize EMSANet};%
        \node[left=7.5mm of pred_ndt_05_panoptic, rotate=90, anchor=mid, align=center] (l) {\footnotesize PanopticNDT\\[-1mm]\footnotesize(5cm)};%
        \node[left=7.5mm of pred_ndt_10_panoptic, rotate=90, anchor=mid, align=center] (l) {\footnotesize \bf PanopticNDT\\[-1mm]\footnotesize\bf (10cm)};%
        \node[left=7.5mm of pred_pmtsdf_panoptic, rotate=90, anchor=mid, align=center] (l) {\footnotesize Panoptic\\[-1mm]\footnotesize Multi-TSDFs};%
        \coordinate (p) at ($(gt_panoptic)!0.5!(emsanet_panoptic)$);%
        \node[rotate=90, anchor=mid, align=center] at (l.mid |- p) (lsf) {\footnotesize Single frame\\[-1mm]\footnotesize(given camera pose)};%
        \coordinate (p) at ([yshift=-2.5mm]lsf.north west |- emsanet_panoptic.south west);%
        \draw[gray!80, dashed, thick] (p) -- (p -| emsanet_instance.south east);%
        \draw[gray!80, dashed, thick] ([yshift=-2.5mm]pred_ndt_05_panoptic.south west) -- ([yshift=-2.5mm]pred_ndt_05_instance_2d.south east);%
        \draw[gray!80, dashed, thick] ([yshift=-2.5mm]pred_ndt_10_panoptic.south west) -- ([yshift=-2.5mm]pred_ndt_10_instance_2d.south east);%
    \end{tikzpicture}%
    \vspace{-3mm}%
    \caption{%
        Qualitative results for \emph{scene\_0757\_00}~(top) and \emph{scene\_0761\_00}~(bottom) of the hidden ScanNetV2 test split. %
        The upper part for each scene shows the thresholded predictions of EMSANet~\cite{emsanet2022ijcnn}~(see Sec.~\ref{sec:experiments:implementation}). %
        The lower part for each scene compares our proposed PanopticNDT with voxel sizes 5\si{\centi\meter} and 10\si{\centi\meter} to Panoptic Multi-TSDFs~\cite{panoptic-multi-tsdf-2022-icra}. %
        Best viewed in color at 300\%. %
        Black indicates \emph{void/no\_instance}, for the semantic colors, we refer to Fig.~\ref{fig:experiments:hypersim_radar_chart}. %
        Panoptic labels are visualized by small color differences based on the semantic color.%
    }        
    \label{fig:appendix:qualitative_results_scannet}
\end{figure*}
}{}

\end{document}